\documentclass[12pt]{article}
\usepackage[margin=1in]{geometry}
 
\usepackage{times,upgreek,rotating}

\usepackage[utf8]{inputenc} % allow utf-8 input
\usepackage[T1]{fontenc}    % use 8-bit T1 fonts
\usepackage{url}            % simple URL typesetting
\usepackage{booktabs}       % professional-quality tables
\usepackage{amsfonts}       % blackboard math symbols
\usepackage{nicefrac}       % compact symbols for 1/2, etc.
\usepackage{microtype}      % microtypography

\usepackage{rotating}
\usepackage{array,colortbl,multirow}
\usepackage{textcomp}

\usepackage{upgreek}

\usepackage{graphicx}
\usepackage[dvips]{epsfig}
\usepackage{color,soul,textcomp,xcolor}

\usepackage{array,natbib}
\usepackage{graphicx} 
\usepackage{subfigure} 
\usepackage{algorithm}
\usepackage{algorithmic}
\usepackage{eucal,upgreek}
\usepackage{amsfonts}       % blackboard math symbols
\usepackage{amsmath}
\usepackage{xspace,url}

\title{Semi-parametric Network Structure Discovery Models}

\author{
  Amir Dezfouli\\
{\normalsize The University of New South Wales}\\
  \texttt{akdezfuli@gmail.com}\\
\and
Edwin V. Bonilla\\
{\normalsize The University of New South Wales}\\
  \texttt{e.bonilla@unsw.edu.au}\\
\and
  Richard Nock\\
{\normalsize Data61, The Australian National University \& The
  University of Sydney}\\
  \texttt{richard.nock@data61.csiro.au}
}
\date{}

\newcommand{\Eq}{Eq.~}
\newcommand{\eqs}{eqs.~}
\newcommand{\eq}{eq.~}
\newcommand{\supplement}{appendix\xspace}

\newcommand{\mat}[1]{\mathbf{#1}}						% Matrix notation
\renewcommand{\vec}[1]{\mathbf{#1}}						% Vector notation
\newcommand{\I}{\mat{I}}								% Identity matrix
\newcommand{\hada}{\odot}          						% Hadamard product
\newcommand{\matentry}[3]{{[#1]_{#2,#3}}}				% single entry of matrix
\newcommand{\matrow}[2]{[{#1}]_{#2,:}}					% Row of matrix
\newcommand{\vect}{\text{vec}}							% Vec operation 
\newcommand{\kron}{\otimes}								% Kronecker product (outer)
\newcommand{\trace}{\text{tr}}							% Trace of a matrix
\newcommand{\mth}{\text{th}}						

\newcommand{\z}{\bz}
\newcommand{\Y}{\mat{Y}}							% Matrix of targets
\newcommand{\y}{\vec{y}}							% Vector of targets
\newcommand{\f}{\vec{f}}							% Vector of latent functions
\newcommand{\A}{\bA} 								% Adjacency matrix
\newcommand{\W}{\bW}								% Edge eights
\newcommand{\epsilony}{\epsilon_y}					% Standard Gaussian added to y
\newcommand{\varnoise}{\sigma_y^2}					% Observarion noise variance
\newcommand{\varw}{\sigma_w^2}						% Prior variance on W
\newcommand{\epsilonf}{\epsilon_f}					% Standard Gaussian added to f 
\newcommand{\bepsilonf}{\boldsymbol{\epsilon}_f}	% vector above
\newcommand{\varf}{\sigma_f^2}						% latent noise variance
\newcommand{\Kf}{\mathbf{K}_f}						% Covariance between latent functions
\newcommand{\B}{\mat{B}}							% B = A \hada W
\newcommand{\G}{\mat{G}}							% G = (I - B)^{-1} 
\newcommand{\E}{\mathbf{E}}							% E = L B B^T L^T
\newcommand{\Covy}{\boldsymbol{\Sigma}_y}			% Covarinace matrix of y
\newcommand{\Kt}{\mathbf{K}_t}						% Time covariance
\newcommand{\matLambda}{\mat{\Lambda}}
\newcommand{\Q}{\mat{Q}}
\newcommand{\Covnoise}{\boldsymbol{\Omega}}         % Noise covarinace Omega = varf E + varnoise I
\newcommand{\Qnoise}{\Q_\Omega} 
\newcommand{\Lambdanoise}{\matLambda_\Omega} 
\newcommand{\lambdanoise}{\lambda_\Omega}           % Scalar vesion of Lambdanoise
\newcommand{\Sigmai}{\boldsymbol{\Sigma}_\text{I}}
\newcommand{\Qi}{\mat{Q}_\text{I}}
\newcommand{\Lambdai}{\matLambda_\text{I}}

\newcommand{\Ytilde}{\tilde{\mat{Y}}}
\newcommand{\ytilde}{\tilde{\vec{y}}}
\newcommand{\Ktildef}{\tilde{\mat{K}}_f}
\newcommand{\Ktildet}{\tilde{\mat{K}}_t}
\newcommand{\Lambdatildef}{\tilde{\matLambda}_f}
\newcommand{\Lambdatildet}{\tilde{\matLambda}_t}
\newcommand{\lambdatilde}{\tilde{\lambda}}
\newcommand{\lambdatildef}{\lambdatilde_f}			% Scalar eigenvalue of Kftilde
\newcommand{\lambdatildet}{\lambdatilde_t}			% Scalar eigenvalue of Kttilde
\newcommand{\blambdatildef}{\boldsymbol{\lambdatilde}_f}			% Scalar eigenvalue of Kftilde
\newcommand{\blambdatildet}{\boldsymbol{\lambdatilde}_t}			% Scalar eigenvalue of Kttilde
\newcommand{\Qtildef}{\tilde{\Q}_f}
\newcommand{\Qtildet}{\tilde{\Q}_t}
\newcommand{\Ytildetf}{\tilde{\mat{Y}}_{tf}}

\newcommand{\meana}{\rho}			% prior mean over A_{ij}
\newcommand{\posmeanw}{\mu}			% posterior mean over W
\newcommand{\posstdw}{\sigma}		% posterior std over W
\newcommand{\alphaconc}{\alpha}     % Posterior parameter of Concrete distribution
\newcommand{\lambdaconc}{\lambda_c}   % Temperature parameter of concrete distribution 
 
\newcommand{\Uniform}{\text{Uniform}}  
\newcommand{\Normal}{\mathcal{N}}  
\newcommand{\Bernoulli}{\text{Bern}}

\newcommand{\GP}{\mathcal{GP}}
\newcommand{\Concrete}{\text{Concrete}}

\newcommand{\KL}{\text{KL}}
\newcommand{\expectation}[2]{\mathbb{E}_{#1} \left[#2\right] }
\newcommand{\covariance}[1]{\mathbb{C}\text{ov}[#1]}
\newcommand{\kernelt}{\kappa}

\newcommand{\elbo}{\mathcal{L}_{\text{elbo}}}
\newcommand{\ellterm}{\mathcal{L}_{\text{ell}}}
\newcommand{\klterm}{\mathcal{L}_{\text{kl}}}

\newcommand{\bY}{\mathbf{Y}}

\newcommand{\by}{\mathbf{y}}

\newcommand{\bt}{\mathbf{t}}
\newcommand{\bA}{\mathbf{A}}
\newcommand{\bW}{\mathbf{W}}

\newcommand{\cD}{\mathcal{D}}
\newcommand{\btheta}{\boldsymbol{\theta}}

\newcommand{\bF}{\mathbf{f}}
\newcommand{\bz}{\mathbf{z}}
\newcommand{\K}{\mathbf{K}}
\newcommand{\bigO}{\mathcal{O}}

\renewcommand{\det}[1]{|#1|}

\newcommand{\latnet}{\textsc{latnet}}
\newcommand{\pc}{\textsc{pc}}
\newcommand{\cpc}{\textsc{cpc}}
\newcommand{\iamb}{\textsc{iamb}}
\newcommand{\lingam}{\textsc{l}i\textsc{ngam}}
\newcommand{\pwlingam}{\textsc{pw-lingam}}
\newcommand{\ges}{\textsc{ges}}
\newcommand{\gds}{\textsc{gds}}
\newcommand{\resit}{\textsc{resit}}

\newcommand{\corrected}{\textsc{corrected}}
\newcommand{\uncorrected}{\textsc{uncorrected}}

\newenvironment{myproof}{{\bf Proof:} }{{ \cqfd }}

\def\cqfd{\hfill\hbox{$\hbox{\vrule width 0.8pt
			\vbox to6pt{\hrule depth 0.8pt width 5.2pt
				\vfill\hrule depth 0.8pt}\vrule width 0.8pt}$}} 

\newtheorem{theorem}{Theorem}

\newcommand{\ve}[1]{\textbf{#1}}
\newcommand{\ra}{\mathsf{A}}
\newcommand{\rp}{\uprho}
\newcommand{\rw}{\mathsf{W}}
\newcommand{\rx}{\mathsf{X}}
\newcommand{\rn}{\mathsf{N}}
\newcommand{\rb}{\mathsf{B}}
\newcommand{\expect}{\mathbb{E}}
\newcommand{\pr}{\mathbb{P}}
\newtheorem{lemma}[theorem]{Lemma}
\newtheorem{corollary}[theorem]{Corollary}

\newcommand{\defeq}{\stackrel{\text{\tiny def}}{=}}

\newcommand{\highlight}[1]{\textbf{#1}}

\newcommand{\brain}{\textsc{brain}\xspace}
\newcommand{\sydney}{\textsc{sydney}\xspace}
\newcommand{\yeast}{\textsc{yeast}\xspace}

\begin{document}
\pagenumbering{Alph}
\thispagestyle{empty}
\maketitle
\pagenumbering{arabic}

\begin{abstract}
We propose a network structure discovery model for continuous observations that generalizes linear causal models by incorporating a Gaussian process (GP) prior on a network-independent component, and random sparsity and weight matrices as the network-dependent parameters. This approach provides flexible modeling of network-independent trends in the observations as well as uncertainty quantification around the discovered network structure. We establish a connection between our model and multi-task GPs and develop an efficient stochastic variational inference algorithm for it. Furthermore, we formally show that our approach is numerically stable and in fact numerically easy to carry out almost everywhere on the support of the random variables involved. Finally, we evaluate our model on three applications, showing that it outperforms previous approaches. We provide a qualitative and quantitative analysis of the structures discovered for domains such as the study of the full genome regulation of the yeast \textit{Saccharomyces cerevisiae}.   
\end{abstract}

%\newpage

\section{Introduction}
\label{sec:intro}

% WHAT PROBLEM WE ARE ADDRESSING AND WHY IT IS IMPORTANT -> NETWORK DISCOVERY, MANY APPLICATIONS
Networks represent the elements of a system and their interconnectedness as a set of \textit{nodes} and \textit{arcs} (connections)  between them. Applications of network analysis range from biological systems such as gene regulatory networks and brain connectivity networks, to social networks and interactions between financial indices. Another application is modeling the relationship between property prices in different suburbs of a city, where each suburb is a node in the network and the property prices over time are the observations. In many such applications the structure of the network is unobserved and we wish to discover this structure from measurements \cite{Linderman2014}. 
% An example is discovering brain connectivity, in which the measurements (obtained by brain functional imaging) correspond to activities of different brain regions (nodes) over time, and the aim is to infer which regions are connected to each other \cite{Smith2011}. 

% ASSUMPTIONS AND PREVIOUS APPROACHES
% In line with the above examples, we assume that continuous-valued observations from different nodes of a network over time are given, and the aim is to discover the structure of the network. 
When dealing with continuous observations, a commonly used framework for this purpose is linear causal models \citep{bollen1989structural,Pearl:2000:CMR:331969,spirtes2000causation}, in which the data-generation process is defined such that the observations from each node are a linear sum of the observations from other nodes and additive noise. Such methods then use techniques such as independent component analysis \citep[e.g.][]{spirtes2000causation} to recover the dependencies between the nodes. 

An assumption in these models is that temporal variations in the observations from a node are either associated to the other nodes in the network, or to the changes in latent confounders; i.e., in the absence of any change in these two components, observations from a node are assumed to follow the noise distribution. However, one can assume that observations from a node can also follow a network-independent trend; for example property prices in a certain region can follow a decreasing/increasing trend over time,  independent of other regions.

%(1) From modelling perspective [already in intro]
% latnet is genereal (no assumption of graph) 
%(a) network-independent component 
%(b) Random W,A components as in Linderman and Adams 
\highlight{Main contribution}. 
In this paper we propose a network structure discovery model that generalizes linear causal models in two directions. Firstly, it incorporates a network-independent component for each node, which is determined by a Gaussian process (GP) prior capturing the inter-dependencies between observations over time. Consequently, the output of a node is now given by a sum of the network-independent component and a (noisy) linear combination of the observations from the other nodes. Secondly, it considers the parameters of this linear combination, which ultimately determine the structure of the network, as random variables. These parameters are given by a binary adjacency matrix and a continuous weight matrix \citep[similar in spirit to the work by][]{Linderman2014}, which allow for representing the sparsity and the strength of the connections in the network. 

The practical advantage of this modeling approach is twofold. Firstly, because of the non-parametric nature of the Gaussian process prior, it provides a more flexible data-generation process, which also allows for network-independent trends in the observations. Secondly, by considering the network-independent component and the network-structure parameters as random variables, it enables the incorporation of probabilistic prior knowledge; a fully Bayesian treatment of the variables of interest; and uncertainty quantification around the discovered network structure.
%
% Alternative
% In this paper we propose a network-structure discovery model that generalizes linear causal models by incorporating a Gaussian process (GP) prior on a network-independent component, and random sparsity and weight matrices as the network-dependent parameters. This approach provides a flexible modeling of  network-independent trends in the observations and uncertainty quantification around the discovered network structure. Furthermore, it enables a full Bayesian treatment of the variables of interest. 
 
\highlight{Inference}. 
In terms of inference in our model we show that, by marginalizing the latent functions corresponding to the network-independent components, our approach is closely related to multi-task GP models under a product covariance \citep{bonilla-et-al-nips-08,rakitsch2013all}. In particular, when conditioning on the network-dependent parameters, our model is a multi-task GP with a task-covariance constrained by the network parameters. This connection allows us to exploit properties of  Kronecker products in order to compute the marginal likelihood (conditioned on the network parameters) efficiently. We estimate the posterior over the network-dependent parameters building upon recent breakthroughs in variational inference \citep{rezende2014stochastic,kingma2013auto,Maddison2016},  making our framework amenable to large-scale stochastic optimization. 
% Furthermore, as we require a posterior over discrete random variables, we use the recently-developed Concrete distribution \citep{Maddison2016} for the application of stochastic back-propagation on these variables. 
 
\highlight{Theoretical analysis}.
%(4) Theory
%Identifiably and stability analysis.
% sampling may have little chance of local non-zero measure falling under machine zero
We investigate the numerical stability of our approach theoretically and discuss practical impacts. In particular, we show that all critical quantities of interests (i) can theoretically be sampled without assumptions, and (ii) can practically be computed ``easily" almost everywhere in their respective supports. In doing so, we show that our approach makes somewhat weaker assumptions than previous work \cite{Linderman2014}.

\highlight{Results}. 
%
%(3) Practical / Results
% Results are amazing, with outstanding quantitative performance and also providing qualitative analysis / understanding 
%  
% (a)FMRI: in the controlled situations that the underlying networks is not arbitrary, e.g., links are unidirectional and there is no cycle in the network, \latnet\ is competitive with more specialised methods designed for discovering DAGs.
% (b) -> more reallistic setting of general neworks: Sydney prices: is indeed more regionally localized in comparison to the networks discovered by other methods. Better than other methods
We investigate problems of discovering brain functional connectivity (\brain), modeling property prices in Sydney (\sydney), and understanding regulation in the yeast genome (\yeast). We provide a qualitative analysis and a quantitative evaluation of our approach showing that in controlled scenarios such as \brain, i.e.~when the underlying network is constrained by a directed acyclic graph, our approach tends to outperform previous methods specifically designed for these settings. In more general settings of unconstrained networks such as \sydney, we outperform previous work and show that our results are more realistic in discovering spatially-constrained trends. Finally, investigating the full yeast genome regulation (\yeast), we find that even in a large network (up to 38,000,000+ arcs), our technique is able to recover both high-level and low-level prior knowledge and hints on original findings.

The rest of this paper is organized as follows:  \S \ref{sec-mod-gp} states our model specifications, \S \ref{sec:marginal} presents the marginal likelihood given network parameters, \S \ref{sec:variational} details variational inference in our model. \S \ref{sec-num-sta} states our theory related to numerical stability, and \S \ref{sec-rel-wor} discusses related works. Finally, \S \ref{sec:experiments} presents our experiments, and a last section discusses and concludes. An appendix, (starting page \pageref{proof_proofs}) details all proofs and full experiments.  

\section{Model Specification}\label{sec-mod-gp}
%We have a network consisting of $N$ nodes, and we have observed outputs $\by_i$ from each node $i$ at time points $\bt_i$. Therefore, we have a data set $\cD$ consisting of $N$ vector observations $\bY = \{ \by_i\}_{i=1}^N$ and their corresponding observation times $\{ \bt_i\}_{i=1}^N$. The output of node $i$ at time $t$ is denoted by $y_i(t)$, and we assume it is generated by a noise-free latent function $f_i$ as follows:
Given a dataset $\cD$ of vector-valued observations $\bY = \{ \by_i\}_{i=1}^N$ and their corresponding times $\{ \bt_i\}_{i=1}^N$ from $N$ nodes in a network, our goal is to infer the existence and strength of the arcs between the nodes. To this end, let $y_i(t)$ be the output of node $i$ at time $t$, 
\begin{equation}
\label{eq:likelihood}
y_i(t) = f_i(t) + \epsilony, \quad \epsilony \sim \Normal(0, \varnoise) \text{,}
\end{equation}
where $\varnoise$ is the observation-noise variance. 
 To model latent function $f_i$, we assume that it is generated by two sources: (i) a network-independent component, which is denoted by $z_i(t)$, and (ii) a network-dependent component, i.e., a weighted sum of the inputs received from the rest of the network:
\begin{align}
\label{eq:prior-f-1}
f_i(t) &= z_i(t) + \sum_{\substack{j=1\\j \neq i}}^NA_{ij}W_{ij}\left[f_j(t) + \epsilonf\right] \text{, } \\
\label{eq:prior-f-2}
z_i(t) &\sim \GP(\mathbf{0}, \kernelt(t,t';\btheta)) \text{,} \quad \epsilonf \sim \Normal(0, \varf)  \text{,} 
\end{align}   
%\noteEB{Need to be consistent with matrix entry notation. In following sections we use boldface.}
%In other words, the output of node $i$ is generated by $z_i(t)$ plus the noisy outputs of all the other nodes.
 %Here, matrix $\bA$ is the adjacency matrix, where $A_{ij} \in \{0,1\}$ represents the existence of a connection (arc) from node $j$ to node $i$. Connections between nodes are weighted by matrix $\bW$, where $W_{ij}\in \mathbb{R}$ determines the weight of the connection from node $j$ to node $i$ (we assume $A_{ii}=W_{ii}=0$). We refer to $\bA$ and $\bW$ as network parameters. It is further assumed that output $f_j(t)$ is not received by node $i$ undistorted, but a zero mean Gaussian noise $\epsilonf \sim \Normal(0, \varf)$ adds to the inputs received by a node from other nodes. 
 %
 %As for the network-independent component, we assume $z_i(t)$ is distributed as a Gaussian process \citep[GP;][]{Rasmussen:2005}  with covariance function $\kernelt(t,t';\btheta)$ and  hyperparameters $\btheta$. The model is therefore a combination of parametric elements ($\bA$ and $\bW$) and non-parametric elements, i.e., $z_i(t)$, and therefore it is a semi-parametric model. \noteAD{noting that $n_i$ is the number of observation in each node.}
 %
where  $A_{ij} \in \{0,1\}$ represents the existence of an arc from node $j$ to node $i$ and $W_{ij}\in \mathbb{R}$ determines the weight of the connection from node $j$ to node $i$ (assuming $A_{ii}=W_{ii}=0$). These are elements of the adjacency matrix $\bA$ and weight matrix $\bW$, respectively, which we will refer to as network parameters. The network-independent component $z_i(t)$ is drawn from a Gaussian process \citep[GP;][]{Rasmussen:2005}   with covariance function $\kernelt(t,t';\btheta)$ and  hyperparameters $\btheta$. Since  $z_i(t)$ is non-parametric and $\bA, \bW$ are parametric components, we refer to the model above as a semi-parametric model.
\subsection{Prior over Network Parameters \label{sec:prior}}
\Eq \eqref{eq:likelihood} defines the likelihood of our observations and \eqs \eqref{eq:prior-f-1} and
\eqref{eq:prior-f-2} define the prior over the latent functions given the network parameters $\bA, \bW$. As our goal 
is to infer the structure of the network, these parameters are also random variables and their prior is defined as:
\begin{align}
\label{eq:prior-AW-1}
	p(\bA, \bW) = p(\bA) p(\bW) = \prod_{ij} p(A_{ij}) p(W_{ij}) \text{,} \\
\label{eq:prior-AW-2}
	p(A_{ij}) = \Bernoulli(\meana) \text{,} \quad p(W_{ij}) = \Normal(0, \varw) \text{,}
\end{align}
 where $\Bernoulli(\meana)$ denotes a Bernoulli distribution with parameter $\meana$.
\subsection{Inference Task}
% The order of ideas
% we marginalize f, and have a marginal conditioned on W, A
% show taht we can compute this efficiently
% Highlight relation to multitask learning 
% Now, A, W go through complex nonlinearities --> intractable posterior
% need variational inference 
% after this, we would like to apply the re-parameterization trick but cannot do it 
% P(A) is a discrete distribution 
% We use the trick of Madison, concrete distribution and do infernece on the relaxed stochastic graph
Our main inference task is to estimate the posterior over the network parameters $p(\bA, \bW | \cD)$. To this end, by exploiting the closeness of GPs under linear operators, we will first show in \S \ref{sec:marginal}  the exact expression for the (conditional) marginal likelihood 
$p(\bY | \bA, \bW)$ obtained when marginalizing latent functions $\bF$ (eq. (\ref{eq:Ft}) below). Furthermore, by establishing a relationship of our model to multi-task learning  \citep{rakitsch2013all,bonilla-et-al-nips-08}, we show how to compute this marginal likelihood efficiently.
Subsequently, due to the highly nonlinear dependence of $p(\bY | \bA, \bW)$ on $\bA, \bW$, we will approximate the posterior over these network parameters using variational inference in \S \ref{sec:variational}. 
% In order to carry out optimization of the variational objective efficiently using Monte Carlo (MC) estimates along with automatic differentiation, we provide a continuous relaxation of our model using the recently developed Concrete distribution \citep{Maddison2016}.

\section{Marginal Likelihood Given Network Parameters}
\label{sec:marginal}
Let us denote the values of all latent functions $f_i(t)$ at time $t$ with $\bF(t) = [f_1(t),\dots, f_N(t)]$, and similarly $\bz(t) = [z_1(t),\dots, z_N(t)]$. Hence, we can rewrite \eq \eqref{eq:prior-f-1} as:
\begin{equation}	
\label{eq:Ft}	
\bF(t)= (\I -\bA \odot \bW)^{-1}(\bz(t) + \bA \odot \bW\bepsilonf),
\end{equation}
where $\odot$ is Hadamard product. We refer the model in \eq \eqref{eq:Ft} as the \emph{inverse model}. 
%
% Given $\bz(t)$ and $\bepsilonf$, the random variables $f_i(t)$ are well defined if matrix $(\I -\bA \odot \bW)$ is invertible under the model's sampling conditions for $\A, \W$. We put this under a more general hat of numerical stability that we discuss below.
% \noteRN{Amir, make sure paramater fixing, which was previously discussed here, appears in experiments}
%
%
%In order to integrate out all the network-independent parameters it is easier 
%to work with the inverse model in eq. \eqref{eq:Ft}. Indeed,
Using this inverse model, we can see now that, for fixed $\A, \W$, since all the distributions are Gaussians and we are only applying linear operators,  the resulting distribution over $f_i$, and consequently over $y_i$, is also a Gaussian process. Hence, we only need to figure out the mean function and the covariance function of the resulting process. Below we present the main results and leave the details of the derivations to the \supplement.

Let $\B \defeq \A \hada \W$ and define the following intermediate matrices (which are a
function of the network parameters):
%\noteRN{I have simplified notations: no need for G in the main file}
%
\begin{eqnarray}
 \E & = & (\I -  \B)^{-1}  \B \B^T (\I -  \B)^{-T}   \label{eqEE} \\
\Kf & =  & {(\I -  \B)^{-1} } (\I -  \B)^{-T} \:\:.\label{eqFF}
\end{eqnarray}
Then we have that the mean function and covariance function of  latent process $f_i$  are  
 given by:
\begin{align}
	\mu_i(t) &= \expectation{}{f_i(t)} = 0 \text{,}\\ 
	\covariance{f_i(t), f_j(t')} &=  \matentry{\Kf}{i}{j} \kernelt(t,t'; \btheta) 
	+ \matentry{\E}{i}{j} \varf  \text{,}
\end{align}
where $\matentry{\mat{M}}{i}{j}$ denotes the $i,j$ entry of matrix $\mat{M}$. 

Consequently, the distribution of the noisy process $y_i$ is also a Gaussian process and  
can be further understood by assuming synchronized observations, i.e.~that 
the observations for all nodes lie on a grid in time, $t=1, \ldots, T$.  
Let  $\Y$ be the $N \times T$ matrix of observations and define $\y = \vect(\Y)$, 
where $\vect(\cdot)$ takes the columns of the matrix argument and stacks them into a single vector.
Therefore, the log-marginal likelihood conditioned on the network
parameters is given by ($\kron$ is Kronecker product):
\begin{align}
\label{eq:margl-aw1}
	\log p(\y | \A, \W) &=  - \frac{1}{2} \log \det{\Covy} - \frac{1}{2} \y^T \Covy^{-1} \y + C \text{,} \\
\label{eq:margl-aw2}
\text{with } \quad \Covy  &=  \Kf \kron \Kt + (\varf \E + \varnoise \I)  \kron \I \text{,} 
\end{align}
where $C= - 0.5 * n \log (2\pi)$; $\Kt$ is the $T \times T$ covariance matrix induced by evaluating 
the covariance function $\kernelt(t,t';\btheta)$ at all observed times;  
$\E$ and $\Kf$ are defined as in \eqs \ref{eqEE}, \ref{eqFF};
and $n = N \times T$ is the total number observations.
%\noteRN{Is it really +C (detail)?}
%
%
\subsection{Relationship with Multi-task Learning}  
\label{sec:mt}
Remarkably, the marginal likelihood of the model described in \eqs \eqref{eq:margl-aw1} and 
\eqref{eq:margl-aw2} reveals an interesting relationship with multi-task learning when using Gaussian process 
priors. Indeed, it boils down to the marginal likelihood of multi-task GP models under 
a product covariance \citep{bonilla-et-al-nips-08,rakitsch2013all}. 

In our case, the nodes in the network can be seen as the tasks in a multi-task GP model and are associated with a 
task-dependent covariance $\Kf$, which is fully determined by the parameters of the network $\A, \W$.  This contrasts 
with  multi-task models where $\Kf$ is, in general, a free parameter \citep{bonilla-et-al-nips-08}. 
Similarly, the input covariance $\Kt$ is the covariance of the observation times.  

Finally, conditioned on $\A, \W$, our model's marginal likelihood exhibits a more complex  noise covariance 
 $\varf \E + \varnoise \I$, which depends strongly on the network parameters. Such a 
covariance structured was not studied by \citet{bonilla-et-al-nips-08}, as they considered only diagonal noise-covariances.
However, \citet{rakitsch2013all} did consider the more general case of Gaussian systems with 
 a covariance given by the sum of two Kronecker products.  In the following section, we exploit 
 their results in order to compute, for fixed $\A, \W$,  the marginal likelihood of our model. 
%
%\noteRN{Pushed the Efficiency in a separate section with computational
%+ numerical properties}
% --> Getting the right organisation is tricky ..

\subsection{Computational efficiency}
In this section we show an efficient expression for the computation of the log-marginal likelihood in \eq \eqref{eq:margl-aw1}. For simplicity, we consider the synchronized case where all the $N$ nodes in the network have $T$ observations at the same times and, as before, we denote the total number of observations with $n = N \times T$. 
The main difficulties of computing the log-marginal likelihood above are the calculation of the log-determinant of an $n$ dimensional matrix, as well as solving an $n$-dimensional system of linear equations.
Our goal is to show that we never need to solve these operations on an $n$-dimensional matrix, 
which are $\bigO(n^3)$ but instead use $\bigO(N^3 + T^3)$ operations. The results in this section have been previously shown by  \citet{rakitsch2013all} for covariances with a sum of two Kronecker products. 

We show our derivations in the \supplement and present the results specific to our model here. To give some intuition behind such derivations, the main idea is to ``factor-out" the noise matrix $\varf \E + \varnoise \I$  from the covariance matrix $\Covy$ and then apply properties of the Kronecker product. 
Hence, given the following matrix definitions along with their eigen-decompositions:
\begin{eqnarray*}
\Covnoise & \defeq & (\varf \E + \varnoise \I) = \Qnoise \Lambdanoise \Qnoise^T \text{,} \\
\Ktildef & \defeq & \Lambdanoise^{-1/2} \Qnoise^T \Kf \Qnoise
\Lambdanoise^{-1/2} =   \Qtildef \Lambdatildef \Qtildef^T \text{;} 
\end{eqnarray*}
the log-determinant term in \eq \eqref{eq:margl-aw1} is given by 
\begin{equation}
 \log \det{\Covy}  = T \sum_{i=1}^N \log \lambdanoise^{(i)}  +  \sum_{i=1}^N \sum_{j=1}^T\log ( \lambdatildef^{(i)} \lambdatildet^{(j)} + 1 )
\end{equation}	
and the quadratic term can be computed as:
\begin{equation}
 \y^T \Covy^{-1} \y = \trace(\Ytilde^T \Qtildet  \Ytildetf \Qtildef^T ) \text{,}
\end{equation}
where 
$\matentry{\Ytildetf}{i}{j} =  \matentry{\Qtildet^T \Ytilde \Qtildef)}{i}{j} /  \matentry{\blambdatildet \blambdatildef^T + 1}{i}{j}$,
$ \Ytilde =  \Y \Qnoise \Lambdanoise ^{-1/2}$ and $\Qtildet
\Lambdatildet \Qtildet^T$ is the eigen-decomposition of $\Kt$.

We see that the above computations only require the eigen-decomposition of 
the $N \times N$ matrix  $\Ktildef$ and the $T \times T$ matrix $\Kt$, while avoiding 
matrix operations on the whole $n \times n$ matrix of
covariances $\Covy$. 
%
%\noteRN{Edwin, I have simplified notations here.}
% Note sure what is different but thanks

\section{Variational Inference}
\label{sec:variational}
Having marginalized the latent functions $\f$ corresponding to the network-independent component, our next step is to  use variational inference to approximate the true posterior $p(\A, \W | \cD)$ with a tractable family of distributions $q(\A,\W)$ that factorizes as
\begin{equation}
\label{eq:qAW}
q(\A,\W) = q(\A) q(\W) = \prod_{i,j} q(A_{ij}) q(W_{ij}) \text{,}
\end{equation}
where $i,j=1\dots N$, and $i \neq j$. Following the variational-inference desiderata we aim to optimize the variational objective, so-called evidence lower-bound ($\elbo$), which is given by:
\begin{align}
\label{eq:elbo}
\elbo &\defeq \klterm + \ellterm \text{, where} \\
\label{eq:kl}
\klterm &=  -  \KL(q(\A, \W)||p(\A, \W)) \text{ and } \\
\label{eq:ell}
\ellterm &= \mathbb{E}_{q(\A,\W)} [\log p(\bY|\bA,\bW)] \text{,}
\end{align}  
where $\KL(q||p)$ denotes the Kullback-Leibler divergence between distributions $q$ and $p$, and  $p(\A, \W)$ is the prior over the network-dependent parameters as defined in \eqs \ref{eq:prior-AW-1} and \ref{eq:prior-AW-2}. 

Given a specification of the approximate posteriors $\{q(A_{ij}), q(W_{ij}) \}$, our goal is to maximize $\elbo$ wrt their corresponding parameters. While computing $\klterm$ and its gradients is straightforward, we note that $\ellterm$ requires expectations of the log conditional likelihood, which depends on $\A, \W$ in a highly-nonlinear fashion. Fortunately, we can address this issue by exploiting recent advances in variational inference with regards to large-scale optimization of stochastic computation graphs \citep{rezende2014stochastic,kingma2013auto,Maddison2016,jang2016categorical}. 
\subsection{The Reparameterization Trick}
The main challenge of dealing with $\ellterm$ in the optimization of the variational objective is that of devising low-variance unbiased estimates of its gradients using, for example, Monte Carlo sampling. This can be overcome by re-parametrizing $\ellterm$ as a deterministic function of its parameters and a fixed-noise distribution. Such an approach has come to be known as the reparametrization trick \citep{kingma2013auto} or stochastic back-propagation \cite{rezende2014stochastic}. Hence we can define our approximate posterior over $W_{ij}$ as:
\begin{equation}
	q(W_{ij}) = \Normal(\posmeanw_{ij}, \posstdw^2_{ij}) \text{,}
\end{equation}
which can be reparametrized easily as a function of a standard normally-distributed variable $z_{ij} \sim \Normal(0,1)$, i.e.~$W_{ij} = \posmeanw_{ij} + \posstdw_{ij} z_{ij}$. 
%
%\noteAD{noting that we are using monte carlo samples here for approximating expected log-likelihood}
\subsubsection{The Concrete Distribution}
In order to define our approximate posteriors over $A_{ij}$ we face the additional challenge that the reparametrization trick cannot be applied to discrete distributions \cite{kingma2013auto}. We address this problem by using a continuous relaxation of discrete random variables known as the Concrete distribution \citep{Maddison2016}. We note that, contemporary to the Concrete distribution, a similar approach has  been proposed  by \citet{jang2016categorical} and it is also known as the Gumble-Softmax trick. 

The main idea of this trick is to replace the discrete random variable with their continuous relaxation, which is simply obtained by taking the softmax of logits perturbed by additive noise. Interestingly, in the zero-temperature limit the Concrete distribution corresponds to its discrete counterpart. More importantly, this continuous relaxation has a closed-form density and a simple reparameterization. For our purposes, we focus on using the Concrete distribution corresponding to the Bernoulli case for $A_{ij}$, i.e.~$q(A_{ij}) = \Concrete(\alphaconc_{ij}, \lambdaconc)$, and show how we sample from it using its reparameterization:
\newcommand{\cU}{\mathcal{U}}
\begin{align}
	 \cU & \sim \Uniform(0,1) \text{,}\\
	a_{ij} &= ( \log \alphaconc_{ij} + \log \cU - \log(1-\cU) )/\lambdaconc \text{,} \\
	A_{ij} &= 1/(1 + \exp(- a_{ij})) \text{,}
\end{align}
where $\alphaconc_{ij}$ are variational parameters and $\lambdaconc$ is a constant. 
\subsubsection{Preservation of the Variational Lower-Bound}
%\noteEB{Explain re-defien prior and show log probability}
As pointed out by \citet{Maddison2016}, optimization of $\elbo$ now implies replacing all discrete variables with their Concrete versions using the relaxation described above. This means that we also relax our priors $p(A_{ij})$ using the same procedure. Furthermore, in order to preserve the evidence-lower-bound nature of the variational objective, we  need to match the log-probabilities in  $\klterm$ with their sampling distribution. For the approximate posterior these are given by:
\begin{align}
\nonumber
	\log q (A_{ij}) &= \log \lambdaconc - \lambdaconc a_{ij} + \log \alphaconc_{ij} \\
	\label{eq:logprob-A}
		&	- 2 \log (1 + \exp(-\lambdaconc a_{ij} + \log \alphaconc_{ij})) \text{,}
\end{align}
and similarly for $p(A_{ij})$. 

Having relaxed our discrete variables, we proceed with optimization of the $\elbo$ in \eq \eqref{eq:elbo} by using Monte Carlo samples from $q(\W,\A)$ to estimate $\ellterm$. For computing $\KL(q(\A) || p(\A))$ we use samples from  $q(A), p(A)$ and their log-probabilities as defined in \eq \eqref{eq:logprob-A}. 
Finally, for $\KL(q(\W) || p(\W))$ we use the analytical form for the KL-divergence between two Gaussians.

\section{Numerical Stability}\label{sec-num-sta}
% EB: I think I've done this
%\noteRN{(Re)define parameters $p,\mu,\sigma$}
%
%\noteRN{Provide the definition of concrete distribution parameters
%  before}

Because $\klterm$ is straightforward to compute, we
investigate the numerical impact of computing  $\ellterm$. We first show that we
get samplability % (Subsection \ref{sec:prior})
``for free" --- \textit{i.e.}, compared to other sophisticated methods whose formal operating regime calls for additional 
 assumptions
\cite{Hyvarinen2013,Linderman2014,sishkwhbDA}.
% Commented out by EB -> not sure what this is about
% there is less formal obstruction to solving eq. \ref{eq:Ft} through sampling $\A$ and $\W$.
%\noteEB{Are we sure we are talking about \emph{identifiability} here?}

\begin{theorem}\label{idenfiabilityAW}
For any parameterization of the concrete distributions ($\lambdaconc \geq
0$ and $\alphaconc_{ij} \geq 0$ ($\forall i\neq j$)), $\I -\bA \odot \bW$
is non-singular with probability one.
\end{theorem}
\begin{theorem}\label{finiteE}
For any parameterization of the concrete distributions ($\lambdaconc \geq 0$ and $\alphaconc_{ij} \geq 0$ ($\forall i\neq j$))
and any $\varnoise>0$, $|\log p(\y | \W,\A)|\ll \infty$ with
probabiliy one for {any} $\y$.
\end{theorem}
(Proofs in \supplement, \S \ref{proof_thm_idenfiabilityAW}, \S \ref{proof_thm_finiteE}.) Therefore, it also holds that $|\ellterm|$ is
finite. 
The proofs build upon a trivial but key feature of concrete distributions: 
they can be designed so as not to break absolute
continuity of their input densities. 

Under specific assumptions, reminiscent to those of \citet{Hyvarinen2013,Linderman2014},
we show that $|\log p(\y | \W,\A)|$ can be sandwiched in a precise interval which makes its
computation numerically easy almost everywhere on its support,
\textit{i.e.}~sampling may have little chance of local non-zero measure falling
under machine zero (with the potential trouble when inverting relevant matrices). 
Most importantly, this holds \textit{for a sampling model (M) which is more
  general than ours}, meaning that one could make different choices
from the concrete distributions we use and yet keep the \textit{same} property:

\begin{itemize}
\item [($M$)] ($\forall i,j$) (i) weight $W_{ij}$ is picked as $\Normal(\mu_{ij}, \sigma^2_{ij})$
($\mu_{ij} \in \mathbb{R}, \sigma_{ij} > 0$), and
(ii) adjacency $A_{ij}$ is picked as $\Bernoulli(\rp_{ij})$ with
$\rp_{ij} \sim \mathcal{V}$, where $\mathcal{V}$ is any random
variable with support in $[0,1]$ (letting $p_{ij}
\defeq \expect[\rp_{ij}]$).
\end{itemize}
We define the total (squared)
expected input (resp. output) to node $i$ as $\mu^+_{i.} \defeq
\sum_j \mu_{ij}^2$ (resp. $\mu^+_{.i} \defeq
\sum_j \mu_{ji}^2$), and the total input (resp. output) variance as ${\sigma}^+_{i.} \defeq
\sum_j \sigma_{ij}^2$ (resp. ${\sigma}^+_{.i} \defeq
\sum_j \sigma_{ji}^2$). We also define averages, $\overline{\mu}^+_{i.} \defeq
\mu^+_{i.} / N$, $\overline{\sigma}^+_{i.} \defeq
\sigma^+_{i.} / N$ (same for outputs), and biased weighted
proportions, $\tilde{p}^\mu_{i.} \defeq  \sum_j
p_{ij}\mu_{ij}^2/\mu^+_{i.}$, $\tilde{p}^\sigma_{i.} \defeq  \sum_j
p_{ij}\sigma_{ij}^2/\sigma^+_{i.}$ (again, same for outputs).
Finally, we define two functions $U, S : \{1, 2, ..., 2N\} \rightarrow
\mathbb{R}_+$ as:
\begin{eqnarray*}
U(i) & \defeq & \left\{
\begin{array}{ll}
2 \tilde{p}^\mu_{i.}\overline{\mu}^+_{i.} + 2
\tilde{p}^\sigma_{i.}\overline{\sigma}^+_{i.} & (i \leq N)\\
2 \tilde{p}^\mu_{.j}\overline{\mu}^+_{.j} +
2 \tilde{p}^\sigma_{.j}\overline{\sigma}^+_{.j}: j\defeq i-N & (i>N)
\end{array}\right. \:\:,\\
S(i) & \defeq & \left\{
\begin{array}{ll}
    \overline{\mu}^+_{i.} +
    \overline{\sigma}^+_{i.}& (i \leq N)\\
\overline{\mu}^+_{.j} +
    \overline{\sigma}^+_{.j}: j\defeq i-N & (i>N)
\end{array}\right. \:\:.
\end{eqnarray*}
For any
diagonalizable $\mat{U}$, $\lambda(\mat{U})$ denotes its
eigenspectrum, and $\lambda^\uparrow(\mat{U}) \defeq
\max|\lambda(\mat{U})|$, $\lambda^\downarrow(\mat{U}) \defeq
\min|\lambda(\mat{U})|$.

\begin{theorem}\label{thEASYM}
Fix any constants $c>0$ and $0<\upgamma<1$ and let $\lambda_\circ
\defeq (\lambda^{\downarrow}(\Kt)/2)  + \varnoise$ and $\lambda_\bullet \defeq 2\lambda^{\uparrow}(\Kt) + \varf + \varnoise$.
Under sampling model $M$, suppose that
\begin{eqnarray}
\max_i U(i) & \in &
\left[\frac{\max_i S(i)}{N^{\gamma}} 
 , \frac{1}{100 N^2}\right]\:\:.\label{intUS2}
\end{eqnarray}
If $N$ is larger
than some constant depending on $c$ and $\gamma$, then with probability $\geq 1 - (1/N^c)$ over the sampling of $\W$ and
$\A$, we have that $\lambda(\Covy) \subset \left[\lambda_\circ, \lambda_\bullet\right]$.
\end{theorem}
(Proof in \supplement, \S  \ref{proof_thm_thEASYM}.) To be
non empty, the
interval puts the implicit constraint that $\max_i S(i) =
O(1/N^{\zeta})$ for some constant $\zeta$,
\textit{i.e.}~roughly, the expected square signal (node-wise) has to be
bounded. Such a bound in the signal's statistics or values is an assumption
that can be found in \citet{Hyvarinen2013,Linderman2014}.
As a corollary,
if the kernel ($\lambda^{\downarrow}(\Kt)$) or noise parameters ($\varnoise$) are, say, above machine zero, there is reduced risk for numerical
instabilities in sampling $\log p(\y | \W, \A)$.
\begin{corollary}\label{corEASYM}
Define $g(z, \y) \defeq (N/2)\log z + z \|\y\|_2^2 - C$,
where $C$ is defined as in (\ref{eq:margl-aw1}). Then under the
settings of Theorem \ref{thEASYM}, with probability $\geq 1 - 1/N^c$,
we have:
\begin{eqnarray}
-\log p(\y | \W, \A) & \in & [g(\lambda_\circ, \y), g(\lambda_\bullet,
\y)]\:\:, \forall \y\:\:.\nonumber
\end{eqnarray}
\end{corollary}
As discussed in the \supplement, the constraint of
(\ref{intUS2}) can be weakened for specific
$\mathcal{V}$s (\textit{e.g.} for more ``informative''
distributions). We also remark that we do not face the
sparsity constraints of the model of \citet{Linderman2014}, such as
the mandatory
increase of sparsity with $N$.

\section{Related Work}\label{sec-rel-wor}
Linear causal models with Gaussian noise \citep[e.g.][]{bollen1989structural,Pearl:2000:CMR:331969} are  different from ours in three key aspects: (i) they assume that the underlying network is a directed acyclic graph \citep[e.g.][]{spirtes2000causation}; (ii) they do not represent the connection strengths using random matrices; and (iii) they do not incorporate the network-independent Gaussian process component. 
Unlike our work, other approaches assume a non-Gaussian additive noise \citep{shimizu2006linear} or a nonlinear transformation of the network-dependent component \citep{hoyer2009nonlinear}. 

As observations in our model are generated from several latent Gaussian processes, our framework is related to GP latent variable models \citep{lawrence2005probabilistic,zhang_6629}. However, our goal is to recover the underlying network structure, instead of carrying out dimensionality reduction or predicting observations for the nodes. On a different vein, our use of random matrices representing network structure is similar to the model in \citet{Linderman2014}, but that model is focused on point-process data rather than continuous-valued observations. 
Finally, with regards to multi-task GP models \cite{bonilla-et-al-nips-08,rakitsch2013all} and more general frameworks for modeling vector-valued outputs \citep{wilson2010generalised}, other approaches have considered Bayesian inference in multi-task learning subject to specific constraints, such as rank constrains \citep{koyejo-2013}. However, their work is mostly focused on dealing with the problem of large-dimensionality data instead of network discovery.

\section{Experiments}
\label{sec:experiments}
We evaluate our approach on three distinct domains: discovering brain
functional connectivity (\brain), modeling property prices in Sydney
(\sydney) and regulation in the yeast
genome (\yeast). We used the squared exponential covariance function and optimized variational parameters, hyperparameters, and likelihood parameters in an iterative fashion using Adam \cite{Kingma2014}. For details of prior setting and optimization specifics see the \supplement. 
%  
%\brain), modelling property prices in Sydney (\sydney), and understanding regulation in the yeast genome (\yeast
\subsection{Methods compared}
% There is a large variety of baseline methods that the current method could be compared against. 
We considered the methods used in \citet{peters2014causal} as baselines for comparison. 
These  include: (1) \pc\ algorithm \citep{spirtes2000causation}; 
(2) Conservative \pc\ algorithm denoted by \cpc\ \citep{Ramse2006}; and 
(3) \lingam\ \citep{shimizu2006linear}; 
In addition to the above, we  considered two more algorithms: 
(4) \iamb\ \cite{tsamardinos2003algorithms},  
and 
(5) 
Pairwise LiNGAM (\pwlingam), which has been recently developed for discovering connectivity between different brain regions \cite{Hyvarinen2013}. 
For the reasons detailed in the \supplement other methods used in \citet{peters2014causal} were not applicable to the datasets analyzed here.

\subsection{\brain~domain}
\label{sec:fmri}

\begin{figure}[t]
	\centering
	\includegraphics[scale=1]{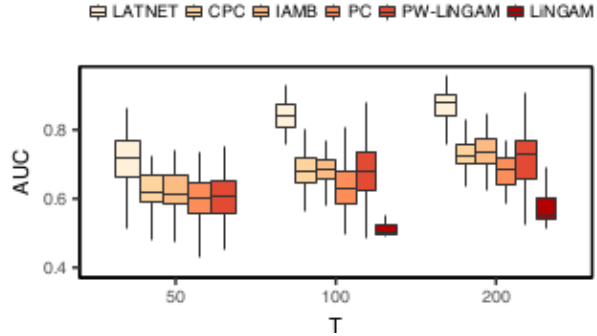}
	\caption{AUC obtained on the \brain~data ($N=15$) for performing link prediction (connectivity between brain regions) for different number of observations from each node ($T$).}
\label{fig:fmri}
\end{figure}

The aim is to discover the connectivity between different brain regions is crucial in neuroscience studies. We analyze the benchmarks of \citet{Smith2011}, in which the activity of different brain regions is recorded at different time points and the aim is to find which brain regions are connected to each other. Each benchmark consists of data from 50 subjects, each consisting of 200 time points ($T=200$). We used network sizes $N=5, 10, 15$, and the true underlying connectivity network is a directed acyclic graph (DAG). Networks discovered by \latnet\ are not restricted to DAGs, and therefore baseline methods assuming the underlying network is a DAG have a \textit{de facto} favorable bias. 

Results are evaluated using the area under the ROC curve (AUC) and shown in Figure~\ref{fig:fmri} for $N=15$ using box-plots (top and bottom edges of the box correspond to the first and third quartiles respectively). The \supplement (\S \ref{secsup-fmri}) presents details on how the AUC is computed for the different methods and the results for $N=5, 10$. We see that, although other methods are favorably biased about the underlying structure, \latnet\  provides significantly better performance than such baselines. We note that \lingam\ was unable to perform inference for $T$ small and its output is reported only for $T \ge 100$. We also note that \pc, \cpc\ and \iamb\ may generate non-concave ROCs. Curves can be post-processed for concave envelopes which improves the AUC, but this artificial post-processing that equivalently mixes outputs does not guarantee the existence of parameters that will in effect produce networks with the corresponding performances. This is discussed in the \supplement \S \ref{secsup-fmri}, along with the results with concavification.

\label{sec:prices}
\begin{figure*}[t]
	\centering
		\includegraphics[scale=1]{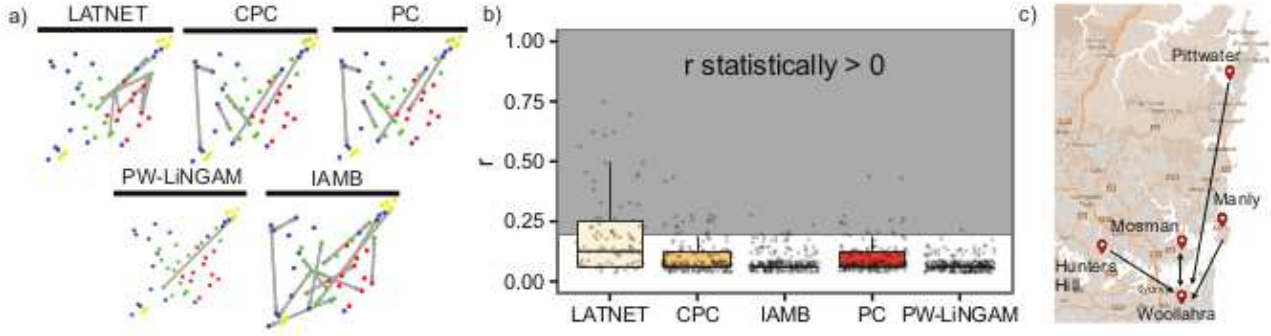} 
	\caption{(a) Networks discovered by the different methods. Points in the graph are located according to their geographical coordinates. red: inner ring suburbs; green: middle ring suburbs; blue: outer ring suburbs; yellow: greater metropolitan region. (b) Distribution of $r$ values for each method. $r$ refers to the proportion of the networks in which a discovered arc presents. (c) Arcs with highest $r$ values discovered by \latnet.}.
	\label{fig:housing}

\end{figure*}
\subsection{\sydney~domain}

The aim is to discover the relationship between property prices in different suburbs of Sydney. The data includes quarterly median sale prices for 51 suburbs in Sydney and surrounding area from 1995 to 2014. 
%Discovering the network structure in this situation involves a relatively large network ($N=51$) in comparison to the number of observations.
We kept the analysis window to five years ($T=20$ since data is quarterly) and starting from 1995--1999 the window is shifted by one year each time until 2010-2014. Some major patterns are expected, like the presence of hubs or authorities related to mass transfers between suburbs.

Figure~\ref{fig:housing}(a) shows the inferred arcs for years 2010-2014, where the nodes are positioned according to their geographical locations. See the \supplement for the full map of the discovered arcs for all years
 and for details of the thresholds used for finding significant arcs. 
We see that the network discovered by \latnet\ is more regionally localized compared to the other methods, and displays major authorities. Note that \lingam\ was unable to perform inference. To complete with a quantitative analysis,
we computed for each algorithm and for each pair of nodes, the proportion $r$ of networks in which an arc was discovered. Then, we computed a distribution of the $r$ values in the networks, emphasizing the ``stable'' arcs with $r$ statistically $>0$ (risk $\alpha = 0.05$). 
Figure~\ref{fig:housing}(b) shows the $r$ values for the different methods.  Less than $8\%$, $5\%$ of \pc\ and \cpc\ arcs were significant, respectively, while more than 29$\%$ of \latnet~arcs are significant.
Interestingly, \iamb\ and \pwlingam\ did not find significant arcs.

What turns out to be interesting is the \textit{actual} arcs found for those in the 29$\%$ that represent the highest $r$-values. Figure~\ref{fig:housing}(c) shows the top-6 of these arcs inferred by \latnet. They clearly indicate that one area of Sydney, Woollahra, acts like an authority in the network, since it receives lots of arcs from other major areas (Hunters Hill, Manly, Mosman, Pittwater). These areas all share common features: they are in central-north Sydney, all have coastal areas, and they happen to be well-known prestigious areas with the highest median property price in  Sydney \cite{pprice_bworld}, so the observed percolation is no surprise.

\subsection{\yeast~domain}
\label{sec:gene}

\begin{figure*}[t]
\centering
\scalebox{0.925}{
\begin{tabular}{m{.33\linewidth}m{.36\linewidth}m{.36\linewidth}}
\includegraphics[trim=240bp 110bp 180bp 100bp,clip,width=\linewidth]{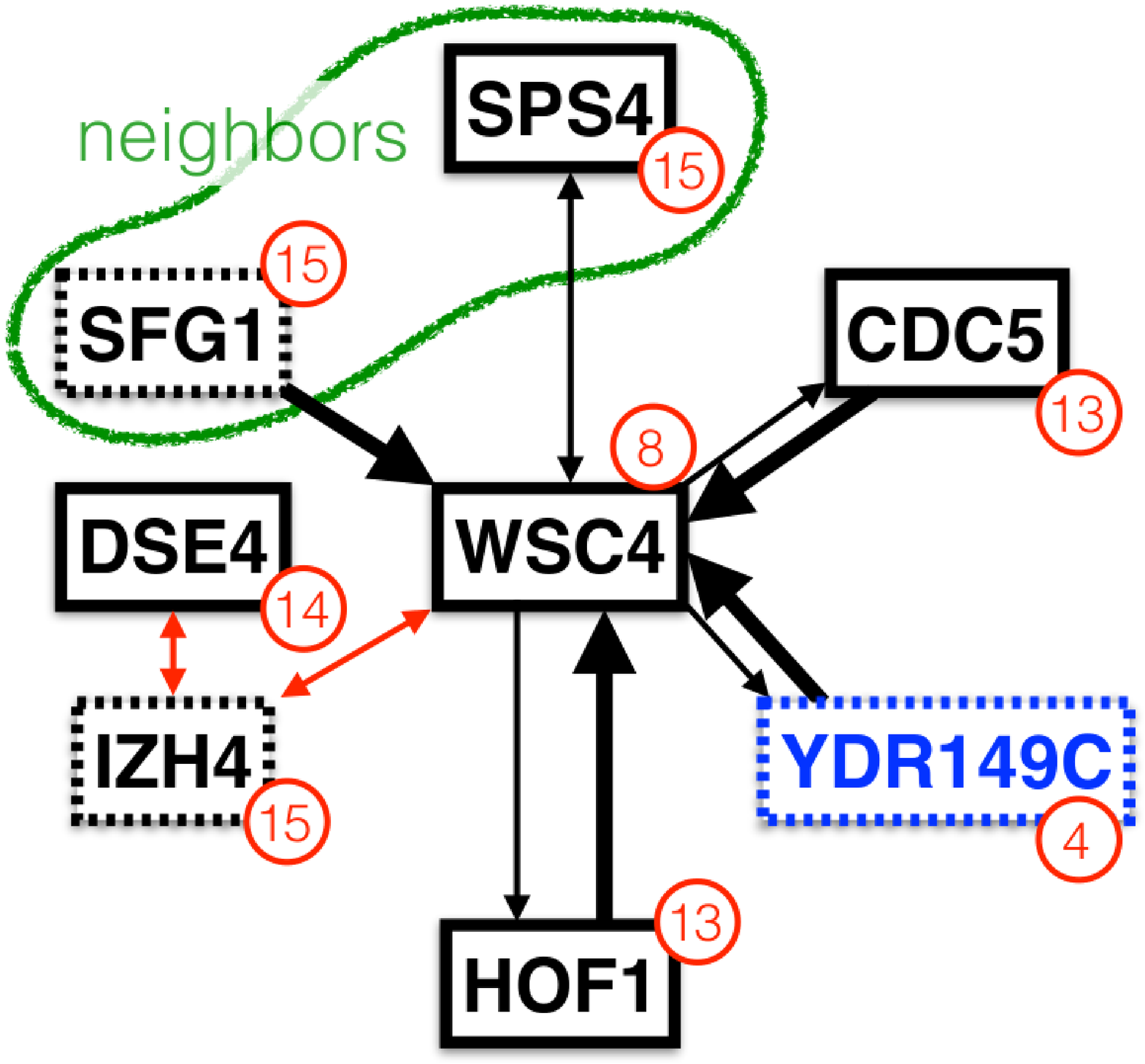} 
&\hspace{-0.7cm}
\includegraphics[trim=120bp 0bp 130bp 25bp,clip,width=\linewidth]{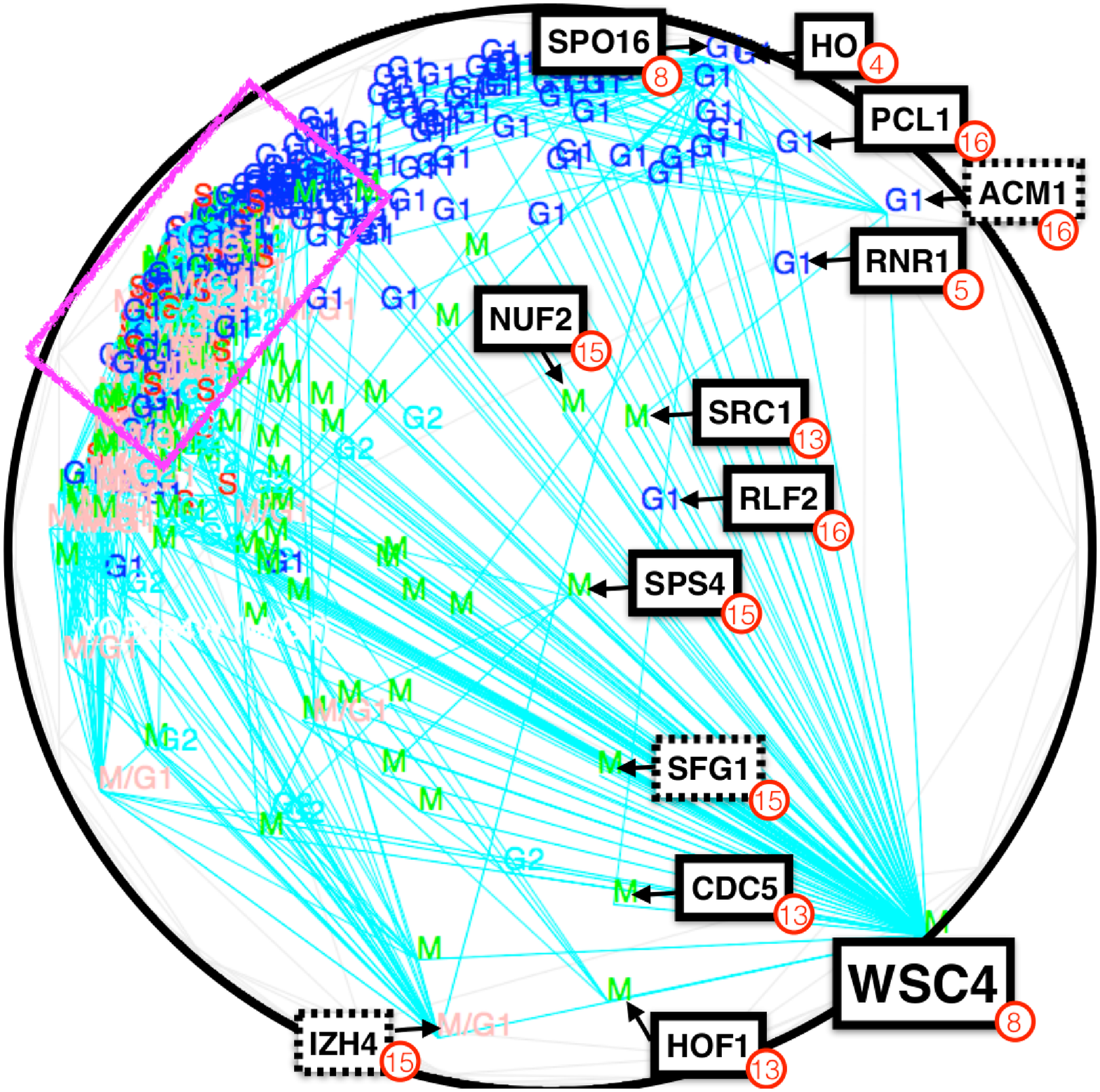} 
&\hspace{-0.7cm}
\includegraphics[trim=250bp 110bp 180bp 100bp,clip,width=\linewidth]{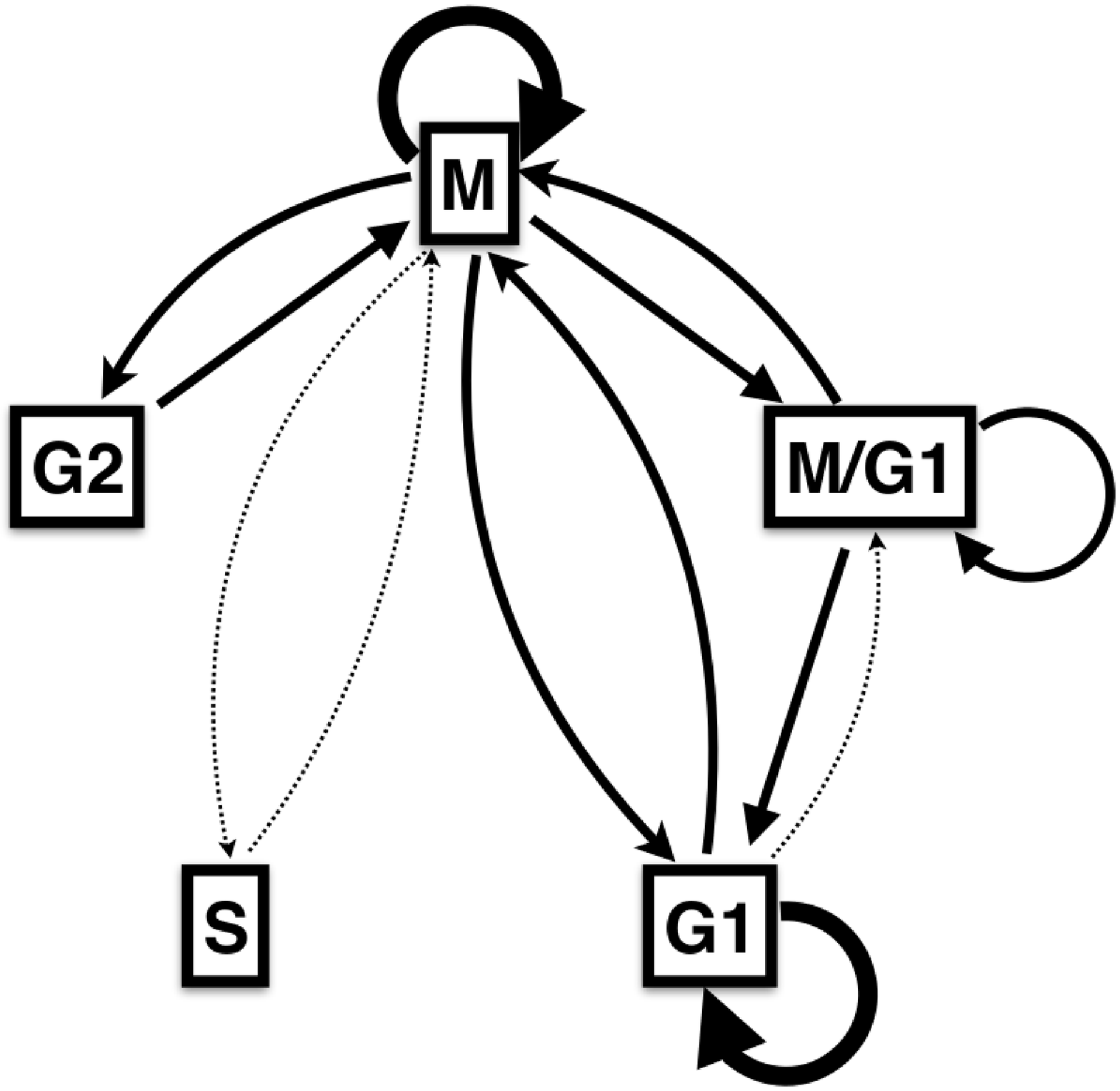} 
\end{tabular}
}
\caption{ (best viewed in color) \textit{Left}: subgraph $G_w$ containing \textit{all} strong arcs with $p_{ij}>0.62$; plain rectangles (vs dashed): reported cell cycle transcriptionally regulated genes \cite{rkt+HR}; thick arcs (resp. thin arcs): $p>0.65$ (resp. $p\in (0.62, 0.65]$); black arcs (resp red arcs): $\mu > 0$ (resp. $\mu < 0$); red disk: chromosome number; in blue: gene with no known biological process/function/compartment; \textit{Center}: manifold learned from strong arcs, displayed in Klein disk (conformal). Strong arcs displayed in blue segments; gene names shown for the most important ones on manifold (conventions follow left pic); pink rectangle: area with comparatively few strong arcs; \textit{Right}: network aggregating strong arcs discovered for the YCC, between cell cycle phases (see text).\label{fig:Klein}}
\end{figure*}

The aim is to infer genome regulation patterns for one extensively studied species, \citep[\textit{Saccharomyces cerevisiae},][]{ssz+CI}'s. In numbers, this represents 100,000+ data points and a network with up to 38,000,000+ arcs. Biology tells us that this network \textit{is} directed. The complete set of experiments and results is available in the \supplement, \S \ref{sec-spel}. In addition to the \textit{full} genome analysis, we have performed a finer grained analysis on a $\sim$tenth known to be involved in a heavily regulated and key part of the yeast's life, the yeast cell cycle (YCC). These genes are the so-called sentinels of the YCC \cite{ssz+CI}. In each experiments, we have scrutinized the subset of \textit{strong} arcs, for which both $p_{ij}$ ($\A$) \emph{and} $|\mu_{ij}|$ ($\W$) are large, that is, $p_{ij}$ in top-$0.1\%$, $|\mu_{ij}|$ in top-$1\%$. These arcs happen to be indeed very significant, with $\alpha$ risk $\approx 10^{-4}$ for the rejection of hypothesis that $p_{ij}$ is not larger than random existence ($0.5$). The discovered networks' information is far beyond the scope of this paper, but some striking points can be noted, taking as references the cell cycle transcriptionally regulated genes \citep{rkt+HR} and \textit{http://www.yeastgenome.org} as a more general resource. 

\noindent \textbf{Analysis of the YCC}. Figure \ref{fig:Klein} summarizes qualitatively the results obtained by \latnet. The topmost strong arcs belong to a small connected component ($G_w$) organized around gene WSC4, asymmetric (both in terms of arcs and $p$ values) and with apparent patterns of positive / negative regulation (sign of $\mu$). The key genes are involved in the cell structure dynamics (DSE4, WSC4, CDC5, HOF1). The IZH4 arcs with negative $\mu$ are perhaps not surprising: WSC4 and DSE4 are involved into cell walls (integrity for the former, \textit{degradation} for the latter). IZH4 is a man-in-the-middle: it has activity \textit{elevated} in zinc \textit{deficient} cells, and it turns out that zinc is crucial for hundreds of enzymes, \textit{e.g.}~for protein folding. Strikingly, SPS4 and SFG1 happen to be \textit{neighbors} on the same chromosome. Figure \ref{fig:Klein} (center) shows the broad picture, with a display of the manifold coordinates induced by the network's graph, built upon \citet{msLS}. It displays that a small number of key genes drive the coordinates. Most of these genes have been pinned down as cell cycle transcriptionally regulated \citep{rkt+HR}, and they essentially turn out to be heavily involved in both a/sexual reproduction.
Last, Figure \ref{fig:Klein} (right) summarizes the broad picture of strong arcs between YCC phases: it should come at no surprise that cell splitting, (M)itosis, has the largest number of these arcs.

\noindent \textbf{Analysis of the full genome} (results in \supplement, \ref{sec-yeast}). The following patterns emerge: first, the network is highly asymmetric: more than twice strong arcs go outside the YCC compared to arcs coming in the YCC from non-YCC genes. Second, the leading YCC genes are still major genes, but they tend to be outnumbered by genes that are perhaps more ``all-purpose''. Last, the predominance of gap phase G1 compared to G2 is in fact a known feature of \textit{Saccharomyces cerevisiae} (compared to other yeast species such as \textit{Saccharomyces pombe}).
\section{Conclusion \& Discussion}\label{sec-dis}

We introduced a framework for network structure discovery when continuous-valued observations from the nodes are given,  which can be seen as a generalization of linear causal models. We have established an interesting connection of the model with multi-task learning and developed an efficient variational inference method for estimating the posterior distribution over the network structure. We have demonstrated the benefits of our approach on real applications by providing qualitative and quantitative analyses.
Besides computational efficiency, our theoretical analysis shows that the traditional constraints for \textit{numerical stability} and \textit{identifiability} in networks are alleviated. Indeed, from a theoretical standpoint, the state of the art goes with substantial constraints that can be related to the fact that the true model exists, is unique and can be properly recovered. Such constraints go with an abstraction of the dynamical model that looks like a series of the type $\sum_{t\geq 0} z^t$, where $t$ symbolizes discrete time and $z$ can be a real, complex or matrix argument \cite{Hyvarinen2013,Linderman2014}. The trick of replacing the analysis overall $t$s by one over a window, rather than alleviating the constraint, substitutes it for other stationarity constraints that can be equally restrictive \cite{shimizu2006linear}. 
In our case, such constraints do not appear because time is absorbed in a GP. The finiteness of the evidence lower-bound (ELBO) is essentially obtained ``for free". What we get with additional assumptions that parallel traditional ones is a non-negligible uplift in the easiness of the expected log-likelihood part, the bottleneck of the ELBO. This result also holds for a broader class of posteriors than the ones we use, opening interesting avenues of applications for concrete distributions.

Finally, experiments display that \latnet~is able to perform sound inference already on small domains, and scales to large domains with the ability to pinpoint meaningful local properties of the networks, as well as capture important high-level network features like global patterns of emigration between wealthy suburbs in Sydney or species characteristics for the yeast.

{
\bibliography{refs_gplatnet,BIB/bibgen,refs_variational}
}

\section{Proofs and algorithms}\label{proof_proofs}

\subsection{Proof of Theorem \ref{idenfiabilityAW}}\label{proof_thm_idenfiabilityAW}

\begin{figure}[t]
\centering
\scalebox{0.925}{
\begin{tabular}{c}
\includegraphics[trim=170bp 190bp 490bp 100bp,clip,width=.50\linewidth]{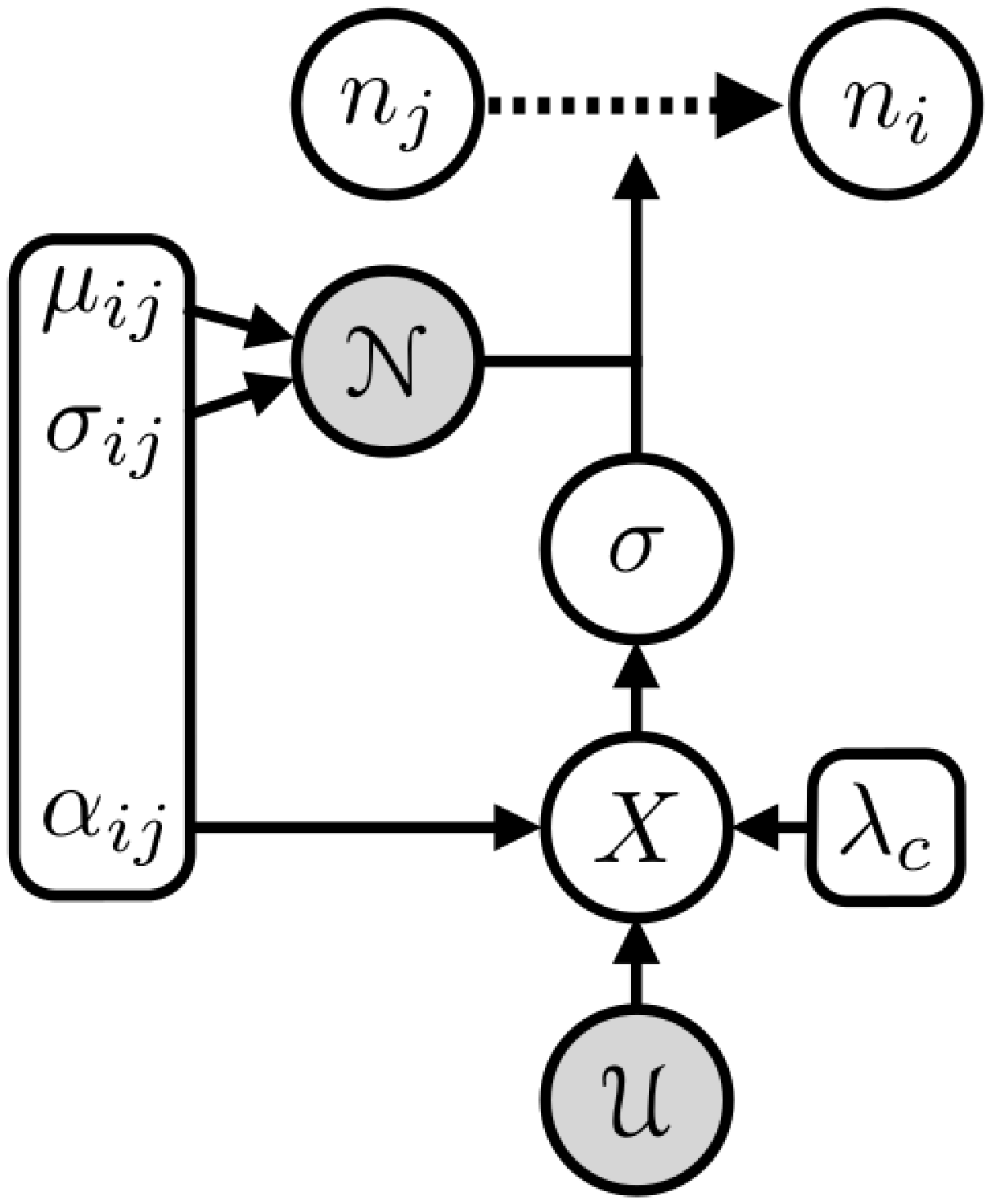} 
\end{tabular}
}
\caption{Sampling graph for arc going from node $n_j$ to node
  $n_i$ ($i\neq j$). Grey nodes denote random variables. Notations in part borrowed from \cite{Maddison2016}. $\lambda_c$ is a
  constant that does not depend on the arc.\label{fig:summ}}
\end{figure}

\begin{theorem}\label{th00}
For any $\lambda_c \geq 0$ and $\alpha_{ij} \geq 0$ ($\forall i\neq j$), $\I-\B$ is non-singular with probability
one.
\end{theorem}
\begin{myproof}
Denote for short $\mat{G} \defeq \I-\B$. The proof is split in three
cases, (I) $\lambda_c > 0$ and $\alpha_{ij} > 0, \forall i\neq j$, (II)
$\lambda_c = 0$ and $\alpha_{ij} > 0, \forall i\neq j$, and finally
(III) $\lambda_c = 0$ and $\exists i\neq j, \alpha_{ij} = 0$.\\
\noindent (Case I: $\lambda_c > 0$, $\alpha_{ij} > 0, \forall i\neq j$) The coordinates $g_{ij}$ take on constant values $g_{ii} = 1$ on the diagonal
($\forall i\in [N]$), and random values $G_{ij}$ outside the diagonal
($i\neq j$). The density of $G_{ij}$ equals $q(A_{ij})\cdot q(W_{ij})$, where $q(W_{ij})
\defeq \Normal(\mu_{ij}, \sigma_{ij}^2)$ and $q(A_{ij}) \defeq
\sigma_{\alpha_{ij}, \lambda_c}(U)$ with
\begin{eqnarray}
\sigma_{\alpha,\lambda_c}(U) & \defeq & \frac{1}{1+\exp\left(-\frac{\log \alpha +
      \log U - \log(1-U)}{\lambda_c}\right)}\:\:,
\end{eqnarray}
 and $U \sim \mathcal{U}(0,1)$ is uniform on interval $(0,1)$
\citep{Maddison2016}. The proof that $\mat{G}$ is invertible adapts a standard
argument (for example, \cite{tSA}). For any\footnote{Whenever $N=1$,
  $\mat{G}\defeq [1]$ is always invertible.} $N\geq 2$, denote $\ve{g}_1,\ve{g}_2, ...,
\ve{g}_N$ the columns of $\mat{G}$, that is, $\mat{G} = [\ve{g}_1 | \ve{g}_2| ...|
\ve{g}_N]$. Each of them can be thought of as a
random vector where one coordinate takes value 1 with probability 1,
an this coordinate is different for all vectors. $\mat{G}$ is non
invertible iff $\ve{g}_1,\ve{g}_2, ...,
\ve{g}_N$ is linearly dependent. Remark that \textit{none} of the
$\ve{g}_j$s can be the null vector, so if $\mat{G}$
is not invertible, then 
\begin{eqnarray}
\exists j>1 & : & \ve{g}_j \in \mathrm{span}(\ve{g}_1,\ve{g}_2, ...,
\ve{g}_{j-1})\:\:.
\end{eqnarray}
As a consequence,
\begin{eqnarray}
\Pr(\mathrm{det}(\mat{G}) = 0) & \leq & \sum_j \Pr(\ve{g}_j \in \mathrm{span}(\ve{g}_1,\ve{g}_2, ...,
\ve{g}_{j-1}))\:\:,\label{detG}
\end{eqnarray}
where the distribution is the product distribution over the columns of
$\mat{G}$. Fix \textit{any} $\ve{g}_1,\ve{g}_2, ...,
\ve{g}_{j-1}$ belonging to the respective supports of the columns, and let 
\begin{eqnarray}
q_j & \defeq & \Pr(\ve{g}_j \in \mathrm{span}(\ve{g}_1,\ve{g}_2, ...,
\ve{g}_{j-1}) | \ve{g}_1,\ve{g}_2, ...,
\ve{g}_{j-1})\:\:.
\end{eqnarray}
Because the uniform and normal distributions are both absolutely
continuous with respect to Lebesgue measure and
$\sigma_{\alpha,\lambda_c}(x)\leq 1 \ll \infty$ (it is also Lipschitz) for any $\alpha > 0, \lambda_c
\neq 0, U \in (0,1)$, so is the density of $G_{ij}$ for any $i\neq j$, and
thereby the density of $\ve{g}_j$ for any $j\geq 1$. Along with the
fact that $\mathrm{span}(\ve{g}_1,\ve{g}_2, ...,
\ve{g}_{j-1})$ has strictly positive codimension for any $j\leq N$, it comes
\begin{eqnarray}
q_j & = & 0, \forall j\geq 2, \forall\ve{g}_1,\ve{g}_2, ...,
\ve{g}_{j-1} \mbox{ fixed}\:\:.
\end{eqnarray}
Integrating over the choices of $\ve{g}_1,\ve{g}_2, ...,
\ve{g}_{j-1}$, we get $\Pr(\ve{g}_j \in \mathrm{span}(\ve{g}_1,\ve{g}_2, ...,
\ve{g}_{j-1})) = 0, \forall j\leq N$ and so $\Pr(\mathrm{det}(\mat{G}) = 0)
= 0$ from ineq. (\ref{detG}). As a consequence, $\I-\B$ is non-singular with probability
one, as claimed.\\
\noindent (Case II: $\lambda_c = 0$, $\alpha_{ij} > 0, \forall i\neq j$) this boils down to
choosing a Bernoulli $B(p_{ij})$
distribution over $A_{ij}$, corresponding to the limit case
$\lambda_c \rightarrow 0$ with \citep{Maddison2016}:
\begin{eqnarray}
p_{ij} & = & \frac{\alpha_{ij}}{1+\alpha_{ij}}\:\:.\label{defPij}
\end{eqnarray}
In this case, the distribution of $\ve{g}_j$ is not
absolutely continuous but a trick allows to truncate the distribution
on a subset over which it is absolutely continuous, and therefore
reduce to Case I to handle it.\\
The \textit{only} atom eventually having non-zero
probability is the canonical
basis vector $\ve{1}_j$, which has probability $\prod_{i\neq
  j}(1-p_{ij})$ to be sampled. We now perform a sequence of recursive row-column
(row followed by column or the reverse)
permutations, starting on $\mat{G}$, which by definition do not change its
invertibility status but only the sign of its determinant. The first
row-column permutation is carried out in such a way that the first
column of the new matrix, $\Pi_1(\mat{G})$, is the first canonical
basis vector, $\ve{1}_1$. We then repeat this operation to have the
second canonical basis vector in the second column, and so on until 
until it cannot
be done anymore to make appear on the left block a new canonical basis
vector. Assuming we have done $N-k$ sequences, we obtain from
$\mat{G}$ the final matrix $\Pi_1(\mat{G})$ with:
\begin{eqnarray}
\Pi_1(\mat{G}) & = & \left[
\begin{array}{ccc}
\mat{I}_{N-k} & |  & \mat{A}_{(N-k)\times k}\\
\mat{0}_{k\times (N-k)} & |  & \hat{\mat{G}}_{1,k}
\end{array}
\right]\:\:.\label{defGtilde}
\end{eqnarray}
Here, $\hat{\mat{G}}_{1,k} \in {\mathbb{R}}^{k\times k}$. 
Now, we are going to carry out $\Pi_1$ again, but on the lower-right block,
$\hat{\mat{G}}_{1,k}$. Removing dimension-dependent indexes, we obtain matrix
\begin{eqnarray}
\Pi_2(\mat{G}) & = & \left[
\begin{array}{ccc}
\mat{I} & |  & \mat{A}\\
\mat{0} & |  & \Pi_1(\hat{\mat{G}}_{1})
\end{array}
\right]\label{eq001}\\
 & = & \left[
\begin{array}{ccc}
\mat{I} & |  & \mat{A}_1\\
\mat{0} & |  & \left[
\begin{array}{ccc}
\mat{I} & |  & \mat{A}_2\\
\mat{0} & |  & \hat{\mat{G}}_{2}
\end{array}
\right]
\end{array}
\right]\:\:.\label{defGtilde2}
\end{eqnarray}
We then keep on doing the same transformation on block
$\hat{\mat{G}}_{2}$ until it is not possible anymore. When it is not
possible anymore, we know that the current submatrix, say
$\hat{\mat{G}}_{n}$, does not contain any canonical basis vector as
column, as depicted in Figure \ref{fig:mat}.
\begin{figure}[t]
\centering
\scalebox{0.925}{
\begin{tabular}{c}
\includegraphics[trim=200bp 220bp 510bp 200bp,clip,width=.50\linewidth]{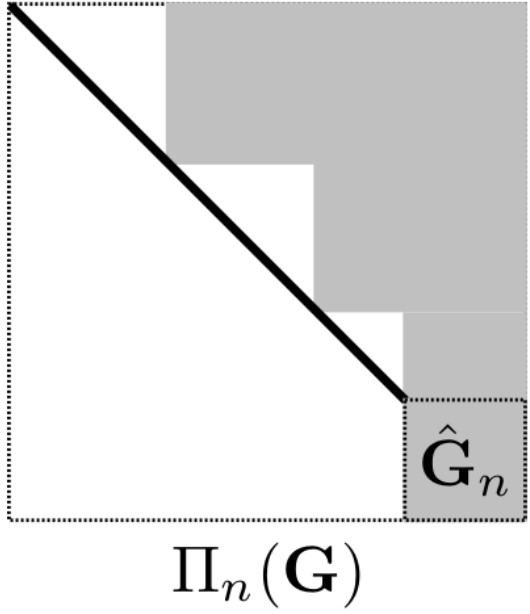} 
\end{tabular}
}
\caption{Final matrix $\Pi_n(\mat{G})$ obtained after recursively
  applying $\Pi_1(.)$ to the lower-right block. Here, white blocks
  mean all-zero, plain dark lines mean all-one, and grey is unspecified.\label{fig:mat}}
\end{figure}
\begin{lemma}\label{ldec}
$|\mathrm{det}(\mat{G})| =
|\mathrm{det}(\hat{\mat{G}}_n)|, \forall n\geq 1$.
\end{lemma}
\begin{myproof}
We proceed by induction. The key observation is the following
standard linear algebra identity. Denoting with a single index the order of a general square
matrix, like $\mat{A}_p$, we have for any $\mat{A}_p$ non-singular, 
\begin{eqnarray}
\left[
\begin{array}{ccc}
\mat{A}_p & |  & \mat{B}\\
\mat{C} & |  & \mat{D}_q
\end{array}
\right] \underbrace{\left[
\begin{array}{ccc}
\mat{I}_p & |  & -\mat{A}_p^{-1}\mat{B}\\
\mat{0} & |  & \mat{I}_q
\end{array}
\right]}_{\defeq \mat{E}} & = & \underbrace{\left[
\begin{array}{ccc}
\mat{I}_p & |  & \mat{0}\\
\mat{C}\mat{A}_p^{-1} & |  & \mat{I}_q
\end{array}
\right]}_{\defeq \mat{F}}  \left[
\begin{array}{ccc}
\mat{A}_p & |  & \mat{0}\\
\mat{0} & |  & \mat{D}-\mat{C}\mat{A}_p^{-1}\mat{B}
\end{array}
\right] \:\:,\label{idGen}
\end{eqnarray}
for any $p>0, q>0, p+q=n, \mat{B}\in {\mathbb{R}}^{p\times q},
\mat{C}\in {\mathbb{R}}^{q\times p}, \mat{D}\in {\mathbb{R}}^{q\times
  q}$. Taking determinants, we note that
$\mathrm{det}(\mat{E})=\mathrm{det}(\mat{F})=1$ because they are
triangular with unit diagonal, and so
\begin{eqnarray}
\mathrm{det}\left(\left[
\begin{array}{ccc}
\mat{A}_p & |  & \mat{B}\\
\mat{C} & |  & \mat{D}_q
\end{array}
\right] \right) & = & \mathrm{det}\left(\left[
\begin{array}{ccc}
\mat{A}_p & |  & \mat{0}\\
\mat{0} & |  & \mat{D}-\mat{C}\mat{A}_p^{-1}\mat{B}
\end{array}
\right]\right)\label{eq000}\\
 & = & \mathrm{det}(\mat{A}_p)\cdot \mathrm{det}(\mat{D}-\mat{C}\mat{A}_p^{-1}\mat{B})\:\:,
\end{eqnarray}
because the right hand-side in eq. (\ref{eq000}) is block
diagonal. Matching the left hand-side of eq. (\ref{eq000}) with
eq. (\ref{defGtilde}), so putting $\mat{A}_p = \mat{I}$ and $\mat{C} =
\mat{0}$, we obtain $\mathrm{det}(\Pi_1(\mat{G})) =
\mathrm{det}(\hat{\mat{G}}_1 -\mat{0}\mat{I}_{N-k}\mat{A}_p) =
\mathrm{det}(\hat{\mat{G}}_1)$, and therefore
$|\mathrm{det}(\mat{G})|=|\mathrm{det}(\hat{\mat{G}}_1)|$. We then
just recursively use eq. (\ref{idGen}) on the lower-right block
($\hat{\mat{G}}_j$, for $j=1, 2, ..., n-1$) and get the
statement of the Lemma.
(End of the proof of Lemma \ref{ldec}).\end{myproof}\\
So, $\mat{G}$ is invertible iff $\hat{\mat{G}}_n$ is invertible and:
\begin{eqnarray}
\Pr(\mathrm{det}(\mat{G}) = 0) & \leq &
\Pr(\mathrm{det}(\Pi_1(\mat{G})) = 0)\nonumber\\
 & & = \Pr(\mathrm{det}(\hat{\mat{G}}_1) = 0)\nonumber\\
 & \leq & \Pr(\exists k\in \{2, 3, ..., N\} : \mathrm{det}(\hat{\mat{G}}_{k})=0 |
 \hat{\mat{G}}_{k}\in {\mathbb{R}}^{k\times k} \wedge \mathcal{P}(\hat{\mat{G}}_{k}))\nonumber\\
 & \leq & \sum_{k=2}^N \Pr(\mathrm{det}(\hat{\mat{G}}_{k})=0 |
 \hat{\mat{G}}_{k}\in {\mathbb{R}}^{k\times k} \wedge \mathcal{P}(\hat{\mat{G}}_{k}))\label{eqb222}\:\:,
\end{eqnarray}
where $\mathcal{P}(\mat{G})$ is the property that no column of
$\mat{G}$ is a canonical basis vector. 
Notice the change: no column in $\hat{\mat{G}}_{k}$ is allowed
to be a canonical basis vector, and therefore the support for the
density of the columns of $\hat{\mat{G}}_{k}$ is such that its
distribution is now absolutely continuous. We are thus left with the
same case as in Case I, which yields $\Pr(\mathrm{det}(\hat{\mat{G}}_{k})=0 |
 \hat{\mat{G}}_{k}\in {\mathbb{R}}^{k\times k} \wedge \mathcal{P}(\hat{\mat{G}}_{k})) = 0, \forall k\in \{2, 3, ..., N\}$, and brings
$\Pr(\mathrm{det}(\mat{G}) = 0) = 0$ as well.\\
\noindent (Case III: $\lambda_c \geq 0$, $\alpha_{ij} = 0$ for some
$i\neq j$) Remark that $\lim_{\alpha\rightarrow
  0}\sigma_{\alpha,\lambda_c}(x) = 0$ if $\lambda_c > 0$, and if $\lambda_c
= 0$, 
this boils down from Case II (eq. (\ref{defPij})) to
choosing a Bernoulli $B(0)$
distribution over $A_{ij}$, so both cases coincide with $A_{ij}$ being
chosen as $B(0)$, implying $G_{ij} = 0$. We are left with the same
transformation as in Case II --- the main difference being that some
$G_{ij}$ is surely zero, but it changes nothing to the reasoning done
in case II. Therefore, $\Pr(\mathrm{det}(\mat{G}) = 0) = 0$ again.
\end{myproof}\\

\subsection{Proof of Theorem \ref{finiteE}}\label{proof_thm_finiteE}

It comes from Theorem \ref{idenfiabilityAW} that $\mat{G}^{-1}$ can always be computed with probability
one with respect to the random sampling of $\B$, and there is no
constraint on the parameterization of the concrete distribution for
invertibility \citep{Maddison2016}. Interestingly perhaps, the story would be
completely different for the invertibility of $\B$, as the argument
for cases (II) and (III) break down because with
positive probability that would be easy to lower-bound, $\B$ would in
fact be not invertible.\\

The
important consequence of Theorem \ref{th00} relies on the computation
of the log likelihood, which we recall:
\begin{eqnarray}
	\log p(\y | \W, \A) &= & - \frac{1}{2}
        \log \det{\Covy} - \frac{1}{2} \y^T \Covy^{-1} \y + C \:\:.\label{eqloglike}
\end{eqnarray}
We now prove Theorem \ref{finiteE}. We recall the main matrix component of
eq. (\ref{eqloglike}):
\begin{eqnarray}
		\Covy  &=  &((\I - \B)^\top(\I - \B))^{-1} \kron \Kt +
                \varf (\I - \B)^{-1} \B ((\I - \B)^{-1} \B )^\top
                \kron \Sigmai + \varnoise \I\:\:.
\end{eqnarray}
We observe that the following two matrices are positive semi-definite\footnote{As remarked
  above, depending on the choices of parameters $\lambda_c$ and $\alpha_{..}$, the null space of $\B$ is indeed not always reduced to the
  null vector. Therefore, $(\I - \B)^{-1} \B ((\I - \B)^{-1} \B
  )^\top$ may be not positive definite with strictly positive probability.}:
$(\I - \B)^{-1} \B ((\I - \B)^{-1} \B )^\top$, $\Kt$, while $\Sigmai,
\I, ((\I - \B)^\top(\I - \B))^{-1}$ are positive definite (with
probability 1 for that last one, see Theorem \ref{th00}). Hence, a
sufficient condition for the combination in $\Covy$ to be positive
definite is $\varnoise>0$, as claimed. This brings the finiteness of $|\log p(\y | \W,
\A)|$ with probability one, and the statement of Theorem \ref{finiteE}.

\subsection{Proof of Theorem \ref{thEASYM}}\label{proof_thm_thEASYM}

We split the proof in two main parts, the first of which focuses on a
simplified version of the model in which the Bernoulli parameter ($\A$) is
sampled according to a Dirac --- \textit{e.g.} in the context of
inference, from the prior standpoint, it is
maximally informed. The results might be useful outside our
framework, if $p$ is sampled from a distribution
different from the ones we use. 

We state the main notations involved
in the Theorem. We define the total (squared)
expected input (resp. output) to node $i$ as $\mu^+_{i.} \defeq
\sum_j \mu_{ij}^2$ (resp. $\mu^+_{.i} \defeq
\sum_j \mu_{ji}^2$), and the total input (resp. output) variance as ${\sigma}^+_{i.} \defeq
\sum_j \sigma_{ij}^2$ (resp. ${\sigma}^+_{.i} \defeq
\sum_j \sigma_{ji}^2$). We also define averages, $\overline{\mu}^+_{i.} \defeq
\mu^+_{i.} / N$, $\overline{\sigma}^+_{i.} \defeq
\sigma^+_{i.} / N$ (same for outputs), and biased weighted
proportions, $\tilde{p}^\mu_{i.} \defeq  \sum_j
p_{ij}\mu_{ij}^2/\mu^+_{i.}$, $\tilde{p}^\sigma_{i.} \defeq  \sum_j
p_{ij}\sigma_{ij}^2/\sigma^+_{i.}$ (again, same for outputs).

Now, we define two functions $U, E : \{1, 2, ..., 2N\} \rightarrow
\mathbb{R}_+$ as:
\begin{eqnarray*}
U(i) & \defeq & \left\{
\begin{array}{ll}
2 \tilde{p}^\mu_{i.}\overline{\mu}^+_{i.} + 2
\tilde{p}^\sigma_{i.}\overline{\sigma}^+_{i.} & (i \leq N)\\
2 \tilde{p}^\mu_{.j}\overline{\mu}^+_{.j} +
2 \tilde{p}^\sigma_{.j}\overline{\sigma}^+_{.j}: j\defeq i-N & (i>N)
\end{array}\right. \:\:,\\
E(i) & \defeq & \left\{
\begin{array}{ll}
\phi(\tilde{p}^\mu_{i.}) \cdot
    \overline{\mu}^+_{i.} +
    \overline{\sigma}^+_{i.}& (i \leq N)\\
\phi(\tilde{p}^\mu_{.j})\cdot \overline{\mu}^+_{.j} +
    \overline{\sigma}^+_{.j}: j\defeq i-N & (i>N)
\end{array}\right. \:\:,
\end{eqnarray*}
where $\phi(z) \defeq 2\sqrt{z(1-z)}$ is Matsushita's entropy. For any
diagonalizable matrix $\mat{M}$, we let $\lambda(\mat{M})$ denote its
eigenspectrum, and $\lambda^\uparrow(\mat{M}) \defeq
\max\lambda(\mat{M})$, $\lambda^\downarrow(\mat{M}) \defeq
\min\lambda(\mat{M})$. Our simplified version of Theorem
\ref{thEASYM}, which we first prove,
is the following one.

\begin{theorem}\label{thEASYM_DIRAC}
Assume $\ra_{ij} \sim \mathcal{B}(\rp_{ij})$ with $\rp_{ij} \sim
\mathrm{Dirac}(p_{ij})$, and $\rw_{ij} \sim \mathcal{N}(\mu_{ij},
\sigma^2_{ij})$, $p_{ij}, \mu_{ij}, \sigma_{ij}$ being fixed for any $i,j$. Fix any constants $c>0$ and $0<\upgamma<1$ and let
\begin{eqnarray}
\lambda_\circ \defeq \frac{\lambda^{\downarrow}(\Kt)}{2}
  + \varnoise \:\:, \:\: \lambda_\bullet \defeq 2\lambda^{\uparrow}(\Kt) + \varf + \varnoise\:\:.\label{defSISI}
\end{eqnarray}
Suppose that:
\begin{eqnarray}
\max_i U(i) & \in &
\left[\frac{\max_i E(i)}{N^{\gamma}} 
 , \frac{1}{100 N^2}\right]\:\:.\label{intUS2SI}
\end{eqnarray}
If $N$ is larger
than some constant depending on $c$ and $\gamma$, then with probability $\geq 1 - (1/N^c)$ over the sampling of $\W$ and
$\A$, the following holds true:
\begin{eqnarray}
\lambda(\Covy) & \subset & \left[\lambda_\circ, \lambda_\bullet\right]\:\:.
\end{eqnarray}
\end{theorem}

\subsubsection{Helper tail bounds and
properties for arcs, row and columns in matrix
$\A\odot \W$}\label{helperB}

To obtain concentration bounds on $\log p(\y | \W, \A)$, we need to
map the arc signal onto the real line, including \textit{e.g.} when
$p=0$ (in which case there cannot exist an arc between the two
corresponding nodes, so there is no observable "weight" \textit{per se}). We follow the convention for the Hawkes model of \cite{Linderman2014},
and associate to these "no signal" events the real zero, which makes
sense since for
example it matches the Dirac case when $\mu, \sigma \rightarrow 0$ ---
which corresponds to an arc with weight always zero ---.
Define for short
\begin{eqnarray}
\mat{H} & \defeq & (\I - \B)^\top(\I - \B)\label{defH}\:\:,\\
\mat{H}' & \defeq & (\I - \B)(\I - \B)^\top\label{defHprime}\:\:,\\
\mat{J} & \defeq & \B \B^\top\label{defJ}\:\:,
\end{eqnarray}
so that
\begin{eqnarray}
		\Covy  &=  &\mat{H}^{-1} \kron \Kt +
                \varf (\I - \B)^{-1} \mat{J} (\I - \B)^{-\top}
                \kron \Sigmai + \varnoise \I\:\:.\label{sumCOVY}
\end{eqnarray}
We remark that the eigenspectrum of $(\I - \B)^{-1} \mat{J} (\I -
\B)^{-\top}$ is the same as for $\mat{J} \mat{H}'^{-1}$: if $\ve{u}$ is an eigenvector of $(\I - \B)^{-1}
\mat{J} (\I - \B)^{-\top}$, then $(\I - \B)^{-1}
\mat{J} (\I - \B)^{-\top} \ve{u} = \lambda \ve{u}$ is equivalent to $
\mat{J} (\I - \B)^{-\top} \ve{u} = \lambda (\I - \B) \ve{u}$,
equivalent to $
\mat{J} (\I - \B)^{-\top} (\I - \B)^{-1}\ve{v} = \lambda \ve{v}$
(letting $\ve{v} \defeq (\I - \B) \ve{u}$), finally equivalent to
$\mat{J} \mat{H}'^{-1}\ve{v} = \lambda \ve{v}$. Therefore, bounding
the eigenspectra of $\mat{H}, \mat{H}', \mat{J}$, plus adequate
assumptions on that of $\Kt$, shall lead to
bounding the eigenspectra of $\Covy$, but to get al these bounds, we
essentially need properties and concentration inequalities for the
coordinates of $\B$ and their row- or column- sums. This is what we
establish in this Section.

We first derive a tail bound for arc weight, removing indexes for clarity, and assuming $q(W)
\defeq \Normal(\mu, \sigma^2)$ and $q(A) \defeq \mathcal{B}(p)$ (see
Figure \ref{fig:summ}). Let $\rw$ denote the random variable taking
the arc weight. We
recall that random variable $\rx$ is $(k,\upbeta)$-sub-Gaussian
($k, \upbeta>0$) iff \citep{vMA}:
\begin{eqnarray}
\expect_{\rx}[\exp(\uplambda (\rx - \expect[\rx] )) & \leq &
k\cdot \exp\left( \frac{\upbeta^2\uplambda^2}{2} \right)\:\:, \forall \uplambda \in \mathbb{R}\:\:.\label{subgdef}
\end{eqnarray}
\begin{theorem}\label{lemSG}
Let $\rw \sim q(W)\cdot q(A)$. The following holds true:
\begin{eqnarray}
\expect_{\rw}[\exp(\uplambda (\rw - \expect[\rw] )) & = & (1-p)\cdot \exp(- p\mu \uplambda) + p \cdot \exp\left(\mu (1-p)\uplambda + \frac{\sigma^2
     \uplambda^2}{2}\right)\:\:, \forall \uplambda \in \mathbb{R}\:\:. \label{tb1}
\end{eqnarray}
Furthermore, $\rw$ is $(1,\upbeta)$-sub-Gaussian with $\upbeta$ satisfying:
\begin{itemize}
\item $\upbeta^2 = p \sigma^2$ if $p\in \{0,1\}$,
\item $\upbeta^2 = 2\sqrt{p(1-p)}\mu^2 + \sigma^2$ if $p\in (0,1)$.
\end{itemize}
\end{theorem}
\begin{myproof}
Denote for short two random variables $\rn \sim \Normal(\mu,
\sigma^2)$ and $\rb \sim \mathcal{B}(p)$. We trivially have $\expect[\rw] = p\mu$ and:
\begin{eqnarray}
\expect_{\rw}[\exp(\uplambda (\rw - \expect[\rw] )) & = &
\expect_{\rw}[\exp((\rw - p\mu  ) \uplambda )] \nonumber\\
 & = &  (1-p)\cdot \expect_{\rn}[\exp(-p\mu\uplambda )] + p\cdot \expect_{\rn}[\exp((\rn - p\mu) \uplambda)] \nonumber\\
 & = &  (1-p)\cdot \exp(- p\mu \uplambda) + p\cdot \expect_{\rn}[\exp((\rn - p\mu) \uplambda)]\nonumber\\
 & = &  (1-p)\cdot \exp(-p\mu\uplambda ) + p \cdot \exp(-
 p\mu \uplambda) \cdot \expect_{\rn}[\exp(\rn \uplambda )]\nonumber\\
 & = &  (1-p)\cdot \exp(- p\mu \uplambda) + p \cdot \exp(-
 p\mu \uplambda) \cdot \exp\left(\mu \uplambda + \frac{\sigma^2
     \uplambda^2}{2}\right)\label{ndis}\\
 & = & (1-p)\cdot \exp(- p\mu \uplambda) + p \cdot \exp\left(\mu (1-p)\uplambda + \frac{\sigma^2
     \uplambda^2}{2}\right)\label{ndis2}\:\:,
\end{eqnarray}
for any $\uplambda \in \mathbb{R}$, as claimed for eq. (\ref{tb1}). Eq. (\ref{ndis}) comes from the moment generating function for
Gaussian $\rn$. Now, it is clear that 
\begin{itemize}
\item $\rw$ is sub-Gaussian with
parameter $\upbeta=\sigma$ in the
following two cases: (i) $p=1$, (ii) $\mu = 0$. For this latter case,
we have indeed $\expect_{\rw}[\exp(\uplambda (\rw - \expect[\rw] ))
= (1-p) + p \cdot \exp(\sigma^2
     \uplambda^2/2) \leq  ((1-p) + p) \cdot \exp(\sigma^2
     \uplambda^2/2) = \exp(\sigma^2
     \uplambda^2/2)$ (using Jensen's inequality on $z\mapsto \exp(z)$). Furthermore, sub-Gaussian parameter
   $\sigma$ cannot be improved in both cases.
\item the trivial case $p=0$ leads to sub-Gaussianity for any $\upbeta \geq
0$.
\end{itemize}
Otherwise (assuming thus $0<p< 1$ and $\mu \neq 0$), we can
immediately rule out the case $\upbeta \leq \sigma$ (for any $k>0$), by
noticing that, for $\upbeta = \sigma$, we have $p \cdot \exp(\mu
(1-p)\uplambda) = k$ for 
\begin{eqnarray}
\uplambda & = & \frac{1}{(1-p)\mu}\log \frac{k}{p} \:\: (\ll \infty)\:\:,
\end{eqnarray}
and so, for this value of $\uplambda$, $\expect_{\rw}[\exp(\uplambda (\rw - \expect[\rw] )) > p \cdot \exp\left(\mu
(1-p)\uplambda + (\upbeta^2
     \uplambda^2)/2\right) = k \exp(\upbeta^2
     \uplambda^2/2)$. In the following, we therefore consider $0<p<
     1$, $\mu \neq 0$ and $\upbeta > \sigma$. 
\begin{lemma}\label{lineq1}
$\forall p \in [0,1], \forall x>0$, we have
\begin{eqnarray}
p(x-1) + 1 & \leq & x^p\exp(\phi_u(p)\cdot \log^2 x)\:\:,\label{pfund}
\end{eqnarray}
where $\phi_u(p)\defeq \sqrt{p(1-p)}$ is (unnormalized) Matsushita's entropy.
\end{lemma}
\noindent \textbf{Remark}: ineq. (\ref{pfund}) is probably close to be
optimal analytically.
Replacing $\phi_u(p)$ by a dominated
entropy like Gini's $\phi_u(p)\propto p(1-p)$ (\textit{i.e.} with
finite derivatives on the right of 0 and left of 1) seems to break the result.\\
\begin{myproof}
The proof makes use of several tricks to counter the fact that
the right-hand side of ineq. (\ref{pfund}) is essentially concave --
but not always -- in $p$, and essentially convex -- but not always --
in $x$, and matches the left-hand side as $p\rightarrow \{0,1\}$. In a
first step, we show that ineq. (\ref{pfund}) holds for $\log x \in
[-1,1]$ (and any $p\in [0,1]$), then Step 2 shows that ineq. (\ref{pfund}) holds for $\log x
\geq -1$ (and $p\in [0,1]$). Step 3 uses a symmetry argument on the right-hand side of
ineq. (\ref{pfund}) to extend the result to any $x>0$ (and any $p\in
[0,1]$), thereby finishing the proof.\\

\noindent Step 1. We remark that $\phi_u(p) \defeq \sqrt{p(1-p)}$ satisfies the
following properties:
\begin{itemize}
\item [(i)] $\lim_0 \phi_u'(p) = + \infty$, $\lim_1 \phi_u'(p) = -\infty$;
\item [(ii)] $\lim_{\{0,1\}} \phi_u''(p)  +(1+\phi_u'(p) \cdot
k)^2 = - \infty$ for any $k$.
\end{itemize}
Denote for short $F(p,x) \defeq x^p\exp(\phi_u(p)\cdot \log^2 x)$. We have:
\begin{eqnarray}
\frac{\partial F}{\partial p} & = & \log x \cdot (1+ \phi_u'(p)\cdot \log x) \cdot F_x(p)\:\:,\\
\frac{\partial^2 F}{\partial p^2} & = & \log^2 x \cdot \left(\phi_u''(p)  +(1+\phi_u'(p)\log x)^2\right)\cdot F_x(p)\:\:.
\end{eqnarray}
It comes $\partial F/\partial p \sim_0 \phi_u'(p) \log^2 x \cdot F_x(p)$ and so
$\lim_0 \partial F/\partial p = + \infty$ because of (i). Since $F(0,x) = 1$, we have $F(p,x) >
p(x-1) + 1$ in a neighborhood of 0. Also, we
can check as well that $\lim_0 \partial^2 F/\partial p^2= -\infty$ because of (ii), so $F(p,x)$ is
concave in a neighborhood of $0$. For the same reasons, $F(p,x)$ is
concave in a neighborhood of $1$ and since $F(1,x) = x$, we also have $F(p,x) >
p(x-1) + 1$ in a neighborhood of 1. Now, to zero the second
derivative, we need equivalently:
\begin{eqnarray}
\log x & = &
\frac{1}{\phi_u'(p)}\cdot\left(\frac{\pm 1}{2\phi_u^{\frac{3}{2}}(p)}-1\right)\:\:,\label{eqlog1}
\end{eqnarray}
or, equivalently again:
\begin{eqnarray}
G(p,x) \defeq 2\phi_u^{\frac{3}{2}}(p)+\log x(1-2p)\phi_u^{\frac{1}{2}}(p) & = & r\:\:,
\end{eqnarray}
with $r\in \{-1,1\}$. We have (letting $z \defeq \log x$ for short and
$h_1(z)\defeq \sqrt{8z^2+9}, h_2(z) \defeq (2 z^2 + h_1(z) + 3)/(z^2+1)$),
\begin{eqnarray}
\max_{p\in [0,1]} G(p,x) & = & \frac{\left(h_2(z)\right)^{\frac{1}{4}}
  \left(\sqrt{h_2(z) (z^2+1)} +
\sqrt{3+4 z^2 -h_1(z)} z \right)}{2^{\frac{5}{4}} 3^{\frac{3}{4}} \sqrt{z^2 + x1}}\:\:,
\end{eqnarray}
and we can check that $\max_{p\in [0,1]} G(p,x)<1$ when $\log(x)\leq
1$. We can also check that $\min_{p\in [0,1]} G(p,x)>-1$ when $\log(x)\geq
-1$, so eq. (\ref{eqlog1}) has in fact no solution whenever $\log x
\in [-1,1]$, regardless of the choice of $r$. Hence, in
this case, $F(p,x)$ is concave in $p$ and we get $F(p,x) \geq
p(x-1)+1$, for any $\log x \in [-1,1]$.\\

\noindent Step 2. Suppose now that $|\log x| >
1$. We have
\begin{eqnarray}
\frac{\partial F}{\partial x} & = & \frac{1}{x}\cdot\left(p+2\phi_u(p)\log x\right) \cdot G_p(x)\:\:,\\
\frac{\partial^2 F}{\partial x^2}  & = & \frac{1}{x^2}\cdot\left(4\phi_u^2(p) \log^2x +
  2\phi_u(p)(2p-1)\log x + 2\phi_u(p) - p(1-p)\right) \cdot G_p(x)\:\:.
\end{eqnarray}
We have $(\partial F/\partial x) (p,1)= p$ and convexity is ensured as long as 
\begin{eqnarray}
\log x & \not\in & \left[\frac{1-2p \pm
    \sqrt{1-8\phi_u(p)}}{4\phi_u(p)}\right] \defeq \mathcal{A}\:\:.
\end{eqnarray}
It happens that $\mathcal{A} \subset [-1,1]$, so whenever $|\log x| \geq
1$, $F(p,x)$ is convex in $x$. To finish Step 2, considering only the
case $\log x\geq 1$, it is sufficient to show
that $(\partial F/\partial x) (p,e)\geq p$, or equivalently,
\begin{eqnarray}
H(p) \defeq \left(p+2\phi_u(p)\right)\exp(p + \phi_u(p)) & \geq & ep\:\:,\label{eqH2}
\end{eqnarray}
It can be shown that the first derivative,
\begin{eqnarray}
H'(p) & = & \left(2-p+\frac{2+p-6p^2}{2\phi_u(p)}\right)\cdot
\exp(p + \phi_u(p))\:\:,
\end{eqnarray}
is $\geq e$ for any $p<0.7$ --- so, since both limits in 0 for eq. (\ref{eqH2}) coincide, eq. (\ref{eqH2}) holds for any
$p<0.7$. The second derivative (fixing $Q(p) \defeq 2-13p+34p^2-12p^3-8 p^4 +\phi_u(p)((3p-2)(1-4p^2)+ 4 \phi_u^2(p)(1-p))$),
\begin{eqnarray}
H''(p) & = &\phi_u''(p) \cdot Q(p) \cdot
\exp(p + \phi_u(p))\:\:,
\end{eqnarray}
is strictly negative for $p\geq 0.7$ --- so, since both limits in 1
for eq. (\ref{eqH2}) coincide, eq. (\ref{eqH2}) is strictly concave
for $p\geq 0.7$, it sits above its chord $[(0.7,H(0.7)),(1,e)]$ which
itself sits above $p\mapsto ep$ for $p\leq 1$, so eq. (\ref{eqH2}) holds for any
$p\geq 0.7$. This achieves the proof of Step 2.\\

\noindent Step 3. We now have that ineq. (\ref{pfund}) holds for any $\log x
\geq -1$ and any $p\in [0,1]$. To finish the argument, we just have to
remark that $F(p,x)$ satisfies the following symmetry:
\begin{eqnarray}
F(p,x) & = & x\cdot F\left(1-p,\frac{1}{x}\right)\:\:,\label{psym}
\end{eqnarray}
so assuming that $\log x < -1$, we have $\log(1/x) \geq 1$, so we
reuse Steps 1 and 2 together with eq. (\ref{psym}) to obtain that for
any $\log x < -1$,
\begin{eqnarray}
F(p,x) & = & x\cdot F\left(1-p,\frac{1}{x}\right)\nonumber\\
 & \geq & x\cdot
 \left((1-p)\left(\frac{1}{x}-1\right)+1\right)\nonumber\\
 & & = (1-p)(1-x)+x = p(x-1)+1\:\:,
\end{eqnarray}
as claimed, where the inequality makes use of Steps 1, 2. This
achieves the proof of Lemma \ref{lineq1}.
\end{myproof}\\
To finish the proof of Theorem \ref{lemSG}, we make use of Lemma
\ref{lineq1} as follows, starting from eq. (\ref{tb1}):
\begin{eqnarray}
\expect_{\rw}[\exp(\uplambda (\rw - \expect[\rw] )) & = & (1-p)\cdot \exp(- p\mu \uplambda) + p \cdot \exp\left(\mu (1-p)\uplambda + \frac{\sigma^2
     \uplambda^2}{2}\right)\nonumber\\
 & \leq & \left\{ (1-p)\cdot \exp(- p\mu \uplambda) + p \cdot \exp\left(\mu (1-p)\uplambda\right)\right\}\cdot \exp\left(\frac{\sigma^2
     \uplambda^2}{2}\right) \label{eq21}\\
 & & = \left\{ (1-p)+ p \cdot \exp\left(\mu \uplambda\right)\right\}\cdot \exp(- p\mu \uplambda) \cdot \exp\left(\frac{\sigma^2
     \uplambda^2}{2}\right) \nonumber\\
 & \leq & \exp(p\mu \uplambda) \cdot \exp\left(\phi_u(p) \mu^2\lambda^2\right) \cdot \exp(- p\mu \uplambda) \cdot \exp\left(\frac{\sigma^2
     \uplambda^2}{2}\right)\label{eq22}\\
 & & = \exp\left(\frac{(\sigma^2+2\phi_u(p)\mu^2)
     \uplambda^2}{2}\right)\:\:, \forall \uplambda\in \mathbb{R}\:\:.\label{bss1}
\end{eqnarray}
Ineq. (\ref{eq21}) uses the fact that $\sigma^2
     \uplambda^2\geq 0$, and ineq. (\ref{eq22}) uses Lemma
\ref{lineq1} with $x = \exp(\mu \uplambda)$. Hence, $\rw$ is
sub-Gaussian with parameters $k=1$ and $\upbeta^2 =
\sigma^2+2\phi_u(p)\mu^2 = \sigma^2+2\sqrt{p(1-p)}\mu^2$, as
claimed. This ends the proof of Theorem \ref{lemSG}.
\end{myproof}\\
Theorem \ref{lemSG} leads to the following concentration inequality
for the row- and column-sums of $\B$, which are key to bound
eigenvalues.
\begin{lemma}\label{lemSG1}
Let $\mu^+_{i.} \defeq
\sum_j \mu_{ij}^2$, $\mu^+_{.j} \defeq
\sum_i \mu_{ij}^2$, ${\sigma}^+_{i.} \defeq
\sum_j \sigma_{ij}^2$, ${\sigma}^+_{.j} \defeq
\sum_i \sigma_{ij}^2$, and let $\overline{\mu}^+_{i.} \defeq
\mu^+_{i.} / N$ (and so on for the other averages
$\overline{{\sigma}}^+_{i.}, \overline{{\sigma}}^+_{.j}$). Finally, let
$\tilde{p}^\mu_{i.} \defeq  \sum_j p_{ij}\mu_{ij}^2/\mu^+_{i.}$, 
$\tilde{p}^\mu_{.j} \defeq  \sum_i p_{ij}\mu_{ij}^2/\mu^+_{.j}$ and
\begin{eqnarray}
\nu^r_i & \defeq & \overline{\mu}^+_{i.}\cdot
    \phi(\tilde{p}^\mu_{i.}) +
    \overline{\sigma}^+_{i.}\:\:, \label{defNU1}\\
\nu^c_j & \defeq & \overline{\mu}^+_{.j}\cdot
   \phi(\tilde{p}^\mu_{.j}) +
    \overline{\sigma}^+_{.j}\:\:,
\end{eqnarray} 
where $\phi(p)\defeq 2\sqrt{p(1-p)}$ is (normalized) Matsushita's entropy.
Then the following holds for any $t>0$:
\begin{eqnarray}
\pr\left[\sum_{i}(\rw_{ij} - p_{ij}\mu_{ij}) \not\in (-Nt,Nt)\right] & \leq
& 2\exp\left(-\frac{N t^2}{2\nu^c_j }\right)\:\:,\label{conc111}\\
\pr\left[\sum_{j}(\rw_{ij} - p_{ij}\mu_{ij}) \not\in (-Nt,Nt)\right] & \leq
& 2\exp\left(-\frac{N t^2}{2\nu^r_i }\right)\:\:.\label{conc112}
\end{eqnarray}
\end{lemma}
\begin{myproof}
Since the sum of $N$ independent random variables respectively
$(k,\upbeta_i)$-sub-Gaussian ($i\in [N]$) brings a $(k,\sum_i \upbeta_i)$
sub-Gaussian random variable, Theorem \ref{lemSG} immediately yields:
\begin{eqnarray}
\pr\left[ \frac{1}{N}\sum_{j}(\rw_{ij} - \expect[\rw_{ij}]) \geq t\right] & \leq
& \exp\left(-\frac{N t^2}{2\cdot \frac{1}{N}\sum_j(2\sqrt{p_{ij}(1-p_{ij})}\mu_{ij}^2 + \sigma_{ij}^2)}\right)\:\:.
\end{eqnarray}
Since $p\mapsto \sqrt{p(1-p)}$ is concave, we have:
\begin{eqnarray}
\sum_j \sqrt{p_{ij}(1-p_{ij})}\mu_{ij}^2 & = & \mu^+_{i.}\cdot 
\sum_j \frac{\mu_{ij}^2}{\mu^+_{i.}}\cdot \sqrt{p_{ij}(1-p_{ij})}\nonumber\\  
& \leq & \mu^+_{i.}\cdot \sqrt{\tilde{p}^\mu_{i.}\left(1-\tilde{p}^\mu_{i.}\right)}\:\:.\label{iconc}
\end{eqnarray}
We finally obtain using ineq. (\ref{iconc}),
\begin{eqnarray}
\pr\left[ \frac{1}{N}\sum_{j}(\rw_{ij} - \expect[\rw_{ij}]) \geq t\right] & \leq
& \exp\left(-\frac{N t^2}{2\cdot (\overline{\mu}^+_{i.}\cdot
    2\sqrt{\tilde{p}^\mu_{i.}(1-\tilde{p}^\mu_{i.})} + \overline{\sigma}^+_{i.})}\right)\:\:,\label{conc11}
\end{eqnarray}
and we would obtain by symmetry:
\begin{eqnarray}
\pr\left[ \frac{1}{N}\sum_{j}(\rw_{ij} - \expect[\rw_{ij}]) \leq -t\right] & \leq
& \exp\left(-\frac{N t^2}{2\cdot (\overline{\mu}^+_{i.}\cdot
    2\sqrt{\tilde{p}^\mu_{i.}(1-\tilde{p}^\mu_{i.})} + \overline{\sigma}^+_{i.})}\right) \label{conc12}
\end{eqnarray}
as well. This ends the proof of Lemma \ref{lemSG1}.
\end{myproof}\\
Let us define function $E : \{1, 2, ..., 2N\} \rightarrow
\mathbb{R}_+$ with:
\begin{eqnarray}
E(i) & \defeq & \left\{
\begin{array}{lcl}

    2\sqrt{\tilde{p}^\mu_{i.}(1-\tilde{p}^\mu_{i.})} \cdot
    \overline{\mu}^+_{i.} +
    \overline{\sigma}^+_{i.} & \mbox{ if } & i \leq N\:\:,\\

    2\sqrt{\tilde{p}^\mu_{.(i-N)}(1-\tilde{p}^\mu_{i.})}\cdot \overline{\mu}^+_{.(i-N)} +
    \overline{\sigma}^+_{.(i-N)} & \multicolumn{2}{l}{\mbox{ otherwise
      }}
\end{array}\right.\:\:,\label{defEE}
\end{eqnarray}
which collects the key parts in the concentration inequalities for
row- / column-sums. We need in fact slightly more than Lemma \ref{lemSG1}, as we do not
just want to bound row- or column-sums, but we need to bound their $L_1$ norms (which,
since $\|\ve{u}\|_1 \geq |\ve{1}^\top \ve{u}|$ by the triangle
inequality, yields a bound on row- or column-sums). It can be
verified that $|\rw_{ij}|$ is $(2,\upbeta)$-sug-Gaussian with the
same $\upbeta$ as for $\rw_{ij}$, but because $|\rw_{ij}|$ now
integrates a folded Gaussian random variable \citep{tbhOT} instead of a
Gaussian, its expectation is non trivial. We have not found any (simple)
bound on the expectation of such a folded Gaussian, so we provide a
complete one here for $\rw_{ij}$, which integrates as
well Bernoulli parameter $p_{ij}$.
\begin{lemma}\label{lemabs}
We have:
\begin{eqnarray}
\expect[|\rw_{ij}|] & \leq & p_{ij}\cdot\left(|\mu_{ij}| + \frac{1}{\upgamma}\cdot
\frac{\sigma^2_{ij}}{\sigma_{ij}+|\mu_{ij}|}\right)\:\:,\label{boundEXP}
\end{eqnarray}
where $\upgamma \defeq \sqrt{\pi/2}$. Furthermore, (\ref{boundEXP}) is optimal in
the sense that both sides coincide when $\mu_{ij} = 0$ (in this case, $\expect[|\rw_{ij}|] = \sigma_{ij}/\upgamma$).
\end{lemma}
\begin{myproof}
We now have (removing indices for readability, \cite{tbhOT}):
\begin{eqnarray}
\expect[|\rw|] & = & p\left(\sqrt{\frac{2}{\pi}}\cdot \sigma
  \exp\left(-\frac{\mu^2}{2\sigma^2}\right) +
  \mu\left[1-2\Phi\left(-\frac{\mu}{\sigma}\right)\right]\right)\:\:,
\end{eqnarray}
where $\Phi$ is the CDF of the standard Gaussian, so it is clear that
the statement of the Lemma holds (and is in fact tight) when $\mu = 0$, as in this case
$\expect[|\rw|] = \sigma\sqrt{2/\pi}$. Otherwise, assume $\mu \neq 0$.
For any $z>0$, let
\begin{eqnarray}
f(z) & \defeq & \frac{1}{1+\sqrt{1+\frac{4}{z^2}}} \cdot
\left(\sqrt{\frac{2}{\pi}}\cdot \frac{1}{z}\exp\left(-\frac{z^2}{2}\right)\right)\:\:,
\end{eqnarray}
where $u>0$ is a constant. It comes from \citep[Inequality 7.1.3]{asHO}:
\begin{eqnarray}
\Phi(z) & \leq &
1-f(z)\:\:,
\end{eqnarray}
and so, if $\mu < 0$, 
\begin{eqnarray}
\expect[|\rw|] & = & p\left(\mu + \sqrt{\frac{2}{\pi}}\cdot \sigma
  \exp\left(-\frac{\mu^2}{2\sigma^2}\right) -
  2\mu\Phi\left(-\frac{\mu}{\sigma}\right)\right) \nonumber\\
& \leq & p\left(-\mu + \sqrt{\frac{2}{\pi}}\cdot \sigma
  \exp\left(-\frac{\mu^2}{2\sigma^2}\right) +
  2\mu f\left(-\frac{\mu}{\sigma}\right)\right) \nonumber\\
 & & = p\left(|\mu| + \sqrt{\frac{2}{\pi}}\cdot \sigma
  \exp\left(-\frac{\mu^2}{2\sigma^2}\right) \left[1 -
    \frac{2}{1+\sqrt{1+\frac{4\sigma^2}{\mu^2}}}\right]\right)\:\:,\label{bs13}
\end{eqnarray}
and we would obtain the same bound for
$\mu>0$. 
There just remains to remark that ($\forall z>0$):
\begin{eqnarray*}
1-\frac{2}{1+\sqrt{1+\frac{1}{z}}} & \leq & 1 - 2\sqrt{z} + 2z\:\:, \\
\left(1-z+\frac{z^2}{2}\right)\cdot \exp\left(-\frac{z^2}{2}\right) &
\leq & \frac{1}{1+z}\:\:,
\end{eqnarray*}
and we obtain the statement of Lemma \ref{lemabs}.
\end{myproof}\\
Using Lemma \ref{lemabs} , we can extend Lemma \ref{lemSG1} and obtain
the following Lemma.
\begin{lemma}\label{lemSG2}
Let $E^\star\defeq \max_i E(i)$ and $\mathbb{A}$
denote the event:
\begin{eqnarray}
\mathbb{A} \equiv
\left(\exists j\in [N] : \|\ve{c}_j\|_1 > \sum_i p_{ij}(|\mu_{ij}| +
  \delta_{ij})+ Nt\right) \vee\left( \exists i\in [N] : \|\ve{r}_i\|_1 >  \sum_j p_{ij}(|\mu_{ij}| + \delta_{ij})+ Nt\right)
\end{eqnarray}
Then for any $t>0$, 
\begin{eqnarray}
\pr\left[\mathbb{A} 
\right]
 & \leq
& 4N\exp\left(-\frac{N t^2}{2 E^\star}\right)\:\:,\label{conc13}
\end{eqnarray}
where $\ve{r}_i \defeq (\B \ve{1})_i$ and $\ve{c}_j \defeq
(\B^\top\ve{1})_j$ are respectively row- and column-sums in $\B$, $\delta_{ij} \defeq \sigma^2_{ij} / (\sigma_{ij} + \upgamma
|\mu_{ij}|)$ and $\upgamma \defeq \sqrt{\pi/2}$.
\end{lemma}
The way we use Lemma \ref{lemSG2} is the following: pick
\begin{eqnarray}
t & = & \sqrt{\frac{2E^\star}{N} \cdot \log \frac{4N}{\updelta}}\:\:.
\end{eqnarray}
We get that with probability $\geq 1 - \updelta$, we shall have both
\begin{eqnarray}
\|\ve{c}_j\|_1 & \leq & \sum_i \tilde{b}_{ij} + \sqrt{2E^\star N \cdot \log
    \frac{8N}{\updelta}}\:\:, \forall j \in [N]\:\:,\label{int1}\\
\|\ve{r}_i\|_1 & \leq & \sum_j \tilde{b}_{ij} + \sqrt{2E^\star N \cdot \log
    \frac{8N}{\updelta}} \:\:, \forall i \in [N]\:\:,\label{int2}
\end{eqnarray}
for all columns and rows in $\B$, with $\tilde{b}_{ij} \defeq
p_{ij}(|\mu_{ij}| + \delta_{ij})$. There is a balance between the two
summands in (\ref{int1}), (\ref{int2}) that we need to clarify to
handle the upperbounds. This is achieved through the following Lemma.
\begin{lemma}\label{boundB}
For any $i,j$,
\begin{eqnarray*}
\frac{1}{N} \sum_j \tilde{b}_{ij} & \leq & \sqrt{2
\tilde{p}^\mu_{i.}\overline{\mu}^+_{i.} + 2
\tilde{p}^\sigma_{i.}\overline{\sigma}^+_{i.}}\:\:,\\
\frac{1}{N} \sum_i \tilde{b}_{ij} & \leq & \sqrt{2 \tilde{p}^\mu_{.j}\overline{\mu}^+_{.j} + 2 \tilde{p}^\sigma_{.j}\overline{\sigma}^+_{.j}}\:\:,
\end{eqnarray*}
where $\tilde{p}^\sigma_{i.} \defeq  \sum_j p_{ij}\sigma_{ij}^2/\sigma^+_{i.}$, 
$\tilde{p}^\sigma_{.j} \defeq  \sum_i p_{ij}\sigma_{ij}^2/\sigma^+_{.j}$.
\end{lemma}
\begin{myproof}
We have for any $i,j$,
\begin{eqnarray}
\left(\frac{1}{N} \sum_j \tilde{b}_{ij}\right)^2 & = & \left(\frac{1}{N} \cdot \sum_j
p_{ij} |\mu_{ij}|\left(1+
\frac{\sigma_{ij}}{|\mu_{ij}|}\cdot \frac{1}{1 + \upgamma
\frac{|\mu_{ij}|}{\sigma_{ij}}}\right)\right)^2\nonumber\\
 & \leq & \left(\frac{1}{N} \cdot \sum_j
p_{ij} |\mu_{ij}| + \frac{1}{N} \cdot \sum_j
p_{ij}\sigma_{ij}\right)^2\nonumber\\
 & \leq & 2\left(\frac{1}{N} \cdot \sum_j
p_{ij} |\mu_{ij}|\right)^2 + 2\left(\frac{1}{N} \cdot \sum_j
p_{ij}\sigma_{ij}\right)^2\label{jeq1}\\
 & \leq & 2\sum_j
p^2_{ij} \mu^2_{ij} + 2\sum_j
p^2_{ij} \sigma^2_{ij}\label{jeq3}\\
 & \leq & 2\sum_j
p_{ij} \mu^2_{ij} + 2\sum_j
p_{ij} {\sigma}^2_{ij}\label{jeq4}\\
 & & = 2 \tilde{p}^\mu_{i.}\overline{\mu}^+_{i.} + 2 \tilde{p}^\sigma_{i.}\overline{\sigma}^+_{i.}\:\:.\label{bUU}
\end{eqnarray}
Ineqs (\ref{jeq1}) and (\ref{jeq3}) follows from $(\sum_{u=1}^v a_u)^2
\leq v\sum_u a_u^2$. Ineq. (\ref{jeq4}) comes from $p_{ij}\in [0,1]$. We
would have similarly
\begin{eqnarray}
\left(\frac{1}{N} \sum_i \tilde{b}_{ij}\right)^2 & \leq & 2 \tilde{p}^\mu_{.j}\overline{\mu}^+_{.j} + 2 \tilde{p}^\sigma_{.j}\overline{\sigma}^+_{.j}\:\:.\label{bUU2}
\end{eqnarray}
This ends the proof of Lemma \ref{boundB}.
\end{myproof}

\subsubsection{Proof of Theorem \ref{thEASYM_DIRAC}}\label{proof_thm_thEASYM_DIRAC}

Let us define function $U : \{1, 2, ..., 2N\} \rightarrow
\mathbb{R}_+$ with:
\begin{eqnarray}
U(i) & = & \left\{
\begin{array}{lcl}
2 \tilde{p}^\mu_{i.}\overline{\mu}^+_{i.} + 2 \tilde{p}^\sigma_{i.}\overline{\sigma}^+_{i.}& \mbox{ if } & i \leq N\:\:,\\
2 \tilde{p}^\mu_{.(i-N)}\overline{\mu}^+_{.(i-N)} + 2 \tilde{p}^\sigma_{.(i-N)}\overline{\sigma}^+_{.(i-N)}& \multicolumn{2}{l}{\mbox{ otherwise
      }}
\end{array}\right.\:\:,\label{defUU}
\end{eqnarray}
which collects the bounds in ineqs (\ref{bUU}) and (\ref{bUU2}), and
let $U^\star \defeq \max_i U(i)$. 
Let
\begin{eqnarray}
\ell & \defeq & N\sqrt{U^\star}+ \sqrt{2E^\star N \cdot \log
    \frac{4N}{\updelta}}\:\:.\label{defELL}
\end{eqnarray}
$\ell$ is be the quantity we need to handle all eigenspectra, but
for this objective, let us define assumption (Z) as:
\begin{itemize} 
\item [(Z)] $(1+\epsilon)
N\sqrt{U^\star} \leq 1/5$ (call it the domination assumption for short) \textit{and}
\begin{eqnarray}
\frac{E^\star}{U^\star} & \leq & \epsilon^2 \cdot \frac{N}{2 \log
    \frac{4N}{\updelta} }\:\:.\label{largeNET2}
\end{eqnarray}
\end{itemize}
Assumption (Z) is a bit technical: we replace it by a simpler one,
(A), which implies (Z). Suppose ${\gamma} \in (0,1)$ a constant, and assume $N\geq
K^{1/(1-\gamma)}$ without loss of generality; fix for some
\textit{constant} $c>0$, 
\begin{eqnarray}
\updelta & = & \frac{1}{N^c}\:\:,\\
\epsilon^2 & = & \frac{2}{N^{1-{\gamma}}}\cdot \log \frac{4N}{\updelta}\nonumber\\
& \geq & \frac{2(c+4)}{N^{1-{\gamma}}}\cdot \log N\:\:.
\end{eqnarray}
Condition (\ref{largeNET2}) is now ensured provided
\begin{eqnarray}
U^\star & \geq &
\frac{1}{N^{\gamma}} \cdot E^\star\:\:,\label{condE2}
\end{eqnarray}
while the domination condition is ensured, with $N =
\Omega(c^{2+\kappa})$ ($\kappa>0$ a constant) large enough so that $\epsilon \leq 1$, as long as
\begin{eqnarray}
U^\star & \leq & \frac{1}{100N^2}\:\:.
\end{eqnarray}
So let us simplify assumption (Z) by the following assumption, which implies (Z):
\begin{itemize}
\item [(A)] $c>0$ and $0 < \gamma
< 1$ being constants such that $N=\Omega(poly(c), 3^{1/(1-\gamma)})$, we have:
\begin{eqnarray}
U^\star & \in &
\left[\frac{E^\star}{N^{\gamma}} 
 , \frac{1}{100 N^2}\right]\:\:.\label{intUS}
\end{eqnarray}
\end{itemize}
Again, (A) implies (Z).\\

\begin{center}
\colorbox{gray!10}{\fbox{%
    \parbox{0.95\textwidth}{
\noindent \textbf{Remark 1}: the upperbound of (\ref{intUS}) is
quantitatively not so
different from \cite{Linderman2014}'s assumptions. They work with two assumptions, the first of which being
\begin{eqnarray}
\sigma^2 & \leq & \frac{1}{N}\label{laDLCM1}
\end{eqnarray}
(we consider variances for the assumption to rely on same scales as ours), and also
pick network parameters $\mu, \sigma$ in such a way that large
deviations for edge weights are
controlled with high probability, with a condition that roughly looks like:
\begin{eqnarray}
\mu^2 +
  \frac{c}{N^2}\cdot \sigma^2 & = & O\left(\frac{1}{N^2}\right)\label{laDLCM2}\:\:,
\end{eqnarray}
for some constant $c>3$. This constraint is relevant to
the same stability issues as the ones we study here, and can be found
in a slightly different form (but equivalent) in \citep[Section
4]{Hyvarinen2013}, where it is mandatory for the estimation of ICA model parameters.. 

Finally, \citep{Linderman2014} make the heuristic choice to enforce at least one of the
two ineqs. (\ref{laDLCM1}, \ref{laDLCM2}). \\

\noindent \textbf{Remark 2}: the sampling constraint akin to
eq. (\ref{laDLCM2}) is in fact very restrictive for ICA estimation of
models \citep[Section
4]{Hyvarinen2013}, since typically \textbf{each} coordinate in $\B$
has to be bounded with high probability, whereas in our case, it is
sufficient to control \textbf{sums} ($L_1$, row- or column-wise) with
high probability. We can therefore benefit from concentration
properties on large networks that
such approaches may not have.
}}
}
\end{center}

\begin{figure}[t]
\centering
\scalebox{0.925}{
\begin{tabular}{c}
\includegraphics[trim=200bp 360bp 510bp 100bp,clip,width=.60\linewidth]{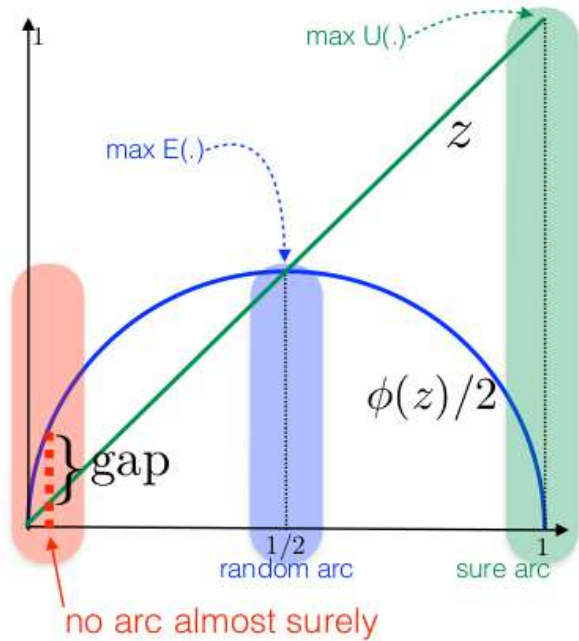} 
\end{tabular}
}
\caption{Region (red cartouche) for which the lowerbound in (\ref{intUS}) may
  fail because $\phi(z)$ happens to be much larger than $2z$ --- a
  worst case corresponding to networks where basically all $p$s are
  very small (\textit{e.g.} $o(1/poly(N))$). The Figure also
  depicts the location of $p$ for "ideal" maximizers of $E(.)$ (hence
  the name, Equivocal arcs) and $U(.)$ (hence the name, Unequivocal arcs).\label{fig:phi}}
\end{figure}

What is interesting from (\ref{intUS}) is the hints that provide the
\textit{lowerbound} of (\ref{intUS}) for Theorem \ref{thEASYM_DIRAC} (main
file) to hold. The main
difference between $U^\star$ and $E^\star$ is indeed (omitting factor
$2\cdot \tilde{p}^\sigma \in [0,1]$ in variance terms) the switch
between $z\mapsto 2z$ (for $U(.)$) and $z\mapsto \phi(z)$ (for
$E(.)$). Figure \ref{fig:phi} explains that the lowerbound may be
violated essentially only on networks with very unlikely arcs almost
everywhere, because $\phi$ has infinite derivative\footnote{And it seems that such entropy-like penalties with infinite derivatives
  in a neighborhood of zero are necessary to obtain Lemma \ref{lineq1}
  --- as explained in the Lemma ---
if we want to keep the sub-Gaussian characterization of the $\rw_{ij}$s.} as $z\rightarrow
0$. Also, it gives a justification for the name of the two
functions $E$ and $U$, where maximizing $E$ tends to favor arcs with $p$ close to $1/2$
($E$ stands for Equivocal), while maximizing $U$ tends to favor arcs with $p$ close to $1$
($U$ stands for Unequivocal).

\noindent ($\star$) We now have all we need to bound the eigenspectra of $\mat{H},
\mat{H}'$. Let $\lambda^{\uparrow}(.)$ (resp. $\lambda^{\downarrow}(.)$) denote
the maximal (resp. minimal) eigenvalue of the argument
matrix. 
We obtain
that with probability $\geq 1 - \updelta$, 
\begin{eqnarray}
\lambda^{\uparrow}(\mat{H}) & \leq & 1 + \max_j \ve{c}^\top_j \sum_k \ve{c}_k -
(\ve{r}_j + \ve{c}_j)^\top \ve{1} \nonumber\\
 & \leq &  1 + \max_j \|\ve{c}_j\|_1 \max_k | \ve{1}^\top \ve{r}_k| -
(\ve{r}_j + \ve{c}_j)^\top \ve{1}\label{holder1}\\
 & \leq & 1 + \max_j \|\ve{c}_j\|_1 \max_i \|\ve{r}_i\|_1 + \max_j
 \|\ve{c}_j\|_1 + \max_i \|\ve{r}_i\|_1\nonumber\\
 & \leq &  \left(1 + \ell\right)^2\:\:,\nonumber
\end{eqnarray}
(ineq. (\ref{holder1}) comes from H\"older inequality) and similarly for the minimal eigenvalue,
\begin{eqnarray}
\lambda^{\downarrow}(\mat{H})  & \geq & 1 + \min_j \ve{c}^\top_j \sum_k \ve{c}_k -
(\ve{r}_j + \ve{c}_j)^\top \ve{1} \nonumber\\
 & \geq &  1 - \ell^2 - 2 \ell\:\:,\nonumber
\end{eqnarray}
which implies that $\ell \leq \sqrt{2}-1$ for this latter bound not to
be vacuous ($\ell$ is defined in eq. (\ref{defELL})). As long as $\updelta = \Omega(1/poly(N))$, it is not
hard to see that $N\sqrt{U^\star}$ dominates in $\ell$ for
large networks so we can assume $N$ large enough so that, for some
small $\epsilon>0$,
\begin{eqnarray}
\frac{E^\star}{U^\star} & \leq & \epsilon^2 \cdot \frac{N}{2 \log
    \frac{4N}{\updelta} }\:\:,\label{largeNET}
\end{eqnarray}
which brings $\ell \leq (1+\epsilon) N\sqrt{U^\star}$.
In this case, if $(1+\epsilon) N\sqrt{U^\star}\leq 1/5$, then
$\lambda^{\downarrow}(\mat{H})\geq 1/2$. Furthermore, it is not hard
to check that we also get $\lambda^{\uparrow}(\mat{H}) \leq 3/2$. To
summarize, as long as assumption (Z) (and so, as long as (A)) holds, 
the \textit{complete} eigenspectra of $\mat{H}$, $\mat{H}^{-1}$ and by
extension $\mat{H}'$, $\mat{H}'^{-1}$, all lie within $[1/2, 2]$ with
high probability. \\

\noindent ($\star$) We finish with the eigenspectrum of $\mat{J}$. 
We also easily obtain that
\begin{eqnarray}
\lambda^{\uparrow}(\mat{J}) & \leq & \max_j \ve{r}^\top_j \sum_k \ve{r}_k  \nonumber\\
 & \leq &  \max_j \|\ve{r}_j\|_1 \max_k | \ve{1}^\top \ve{c}_k| \nonumber\\
  & \leq &  \ell^2\:\:,\nonumber
\end{eqnarray}
and obviously $\lambda^{\downarrow}(\mat{J}) \geq 0$, which is all we
need.

\noindent ($\star$) We now finish the proof of Theorem \ref{thEASYM_DIRAC},
recalling that $\Covy$ can be summarized as:
\begin{eqnarray}
\Covy & = & \mat{A} + \varf \mat{B} + \varnoise \I\:\:,
\end{eqnarray}
with $\mat{A} \defeq \mat{H}^{-1} \kron \Kt$ has an eigensystem which
is the (Minkowski) product of the eigensystems of its two matrices, and
therefore is within $[\lambda^{\downarrow}(\Kt)/2,
2\lambda^{\uparrow}(\Kt)]$; on the other hand, $\mat{B}\defeq (\I - \B)^{-1} \mat{J} (\I - \B)^{-\top}
                \kron \Sigmai$ has eigensystem which is the one of
                $\mat{J}\mat{H}'$ (eigenvalues have different
                algebraic multiplicity though), which therefore is within
                $[0,\ell^2 \cdot (1+\ell)^2] \subset [0,2/25]$. Hence,
                simplifying a bit, we can bound the complete eigenspectrum
                of $\Covy$, $\lambda(\Covy)$, as:
\begin{eqnarray}
\lambda(\Covy) & \subset & \left[\frac{\lambda^{\downarrow}(\Kt)}{2}
  + \varnoise, 2\lambda^{\uparrow}(\Kt) + \varf + \varnoise\right]\:\:,
\end{eqnarray}
under assumption (A), with probability $\geq 1-\updelta = 1 - 1/N^c$,
as claimed. This ends the proof of Theorem \ref{thEASYM_DIRAC}.

\subsubsection{From Theorem \ref{thEASYM_DIRAC} to Theorem \ref{thEASYM}}\label{proof_thm_thEASYM2}

We now assume $\ra \sim \mathcal{B}(\rp_{ij})$ with $\rp_{ij} \sim
\mathcal{V}_{ij}(p_{ij})$, where $\mathcal{V}$ is a random variable
satisfying $p_{ij} \defeq \expect[\mathcal{V}_{ij}]$ and
$\mathrm{supp}(\mathcal{V}) \subseteq [0,1]$ (the support of $\mathcal{V}$). The proof essentially follows
that of Theorem \ref{thEASYM_DIRAC}, with the following minor changes.\\

\noindent ($\star$) The derivation of eq. (\ref{ndis2}) now satisfies,
since $\phi_u(z)$ is maximal in $z=1/2$,
\begin{eqnarray}
\expect_{\rw_{ij}}[\exp(\uplambda (\rw_{ij} - \expect[\rw_{ij}] ))]  &
\leq & \int_{\mathrm{Supp}(\mathcal{V})}
\exp\left(\frac{(\sigma_{ij}^2+2\phi_u(z)\mu_{ij}^2) 
     \uplambda^2}{2}\right) \mathrm{d}\mu(z)\nonumber\\
 & \leq & \exp\left(\frac{(\sigma_{ij}^2+2\phi_u(1/2)\mu_{ij}^2) \uplambda^2}{2}\right)  =  \exp\left(\frac{(\sigma_{ij}^2+\mu_{ij}^2) \uplambda^2}{2}\right)  \:\:.
\end{eqnarray}
\noindent ($\star$) Assumption (A) now reads, for some constants $c>0$ and $0 < \gamma
< 1$ such that $N=\Omega(poly(c), 3^{1/(1-\gamma)})$, we have:
\begin{eqnarray}
U^\star & \in &
\left[\frac{S^\star}{N^{\gamma}} 
 , \frac{1}{100 N^2}\right]\:\:,\label{intUS22}
\end{eqnarray}
where $U$ does not change but
\begin{eqnarray*}
S(i) & \defeq & \left\{
\begin{array}{ll}

    \overline{\mu}^+_{i.} +
    \overline{\sigma}^+_{i.}& (i \leq N)\\
\overline{\mu}^+_{.j} +
    \overline{\sigma}^+_{.j}: j\defeq N-i & (i>N)
\end{array}\right. \:\:.
\end{eqnarray*}

\subsection{Marginal Likelihood Given the Network Parameters}\label{sec-marlik}
When calculating the expected log-likelihood it is easier to work with \textit{the inverse model}:
\begin{align}
z_i(t) &  \sim \GP(\mathbf{0}, \kernelt(t,t';\btheta)), \\
\bepsilonf  & \sim \Normal(0, \varf \I) \\
\epsilony & \sim \Normal(0, \varnoise) \\ 	
f_i(t) & \sim \matrow{ \G }{i} \left(\z(t) + \B \bepsilonf \right) \text{,} \\
y_i(t) & \sim f_i(t) + \epsilony \text{.}
\end{align}
where $\B = \A \hada \W$; $\G = (\I - \B)^{-1}$; $\matrow{\mat{M}}{i}$ denotes the $i\mth$ row of matrix $\mat{M}$.
Here we analyse the conditional likelihood by integrating out everything but $\A, \W$. Clearly,
for fixed $\A, \W$, since all the distributions are Gaussians, and we are only applying linear operators, 
the resulting distribution over $f_i$, and consequently over $y_i$, is also a Gaussian process. Hence, we 
only need to figure out the mean function and the covariance function of the resulting process.
For the expectation we have that:
\begin{equation}
\mu_i(t) = \expectation{}{f_i(t)} = 0  \text{,}
\end{equation}
since both $\z$ and $\epsilonf$ are zero-mean processes. For the covariance function we have that:
\begin{align}
\covariance{f_i(t), f_j(t')} &= \expectation{}{(f_i(t) - \mu_i(t))(f_j(t) - \mu_j(t'))} \\
%	& =  \sum_{\ell=1}^N \matentry{\G}{i}{\ell} \matentry{\G}{j}{\ell} 
%	 \left(\kernelt(t,t'; \btheta) + \delta_{ij} \varn  \right)
&= \matentry{ \G \G^T }{i}{j} \kernelt(t,t'; \btheta) + 
\matentry{\G \B \B^T \G^T}{i}{j} \varf    \\
\label{eq:covfuncyf}
& = \matentry{\Kf}{i}{j} \kernelt(t,t'; \btheta) 
+ \matentry{\E}{i}{j} \varf \text{,}
\end{align} 
where we have defined $\matentry{\mat{M}}{i}{j}$ the $i,j$ entry of matrix $\mat{M}$ and the 
matrix of latent node covariances and noise covariances as:
\begin{align}
\Kf &=  {\G} {\G}^T \\
\E &= \G \B \B^T \G^T \text{.}
\end{align}
The covariance function of the observations is then given by:
\begin{align}
\label{eq:covfuncy}
\covariance{y_i(t), y_j(t')} &= 
\matentry{\Kf}{i}{j} \kernelt(t,t'; \btheta) 
+ \matentry{\E}{i}{j} \varf +
\delta_{ij} \varnoise \text{.}
\end{align}
For further understanding of this model, let us assume that 
the observations lie on a grid in time, $t=1, \ldots, T$ and 
$\Y$ is a $N \times T$ matrix of observations with 
$\y = \vect{\Y}$ 
hence the likelihood  of all observations is:
\begin{align}
p(\y | \W, \A) &= \Normal(\y; \vec{0}, \Covy ) \text{, with} \\
\Covy  &= \Kf \kron \Kt + \E \kron \varf \I  
+ \I \kron \varnoise \I \text{,}
\end{align}
where $\kron$ denotes the Kronecker product;
If we use this setting then we obtain:
\begin{align}
\Covy  &= \Kf \kron \Kt + (\varf \E + \varnoise \I) \kron \I \text{.}
\end{align}

Interestingly, the model above has been studied in statistics and 
in machine learning, see e.g.~\cite{bonilla-et-al-nips-08,rakitsch2013all}. 
Furthermore, inference and hyperparameter estimation can be done 
efficiently by exploiting properties of the Kronecker product, 
e.g.~an evaluation of the marginal likelihood can be done 
in $\bigO(N^3 + T^3)$. Nevertheless, unless there is a substantial  
overlapping between the locations of the observations across 
the nodes (i.e.~times), the Kronecker formulation becomes 
intractable.

\subsection{Marginal likelihood}\label{sec-marlik2}
Assuming the general case (i.e.~non-grid observations), let us refer to 
$\Covy = \K + \varnoise \I$ as the covariance of the marginal process over $\y$, 
as induced by the covariance function in Equation \eqref{eq:covfuncy},
where $\K$ is the covariance matrix  induced by the covariance function in 
Equation \eqref{eq:covfuncyf}. Therefore, the prior, conditional likelihood,
and marginal likelihood of the model are: 
\begin{align}
p(\f) &= \Normal(\f; \vec{0}, \K) \text{,} \\
p(\y | \f) &= \Normal(\y; \f, \varnoise \I) \text{,} \\
p(\y ) &= \Normal(\y; \vec{0}, \Covy ) \text{,}
\end{align}
where we have omitted the dependencies of the above equation on the network 
parameters $\A, \W$. Because of the marginalization property of GPs it is easy to see 
that all the above distributions are $n$-dimensional, where $n = \sum_{i=1}^N T$, 
where $T$ is the number of observations per node. Hence the cost of evaluating the exact marginal likelihood is $\bigO(n^3)$. 
\subsection{Efficient Computation of Marginal Likelihood Given Network Parameters}\label{sec-effcom}
For simplicity, we consider here the synchronized case where all the $N$ nodes in the network have 
$T$ observations at the same times. i.e.~the total number of observations 
is $n = N \times T$. 
Here we show an efficient expression for the log marginal likelihood:
\begin{align}
\log p(\y | \W, \A) &=  -\frac{n}{2} \log (2\pi) - \frac{1}{2} \log \det{\Covy} - \frac{1}{2} \y^T \Covy^{-1} \y  \text{, where} \\
\Covy  &=  \Kf \kron \Kt + \Covnoise \kron \Sigmai \text{, with } \\
\Covnoise & = (\varf \E + \varnoise \I) \text{ and} \\
\Sigmai &= \I
\end{align}
The main difficulty of computing this expression is the calculation of the log determinant 
of an $n$ dimensional matrix, as well as solving an $n$-dimensional system of linear equations.
Our goal is to show that we never need to solve these operations on an $n$-dimensional matrix, 
which are $\bigO(n^3)$ but instead use $\bigO(N^3 + T^3)$ operations.

Given the eigen-decomposition of the above matrices
\begin{align}
\Covnoise &= \Qnoise \Lambdanoise \Qnoise^T \\
\Sigmai &= \Qi \Lambdai  \Qi^T = \I	\text{,}
\end{align}

It is possible to show that the marginal covariance is given by
\begin{align}
\Covy  &= (\Qnoise \Lambdanoise ^{1/2} \kron \Qi \Lambdai^{1/2})
\left( \Ktildef \kron \Ktildet + \I \kron \I \right) 
(\Qnoise \Lambdanoise ^{1/2} \kron \Qi \Lambdai^{1/2})^T \text{, where} \\
\Ktildef & = \Lambdanoise^{-1/2} \Qnoise^T \Kf \Qnoise \Lambdanoise^{-1/2} \\
\Ktildet &=  \Lambdai^{-1/2} \Qi^T \Kt \Qi \Lambdai^{-1/2} = \Kt
\end{align}
For these matrices we also define their eigen-decomposition analogously to above:
\begin{align}
\Ktildef & = \Qtildef \lambdatildef \Qtildef^ T \\
\Ktildet & = \Kt =  \Qtildet \lambdatildet \Qtildet^ T 
\end{align}
\subsubsection{Log-determinant Term}
\begin{align}
\log \det{\Covy} & =   \log \det{\Covnoise \kron \Sigmai  } + \log \det{\Ktildef \kron \Ktildet + \I \kron \I}  \\
&=	T \sum_{i=1}^N \log \lambda_n^{(i)}  + N \sum_{j=1}^T \log \lambda_\text{I}^{(j)} 
+  \sum_{i=1}^N \sum_{j=1}^T\log ( \lambdatilde_f^{(i)} \lambdatilde_t^{(j)} + 1 ) \\
&= 	T \sum_{i=1}^N \log \lambda_n^{(i)}  +  \sum_{i=1}^N \sum_{j=1}^T\log ( \lambdatilde_f^{(i)} \lambdatilde_t^{(j)} + 1 )
\end{align}
\subsubsection{Quadratic Term}
\begin{align}
\y^T \Covy^{-1} \y & =  \y^T 	( \Lambdanoise ^{1/2} \Qnoise^T \kron \Lambdai^{1/2} \Qi^T)^{-1}
\left( \Ktildef \kron \Ktildet + \I \kron \I \right)^{-1}
(\Qnoise \Lambdanoise ^{1/2} \kron \Qi \Lambdai^{1/2})^{-1} 
\y \\
\y^T \Covy^{-1} \y & =  \y^T 
(  \Qnoise \Lambdanoise ^{-1/2} \kron  \Qi \Lambdai^{-1/2})
\left( \Ktildef \kron \Ktildet + \I \kron \I \right)^{-1}
(\Lambdanoise ^{-1/2} \Qnoise^T  \kron  \Lambdai^{-1/2} \Qi^T) 
\y
\end{align}
Let us define 
\begin{align}
\ytilde &=  (\Lambdanoise ^{-1/2} \Qnoise^T  \kron  \Lambdai^{-1/2} \Qi^T)  \y \\
&= \vect (\Lambdai^{-1/2} \Qi^T \Y \Qnoise \Lambdanoise^{-1/2}) \\
&= \vect ( \Y \Qnoise \Lambdanoise^{-1/2}) 
\end{align}
Hence the quadratic form above becomes:
\begin{align}
\y^T \Covy^{-1} \y & =  \ytilde^T \left( \Ktildef \kron \Ktildet + \I \kron \I \right)^{-1} \ytilde \\
& = \trace(\Ytilde^T \Qtildet  \Ytildetf \Qtildef^T ) \text{, where}
\end{align}
\begin{equation}
\matentry{\Ytildetf}{i}{j} = \left(\frac{1}{ \matentry{\blambdatildet \blambdatildef^T + 1}{i}{j} }\right)  \matentry{\Qtildet^T \Ytilde \Qtildef}{i}{j}
\end{equation}
and $\y = \vect(\Y)$,  $\ytilde = \vect({\Ytilde})$ are the 
vectors obtained by stacking the columns of the 
$T \times N$ matrices $\Y$ and $\Ytilde$ respectively.

%\noteEB{Add properties of Kronecker products and refer to those properties used above}

%\vskip 0.3in
%\newpage
\section{Experiments}\label{exp_expes}
As mentioned in the main paper, the choice of baseline comparisons was based on \citet{peters2014causal}. Other than the methods discussed in the main paper, there are four other methods considered by \citet{peters2014causal}:
(1) Brute-force search;
(2) Greedy DAG Search \citep[\gds, see e.g.][]{chickering2002optimal}; 
(3) Greedy equivalence search \citep[\ges,][]{chickering2002optimal,meek1997graphical}; 
(4) Regression with subsequent independence test \citep[\resit,][]{peters2014causal}. 

In the experiments reported in section~\ref{sec:fmri}, since the ground truth is known, the evaluation criteria is AUC (area under the ROC curve). Calculating AUC values requires a  discriminative threshold to generate ROCs. In the case of \gds\ and \ges\  there was no clear parameter that could be considered as the discriminative threshold, and therefore results for these algorithms are not reported. 
In the case of \resit, there is a  threshold, but the threshold values for which the method produces different results were not provided, making it infeasible to calculate AUC, and therefore the output of this algorithm is not reported. 
In the experiments reported in section~\ref{sec:prices}, the implementations of  \ges, \gds\ and \resit\ that we used returned an error (possibly because the number of nodes was greater than the observations from each node). Therefore their results are not reported. 
Finally, for the experiment  in section~\ref{sec:gene} we compared the
results with \cpc, which provided comparatively good performance in
other experiments. Also, we did not include the brute-force method,
which is not feasible to perform in networks with more than four
nodes, and therefore makes it inapplicable in the experiments studied here.

The \pc\ and \cpc\ algorithms are constrain-based structure learning methods for directed acyclic graphs (DAG). The algorithms require a conditional independence test, for which we used the test for zero partial correlation between variables. The \iamb\ method is a two-phase algorithm for Markov blanket discovery. Linear correlation is used for the test of conditional independence required by this algorithm. The \lingam\ method is a Linear non-Gaussian Additive Model (LiNGAM) for estimating structural equation models. \pwlingam\ provides the direction of connection between the two connected nodes. We used partial correlation for determining whether two nodes are connected, and the magnitude of the correlation was used as the discriminative threshold. For connected nodes at the threshold \pwlingam\ was used to determine the direction of the connection.

For [\pc, \cpc\, \ges], \iamb\ and \lingam\ implementations provided by R packages \citet{kalisch2012causal}, \citet{marco2010learning}, \citet{kalisch2012causal} were used respectively. For \pwlingam\ the code provided by the authors was re-implemented in R and was used. For \gds\ and \resit\ implementation provided by authors of \citet{peters2014causal} in R was used.
\subsection{Prior setting and optimization specifics}
Similarly to \citet{Maddison2016}, different $\lambda_c$ values are used for the prior and posterior distributions. For experiments with $N>15$, following \citet{Maddison2016} we used $\lambda_c=0.5$ for  priors and $\lambda_c=2/3$ for posterior distributions. For the experiments in section~\ref{sec:fmri}, in which $N\le 15$, we used the first subject ($T=200$) as the validation data and selected $\lambda_c=1.0$ for priors and $\lambda_c=0.15$ for posterior distributions. The number of Monte Carlo samples was selected based on computational constraints, and were $200$, $20$ and $2$ samples for small-scale (\S \ref{sec:fmri}, \ref{sec:prices}), medium-scale (\S \ref{sec:gene}), and large-scale (\S \ref{sec:gene}) experiments respectively. 

Prior over $W_{ij}$ is assumed to be zero-mean Gaussian distribution with variance $\varw=2/N$ similar to \citet{Linderman2014}. Prior over $A_{ij}$ is assumed to be $\Concrete(1, \lambdaconc)$, which implies that the probability that a link exists between two nodes is $0.5$:
\begin{align}
p(A_{ij}) = \Concrete(1, \lambdaconc) \text{,} \quad p(W_{ij}) = \Normal(0, 2/N) \text{,}
\end{align}

\subsection{Brain functional connectivity data}\label{secsup-fmri}
\begin{figure*}[t]
	\centering
	\includegraphics[scale=1]{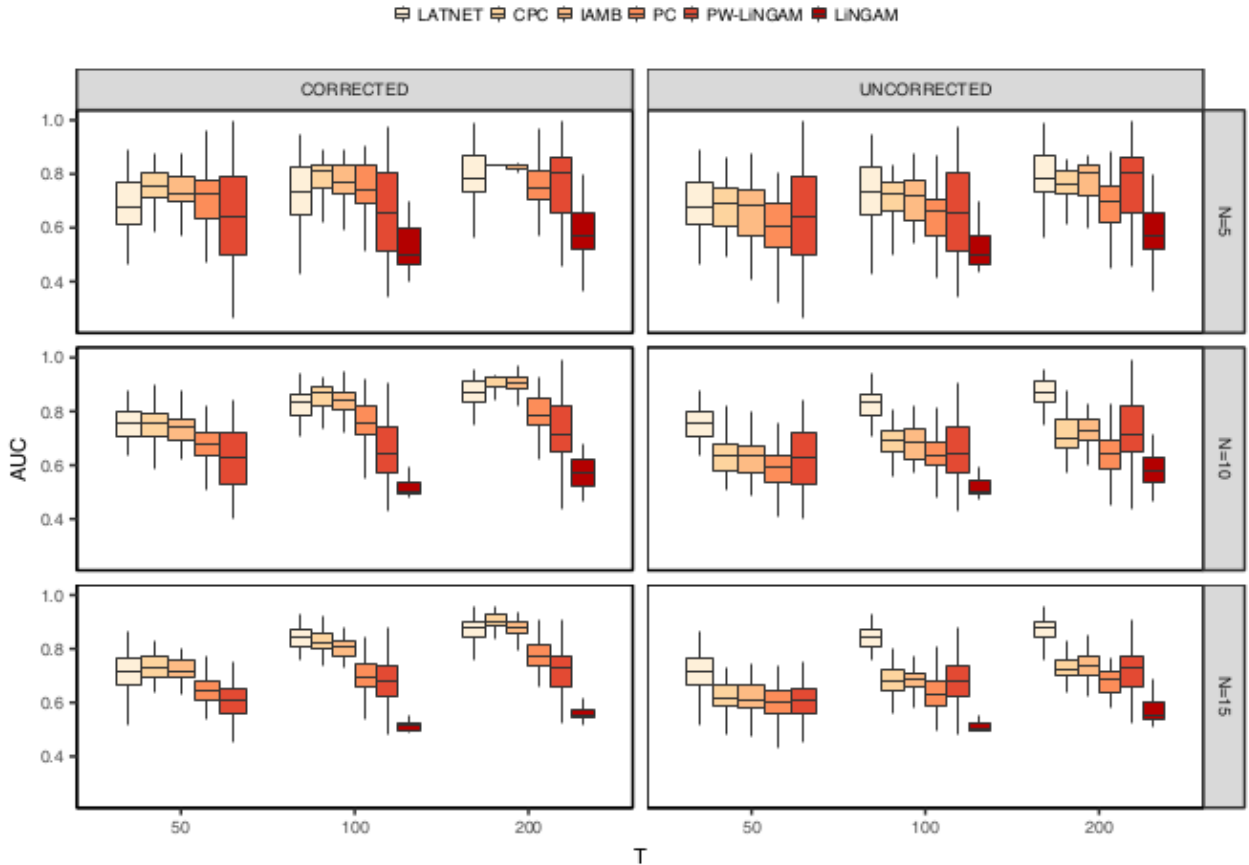}
	\caption{Performance of the methods in link prediction on the
          brain functional connectivity data in terms of AUC. $N$ is
          the number of nodes in the network, and $T$ is the number of
          observations in each node. See text for the definition and
          discussion of
          \corrected~(after concavification of the ROC curve) and
          \uncorrected~(before concavification, \textit{i.e.}
          represents the actual algorithm's outputs).}
	\label{fig:fmri2}
\end{figure*}

\begin{figure*}[t]
	\centering
	\includegraphics[scale=1]{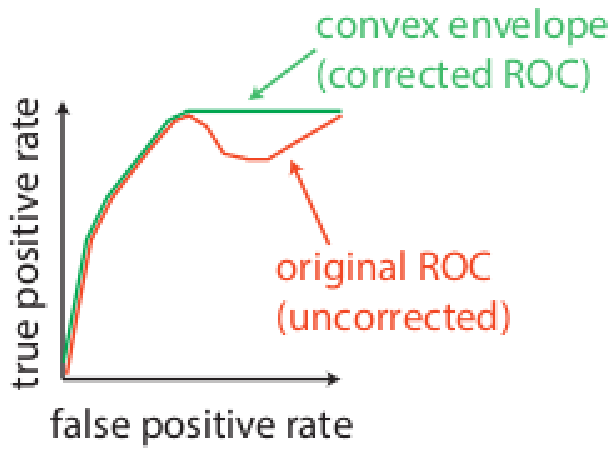}
	\caption{The correction applied to ROC curve in order to calculate AUC. Note that this correction was not applied to the output of \latnet since in the case of \latnet the ROC curve is increasing.}
	\label{fig:roc_correction}
\end{figure*}

\textbf{AUC computation}. This is obtained by varying the discrimination threshold and drawing the false-positive rate (fpr) vs true-positive rate (tpr). In the case of \latnet, this threshold is the absolute expected value of the overall connection strength between the nodes ($|\mu_{ij}p_{ij}|$). In the case of \pc, \cpc\ and \iamb\ algorithms, the discrimination threshold is the $p$-value (target type I error rate) of the conditional independence test, and in the case of \lingam\ and \pwlingam, absolute values of the estimated linear coefficients and partial correlation coefficients are used as the discrimination thresholds respectively. The ROC curve from which the AUC is calculated, is required to be an increasing function, however, in the case of \pc, \cpc\ and \iamb\ algorithms, the fpr/tpr curve can be decreasing in some parts. This is because these algorithms might remove an edge from the graph after increasing the significance level in order to ensure that the resulting graph is a DAG. In such cases, one can removed the decreasing parts by computing the AUC from the non-decreasing portions of the curve (Figure~\ref{fig:roc_correction}; concave envelope of the curve). This correction provides an upper-bound on the AUC of these methods. Figure~\ref{fig:fmri2} shows the AUC of the methods for different network sizes bot both
corrected AUC (\corrected) and uncorrected AUC (\uncorrected). 

\subsection{Spellman's sentinels of the yeast cell cycle}\label{sec-spel}

\begin{figure}[t]
\centering
\scalebox{0.925}{
\begin{tabular}{ccc}
\includegraphics[trim=20bp 0bp 0bp 0bp,clip,width=.32\linewidth]{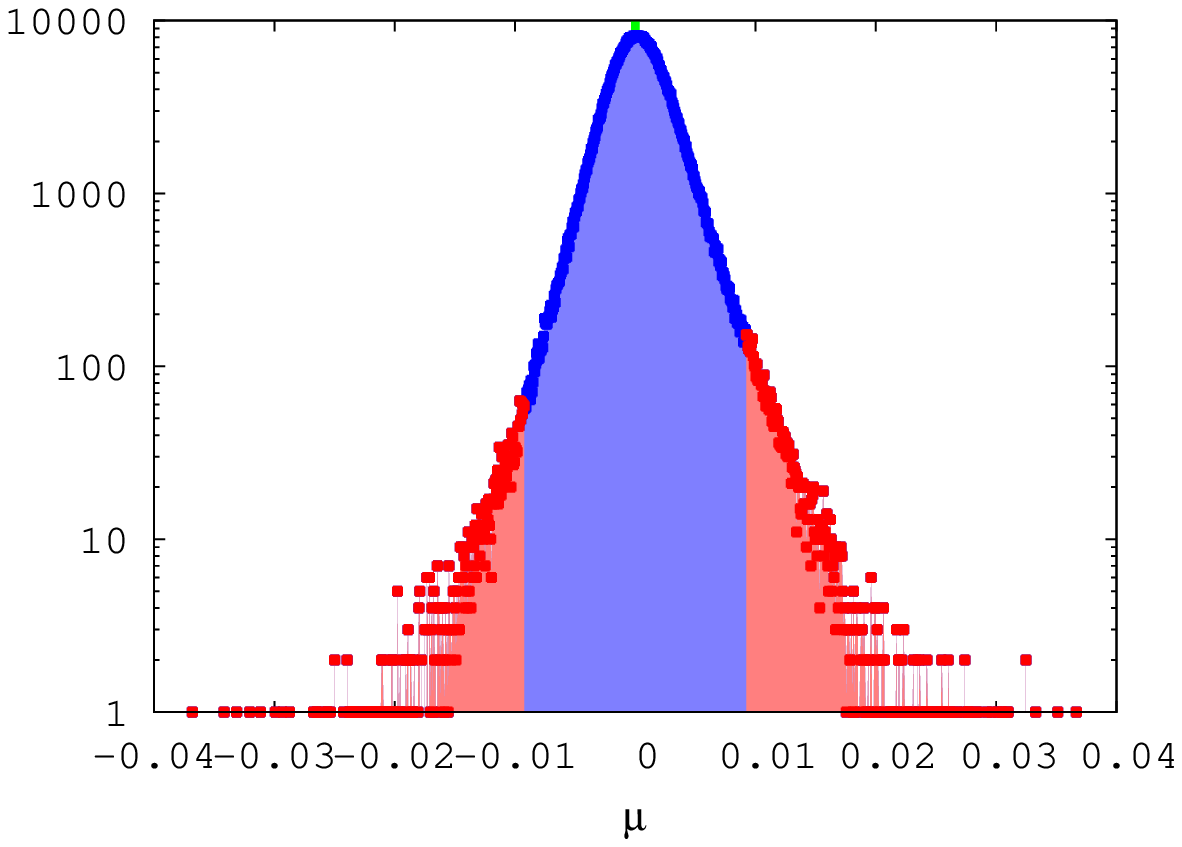}
& \includegraphics[trim=20bp 0bp 0bp 0bp,clip,width=.32\linewidth]{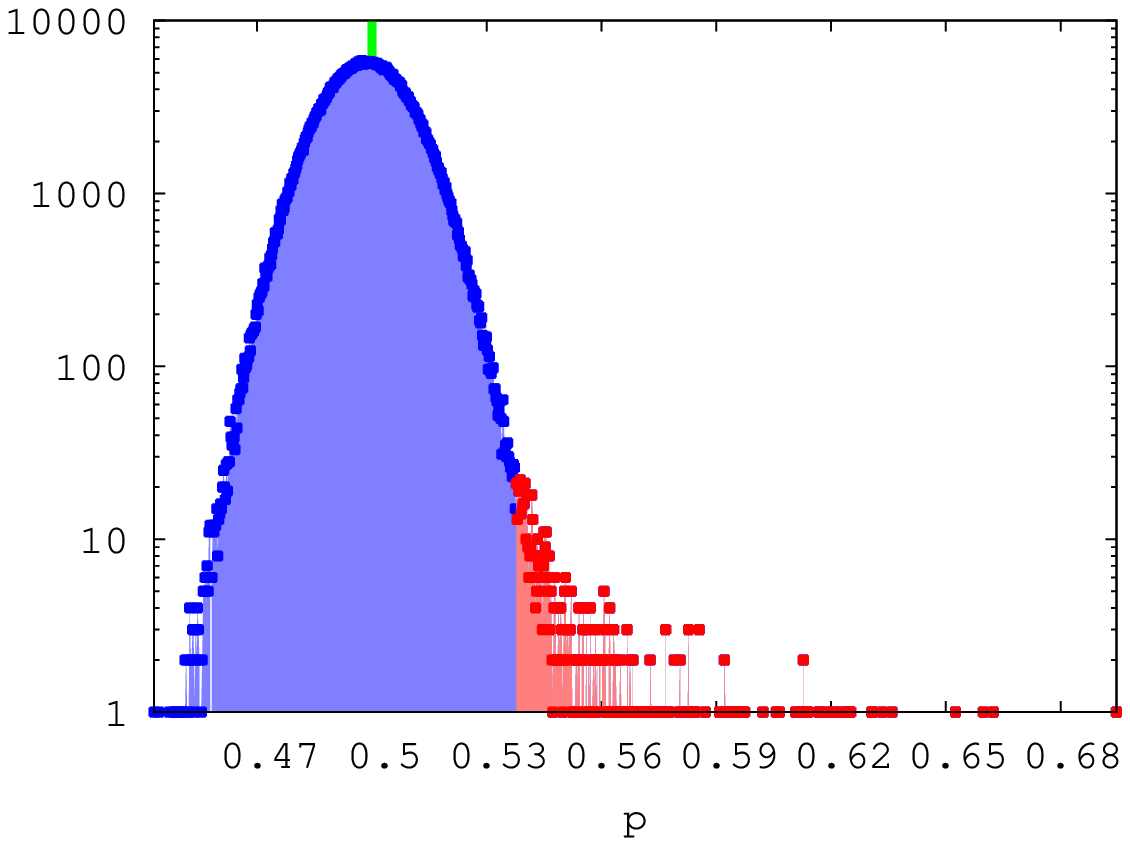} 
& \includegraphics[trim=20bp 0bp 0bp 0bp,clip,width=.32\linewidth]{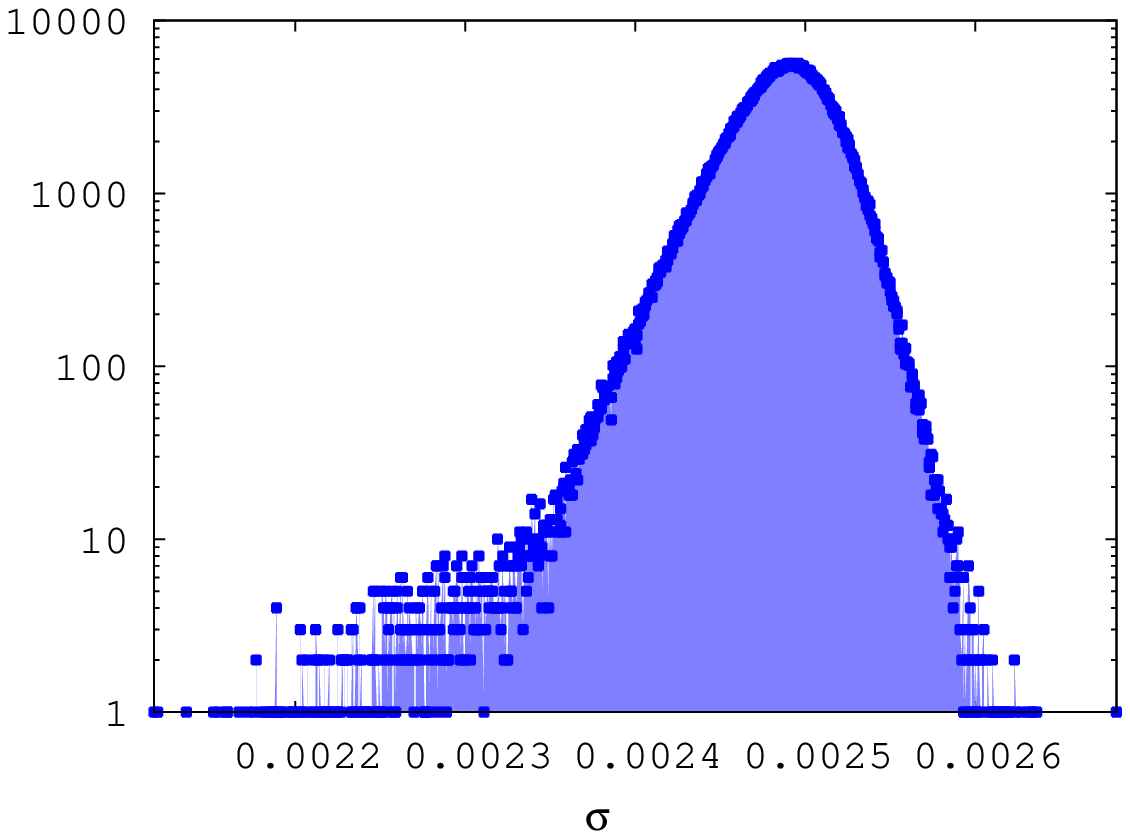} 
\end{tabular}
}
\caption{Counting histograms ($y$, blue + red) for the values of $\mu$ (left), $p$
  (center) and $\sigma$ (right, $y$-scales are log-scales). The vertical green segment
  indicates $\mu = 0$ (left) and $p=.5$ (center). The red part
  displays the upper $99\%$ percentile for $|\mu|$ (left) and upper
  $99.9\%$ percentile for $p$ (center).\label{fig:MUP}}
\end{figure}

We have analyzed the signals of 799 (one gene was missing in our data,
out of the
800 tagged in the original paper) sentinels of the yeast cell
cycle (YCC) from \cite{ssz+CI}, for a total of $\approx$13,600 data
points. Figure \ref{fig:MUP} presents the counting histograms for
$\mu$ and $p$ found among all inferred arcs. Let us denote as
\textit{strong} arcs arcs that jointly belong to the red areas of both
curves (meaning that both $p$ is in top 99.9$\%$ quantile \textit{and} $|\mu|$ in
  top 99$\%$ quantile). We remark that the scale for $\sigma$ is
  roughly in the tenth of that for $\mu$, so that for strong arcs, distributions with $|\mu|$ in
  its top 99$\%$ quantile can be considered encoding non-void
  arc connection (even when small in an absolute scale). We also
  notice that the distribution in $p$ admits relatively large values
  ($\approx 0.7$),
  so that its top $99.9\%$ percentile can be encoding arc probability
  strictly larger than $1/2$.

\begin{table}[t]
\centering
\scalebox{0.7}{
\begin{tabular}{l|l|l}\hline\hline
G1(P) & FKS1, CLN3, CDC47, RAD54, PCL2, MNN1, RAD53, CLB5 & 8/16\\ \hline
G1/S & DPB2, CDC2, PRI2, POL12, CDC9, CDC45, CDC21, RNR1, CLB6, POL1,
MSH2, RAD27, ASF1,
 & \\
 & POL30, RFA2, PMS1, MST1, RFA1, MSH6, SPC42, CLN2, PCL1, RFA3 & 23/28 \\
 \hline
S & MCD1, HTA2, SWE1, HTB1, KAR3, HSL1, HHF2, HHT1, HTB2, CIK1, CLB4  & 11/17\\\hline
G2 & CLB1, CLB2, BUD8, CDC5 & 4/4\\\hline
G2/M & SWI5, CWP1, CHS2, FAR1, DBF2, MOB1, ACE2, CDC6 & 8/9\\\hline
M(P) & CDC20 & 1/2\\\hline
M(M) & TEC1, RAD51, NUM1 & 3/4\\\hline
M(A) & TIP1, SWI4, KIN3, ASF2, ASH1, SIC1, PCL9, EGT2, SED1 & 9/15\\\hline
M(T) & $\emptyset$ & 0/1\\\hline
M/G1 & PSA1, RME1, CTS1 & 3/3\\\hline
G1 & HO & 1/4\\\hline
late G1 & $\emptyset$ & 0/3
\\ \hline\hline
\end{tabular}
}
\caption{Genes found in at least one arc with $p$ in top 99.9$\%$
  quantile or $|\mu|$ in
  top 99$\%$ quantile, in the list of 106 documented genes of the cell
  cycle in \cite{ccwscwwglldAG,rkt+HR}, as a function of the phase as
  defined in \cite{ccwscwwglldAG} (left). The right column mentions the
  number of genes retrieved $/$ total number of genes in the original
  list (for example, \textit{all} G2 genes appear).\label{tab:YCCLIST}}
\end{table}
We have analyzed arcs belonging to at least one of these categories ($p$ is in top 99.9$\%$ quantile \textit{or} $|\mu|$ in
  top 99$\%$ quantile), the intersection of both representing strong
  arcs. Intuitively, this top list should contain most
  of the (much shorter) "A-lists" of cell-cycle genes as recorded in the
  litterature. One of these lists 
  \citep{ccwscwwglldAG} has been curated and can be retrieved from
  \citep[Table 4 SI]{rkt+HR}. It contains 106 genes. Table
  \ref{tab:YCCLIST} gives the genes we retrieve, meaning that at least one
  \textit{significant} arc appear for each of them ($p$ is in top 99.9$\%$ quantile \textit{or} $|\mu|$ in
  top 99$\%$ quantile). The values given in the Table allow to
  concluce that almost $68\%$ of the 106 genes are retrieved as
  having at least one significant arc. Since the total number of
  genes with strong arcs we retrieve is 177, out of the 799, the
  probability that the result observed in Table \ref{tab:YCCLIST} is
  due to chance is zero up to more than \textit{thirty} digits. Hence, assuming
  the list of genes in Table \ref{tab:YCCLIST} is indeed a most
  important one, we can conclude in the reliability of our technique
  for network discovery for this domain.

\begin{table}[t]
\centering
\scalebox{0.925}{
\begin{tabular}{cccccc}\hline\hline
 & M & M/G1 & G1 & S & G2 \\ \hline
M & 18 & 4 & 7 & 3 & 6 \\
M/G1 & 5 & 7 & 3 & 0 & 1\\ 
G1 & 5 & 4 & 23 & 1 & 0 \\
S & 2 & 0 & 0 & 2 & 1 \\
G2 & 4 & 1 & 0 & 0 & 0 \\ \hline
\end{tabular}
}
\caption{Distribution of strong arcs ($p$ in top 99.9$\%$ quantile, $|\mu|$ in
  top 99$\%$ quantile) with respect to phases in the YCC. Each entry has been
  rounded to the nearest integer for readability.\label{tab:YCC}}
\end{table}

\begin{figure}[t]
\centering
\scalebox{0.925}{
\begin{tabular}{c}
\includegraphics[trim=250bp 110bp 180bp 100bp,clip,width=.50\linewidth]{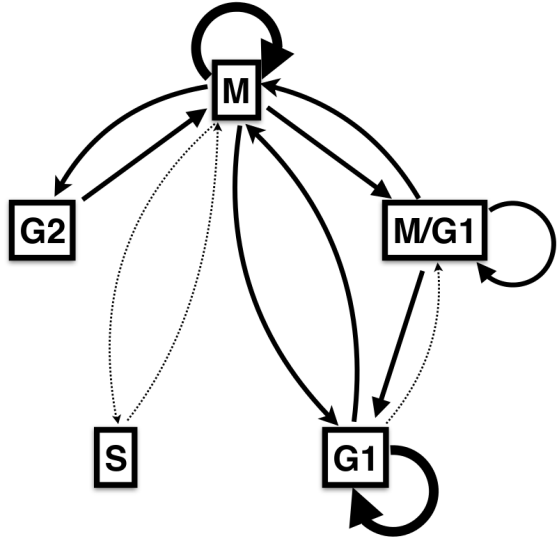} 
\end{tabular}
}
\caption{Distribution of strong arcs ($p$ in top 99.9$\%$ quantile, $|\mu|$ in
  top 99$\%$ quantile) with respect to phases in the YCC (clockwise), displayed as follows: thick plain
  $\geq 8\%$, plain $\in [4\%,8\%)$, dashed $\in (2\%,4\%)$. Reference
  values in Table \ref{tab:YCC}. \label{fig:YCC2}}
\end{figure}

As a next step, Table \ref{tab:YCC} presents the breakdown for the relative distribution
of \textit{strong} arcs in the YCC as a function of the YCC phase, using as reference the original
one from \cite{ssz+CI}, collapsing the vertices in their respective
phase of the YCC to obtain a concise graph of within and between phase
dependences (Figure \ref{fig:YCC2} gives a schematic view of the most
significant part of the distribution --- arcs between different genes
of the \textit{same} YCC phase create the loops observed). We can draw two conclusions: (i) the graph
of dependences between phases is not symmetric. Furthermore, (ii) M and
G1 appear as the phases which concentrate more than half of the strong
arcs, which should be expected given the known regulatory importance in
these two phases \citep{ssz+CI}. 
\begin{table}[t]
\centering
\scalebox{0.8}{\tiny
\begin{tabular}{ccccc}\\ \hline\hline
& Gene & Phase & out/in-degree ratio & \\ \hline
\multirow{33}{*}{\rotatebox[origin=c]{90}{in-degree $<$ out-degree}} & ASH1 & M/G1 & 5.0&\\
&YIL158W & M & 3.0&\\
&MSH6 & G1 & 3.0&\\
&SWI5 & M & 3.0&\\
&RAD53 & G1 & 3.0&\\
&YNR009W & S & 2.5&\\
&MET3 & G2 & 2.3333333333333335&\\
&YOX1 & G1 & 2.0&\\
&SVS1 & G1 & 2.0&\\
&YOL007C & G1 & 2.0&\\
&CDC20 & M & 2.0&\\
&YKR041W & G2 & 2.0&\\
&CDC5 & M & 2.0&\\
&MET28 & S & 2.0&\\
&YML034W & M & 2.0&\\
&SMC3 & G1 & 2.0&\\
&HHO1 & S & 2.0&\\
&YDL039C & M & 2.0&\\
&RAD27 & G1 & 2.0&\\
&FAR1 & M & 2.0&\\
&DIP5 & M & 1.75&\\
&YPL267W & G1 & 1.6&\\
&CDC45 & G1 & 1.5&\\
&RNR1 & G1 & 1.5&\\
&PCL9 & M/G1 & 1.5&\\
&LEE1 & S & 1.5&\\
&YOR314W & M & 1.5&\\
&YIL025C & G1 & 1.4444444444444444&\\
&AGP1 & G2 & 1.3333333333333333&\\
&CWP1 & G2 & 1.3333333333333333&\\
&ALD6 & M & 1.2&\\
&YOL132W & M & 1.1666666666666667&\\
&YNR067C & M/G1 & 1.0666666666666667&\\\hline
\multirow{32}{*}{\rotatebox[origin=c]{90}{in-degree $=$ out-degree}} &YNL078W & M/G1 & 1.0&\\
&YCL013W & G2 & 1.0&\\
&RME1 & G1 & 1.0&\\
&CLB1 & M & 1.0&\\
&RPI1 & M & 1.0&\\
&YIL141W & G1 & 1.0&\\
&BUD4 & M & 1.0&\\
&YLR235C & G1 & 1.0&\\
&YOR315W & M & 1.0&\\
&YER124C & G1 & 1.0&\\
&YPR156C & M & 1.0&\\
&YGL028C & G1 & 1.0&\\
&BUD3 & G2 & 1.0&\\
&STE3 & M/G1 & 1.0&\\
&HST3 & M & 1.0&\\
&ALK1 & M & 1.0&\\
&CHS2 & M & 1.0&\\
&YLL061W & S & 1.0&\\
&YFR027W & G1 & 1.0&\\
&LAP4 & G1 & 1.0&\\
&YNL173C & M/G1 & 1.0&\\
&YML033W & M & 1.0&\\
&SEO1 & S & 1.0&\\
&YOR264W & M/G1 & 1.0&\\
&NUF2 & M & 1.0&\\
&YOR263C & M/G1 & 1.0&\\
&YBR070C & G1 & 1.0&\\
&YNL300W & G1 & 1.0&\\
&YPR045C & M & 1.0&\\
&YOR248W & G1 & 1.0&\\
&MYO1 & M & 1.0&\\
&RLF2 & G1 & 1.0&\\\hline
\multirow{35}{*}{\rotatebox[origin=c]{90}{in-degree $>$ out-degree}} &YOL101C & M/G1 & 0.9444444444444444&\\
&YDR355C & S & 0.8&\\
&HTB2 & S & 0.75&\\
&YRO2 & M & 0.7142857142857143&\\
&YDR380W & M & 0.6666666666666666&\\
&FET3 & M & 0.6666666666666666&\\
&YDL163W & G1 & 0.6666666666666666&\\
&CLB6 & G1 & 0.6666666666666666&\\
&ECM23 & G2 & 0.6666666666666666&\\
&YBR089W & G1 & 0.6666666666666666&\\
&YGR221C & G1 & 0.6666666666666666&\\
&YDL037C & M & 0.6&\\
&MF(ALPHA)2 & G1 & 0.6&\\
&YLR183C & G1 & 0.5714285714285714&\\
&PDR12 & M & 0.5555555555555556&\\
&YER150W & M/G1 & 0.5555555555555556&\\
&POL1 & G1 & 0.5&\\
&YHR143W & G1 & 0.5&\\
&SPS4 & M & 0.5&\\
&PCL1 & G1 & 0.5&\\
&YGL184C & S & 0.5&\\
&EGT2 & M/G1 & 0.5&\\
&CTS1 & G1 & 0.5&\\
&YDR149C & G2 & 0.5&\\
&GAP1 & G2 & 0.5&\\
&HO & G1 & 0.5&\\
&WSC4 & M & 0.4673913043478261&\\
&SPO16 & G1 & 0.46153846153846156&\\
&YMR032W & M & 0.4&\\
&YGP1 & M/G1 & 0.4&\\
&SPH1 & G1 & 0.3333333333333333&\\
&YCL022C & G1 & 0.3333333333333333&\\
&YCLX09W & G2 & 0.3333333333333333&\\
&PIR1 & M/G1 & 0.25&\\
&ARO9 & M & 0.16666666666666666&\\
\hline\hline
\end{tabular}
}
\caption{Imbalancedness of the network: genes in decreasing ratio
  out-degree/in-degree, computed using strong arcs ($p$ in top 99.9$\%$ quantile, $|\mu|$ in
  top 99$\%$ quantile). Only those with $>0$ out-degree and in-degree
  are shown. A star (*) indicates reported targets for cell-cycle
  activators \citep{sbhhrvwzgjySR}.\label{tab:OUTIN}}
\end{table}
To make more precise in observation (i) that
the network is indeed imbalanced, we have computed the ratio
out-degree $/$ in-degree for all genes admitting \textit{strong} edges
of both kinds (\textit{i.e.} with the gene as in- / out- node). Table
\ref{tab:OUTIN} presents all genes collected. A total of 100 genes is
found, the majority of which (68) is imbalanced. We also remark that
roughly 80$\%$ of them is associated to M and/or G1 (only 19 are
associated to phases S or G2), which is consistent with the findings
of Table \ref{tab:YCC}. 

\begin{table}[t]
\centering
\scalebox{0.8}{
\begin{tabular}{ccc}\\ \hline\hline
Gene & Phase & out-degree\\ \hline
WSC4 & M & 43\\
YOL101C & M/G1 & 17\\
YNR067C & M/G1 & 16\\
YIL025C & G1 & 13\\
YDL037C & M & 12\\
YOR264W & M/G1 & 10\\
YER124C & G1 & 10\\
HO & G1 & 9\\
YPL267W & G1 & 8\\
YLR183C & G1 & 8\\
MET3 & G2 & 7\\
DIP5 & M & 7\\
SEO1 & S & 7\\
YOL132W & M & 7\\
ALD6 & M & 6\\
YGL028C & G1 & 6\\
PCL9 & M/G1 & 6\\
SPO16 & G1 & 6\\
YDL039C & M & 6\\
YOL007C & G1 & 6\\
YER150W & M/G1 & 5\\
YRO2 & M & 5\\
PDR12 & M & 5\\
PCL1 & G1 & 5\\
YNR009W & S & 5\\
RME1 & G1 & 5\\
ASH1 & M/G1 & 5\\
AGP1 & G2 & 4\\
GAP1 & G2 & 4\\
YOR263C & M/G1 & 4\\
YNL173C & M/G1 & 4\\
CWP1 & G2 & 4\\
FAR1 & M & 4\\
MCD1 & G1 & 4\\
YOX1 & G1 & 4\\
YDR355C & S & 4\\
YOR314W & M & 3\\
MSH6 & G1 & 3\\
SPT21 & G1 & 3\\
LEE1 & S & 3\\
YIL158W & M & 3\\
YLR049C & G1 & 3\\
RNR1 & G1 & 3\\
HTB2 & S & 3\\
GLK1 & M/G1 & 3\\
SWI5 & M & 3\\
MF(ALPHA)2 & G1 & 3\\
CDC45 & G1 & 3\\
RAD53 & G1 & 3\\ \hline\hline
\end{tabular}
}
\caption{Genes in decreasing out-degree for strong arcs ($p$ in top 99.9$\%$ quantile, $|\mu|$ in
  top 99$\%$ quantile). Only those with out-degree $\geq 3$ are shown.\label{tab:OUT}}
\end{table}

\begin{table}[t]
\centering
\scalebox{0.8}{
\begin{tabular}{ccc}\\ \hline\hline
Gene & Phase & in-degree\\ \hline
WSC4 & M & 92\\
YDL037C & M & 20\\
HO & G1 & 18\\
YOL101C & M/G1 & 18\\
YNR067C & M/G1 & 15\\
YLR183C & G1 & 14\\
SPO16 & G1 & 13\\
YOR264W & M/G1 & 10\\
YER124C & G1 & 10\\
PCL1 & G1 & 10\\
YIL025C & G1 & 9\\
PDR12 & M & 9\\
YER150W & M/G1 & 9\\
GAP1 & G2 & 8\\
SEO1 & S & 7\\
YRO2 & M & 7\\
ARO9 & M & 6\\
SPH1 & G1 & 6\\
YOL132W & M & 6\\
YGL028C & G1 & 6\\
MF(ALPHA)2 & G1 & 5\\
YMR032W & M & 5\\
ALD6 & M & 5\\
YPL267W & G1 & 5\\
YGP1 & M/G1 & 5\\
YDR355C & S & 5\\
RME1 & G1 & 5\\
YOR263C & M/G1 & 4\\
YGL184C & S & 4\\
PCL9 & M/G1 & 4\\
PIR1 & M/G1 & 4\\
HTB2 & S & 4\\
DIP5 & M & 4\\
YNL173C & M/G1 & 4\\
SPS4 & M & 4\\
YCL022C & G1 & 3\\
YOL007C & G1 & 3\\
MET3 & G2 & 3\\
YGR221C & G1 & 3\\
YDL039C & M & 3\\
CWP1 & G2 & 3\\
YCLX09W & G2 & 3\\
YDR380W & M & 3\\
AGP1 & G2 & 3\\
CLB6 & G1 & 3\\
YBR089W & G1 & 3\\
FET3 & M & 3\\
YDL163W & G1 & 3\\
ECM23 & G2 & 3\\\hline\hline
\end{tabular}
}
\caption{Genes in decreasing in-degree for strong arcs ($p$ in top 99.9$\%$ quantile, $|\mu|$ in
  top 99$\%$ quantile). Only those with in-degree $\geq 3$ are shown.\label{tab:IN}}
\end{table}
To finish with the quantitative analyses, Tables \ref{tab:IN} and
\ref{tab:OUT} present the main \textit{strong} genes in term of in or
out degree (genes with in or out-degree $<3$ are not shown). Notice
the preeminence of two well known cell-cycle
regulated genes, HO and WSC4.

\begin{figure}[t]
\centering
\scalebox{0.925}{
\begin{tabular}{cc}
\includegraphics[trim=120bp 0bp 130bp 15bp,clip,width=.50\linewidth]{Figs/Spellman_DICE_Klein_New} 
& \includegraphics[trim=50bp 0bp 30bp 15bp,clip,width=.55\linewidth]{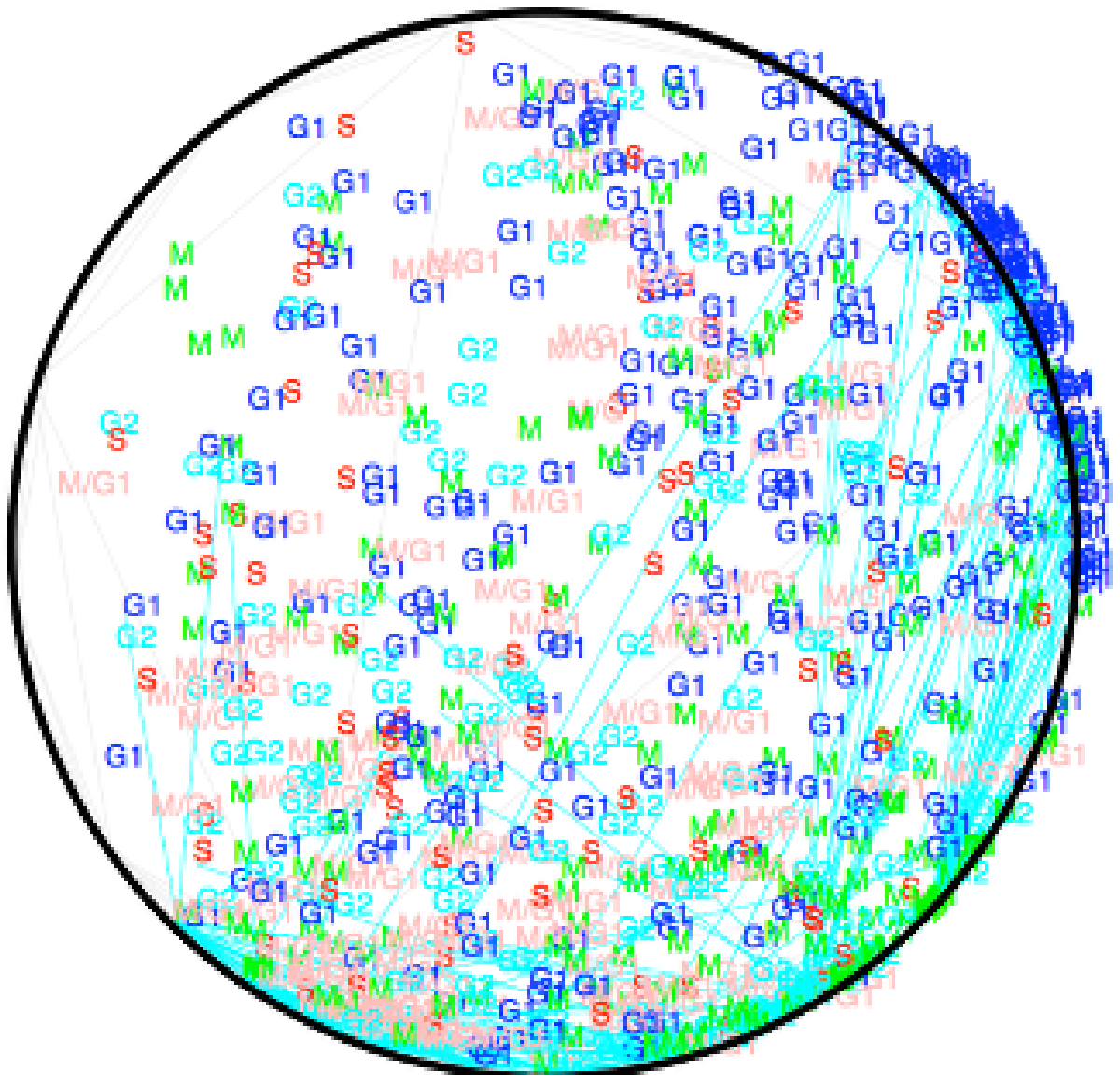} 
\end{tabular}
}
\caption{Manifold coordinates learned from the symmetrized $p$-values
  graph using \cite{msLS}, on Klein disk (we chose it to ease reading: the
  representation is conformal and geodesics are straight lines ---
  WSC4, which is in fact far away from all other genes, does not
  prevent a visually meaningful display of the other main genes). Segments are \textit{strong} arcs
  (arrowheads not represented). \textit{Left}: Major genes influencing the
  computation and known to be Cell Cycle Transcriptionally Regulated
  (CCTR, \cite{rkt+HR}) are displayed in plain boxes. Chromosomes are
  shown in red. \textit{Right}: zoom over the pink area in the left
  plot, showing few strong edges belong to this area, and therefore
  strong edges guide the construction of the manifold's main coordinates.\label{fig:MANI}}
\end{figure}

To catch a glimpse at the overall network found from a more
qualitative standpoint, we have learned a
coordinate system for genes based on a popular manifold learning
technique \cite{msLS}. Since this technique requires the graph to be
symmetric, we have symmetrized the network by taking the max of the
$p$-values to weight each edge. Figure \ref{fig:MANI} presents the
results obtained, for the two leading coordinates --- excluding the
coordinate associated to eigenvalue 1, which encodes the stationary
distribution of the Markov chain and is therefore trivial ---. It is
clear that the first coordinates splits key YCC genes from the rest of
the crowd (\textit{Cf} Tables \ref{tab:IN} and
\ref{tab:OUT}), also highlighting the importance of WSC4 and strong
edges to create the manifold. HO
is used to switch mating type and WSC4 is required for maintenance of
cell wall integrity \cite{sbhhrvwzgjySR}. 

Interestingly, the most prominent genes belong to a small set of
chromosomes (essentially 13, 15, 16). What is quite striking
is the fact that SPS4 and SFG1 are in fact neighbors on chromosome
XV\footnote{http://www.yeastgenome.org/locus/YOR315W/overview}. It is
far beyond the scope of our paper to eventually relate the netork
structure and associated causal
influence in expression --- which we aim to capture --- to the proximity in the (physical)
loci of genes, but this is eventually worthwhile noticing and
exploiting with respect to the already known coexpression of
neighboring genes in yeast \citep{scpIC}.

\begin{figure}[t]
\centering
\scalebox{0.925}{
\begin{tabular}{c}
\includegraphics[trim=230bp 110bp 180bp 100bp,clip,width=.50\linewidth]{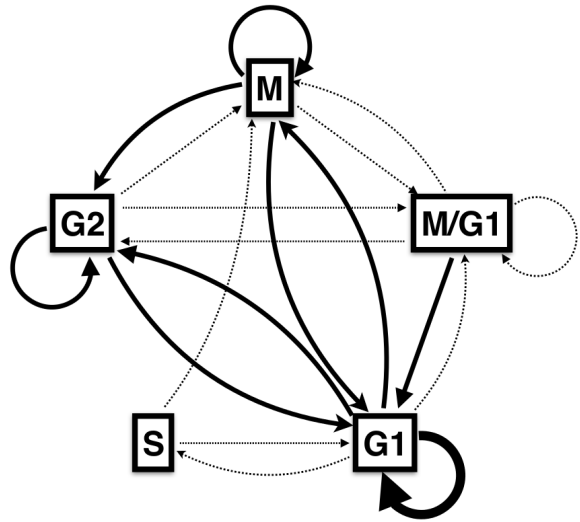} 
\end{tabular}
}
\caption{Distribution of strong arcs ($p\geq 0.05$) found by CPC,
  following Figure \ref{fig:YCC2}. Remark that the figure fails to
  carry the importance of phases M and G1, as Figure \ref{fig:YCC2}
  for \latnet~--- in particular, phase M roughly carries the same weights
  distribution as phase G2, which does not conform to observations
  (G2 is not even mandatory for the YCC while M obviously is).\label{fig:YCC_CPC}}
\end{figure}

\noindent \textbf{Comparison with \cpc} Last, we have compared our results to those of \cpc. Results are shown
in Figure \ref{fig:YCC_CPC} for the manifold (compare to Figure \ref{fig:MANI}
for our technique), and in Figure \ref{fig:YCC_CPC} for the
distribution of strong arcs found (strong in the case of \cpc~means
$p\geq 0.05$). The graph found is much closer to a complete graph,
which is a quite irrealistic observation since cell cycles are
extremely inbalanced in terms of importance with respect to
regulation. Furthermore, as remarked below in a more quantitative way,
it is known that \textit{Saccharomyces cerevisiae} tends to have a predominant
gap phase G1 compared to G2, which is clearly less visible from the
\cpc~results compared to \latnet's results.

\begin{figure}[t]
\centering
\scalebox{0.925}{
\begin{tabular}{cc}
\includegraphics[trim=120bp 0bp 130bp 25bp,clip,width=.45\linewidth]{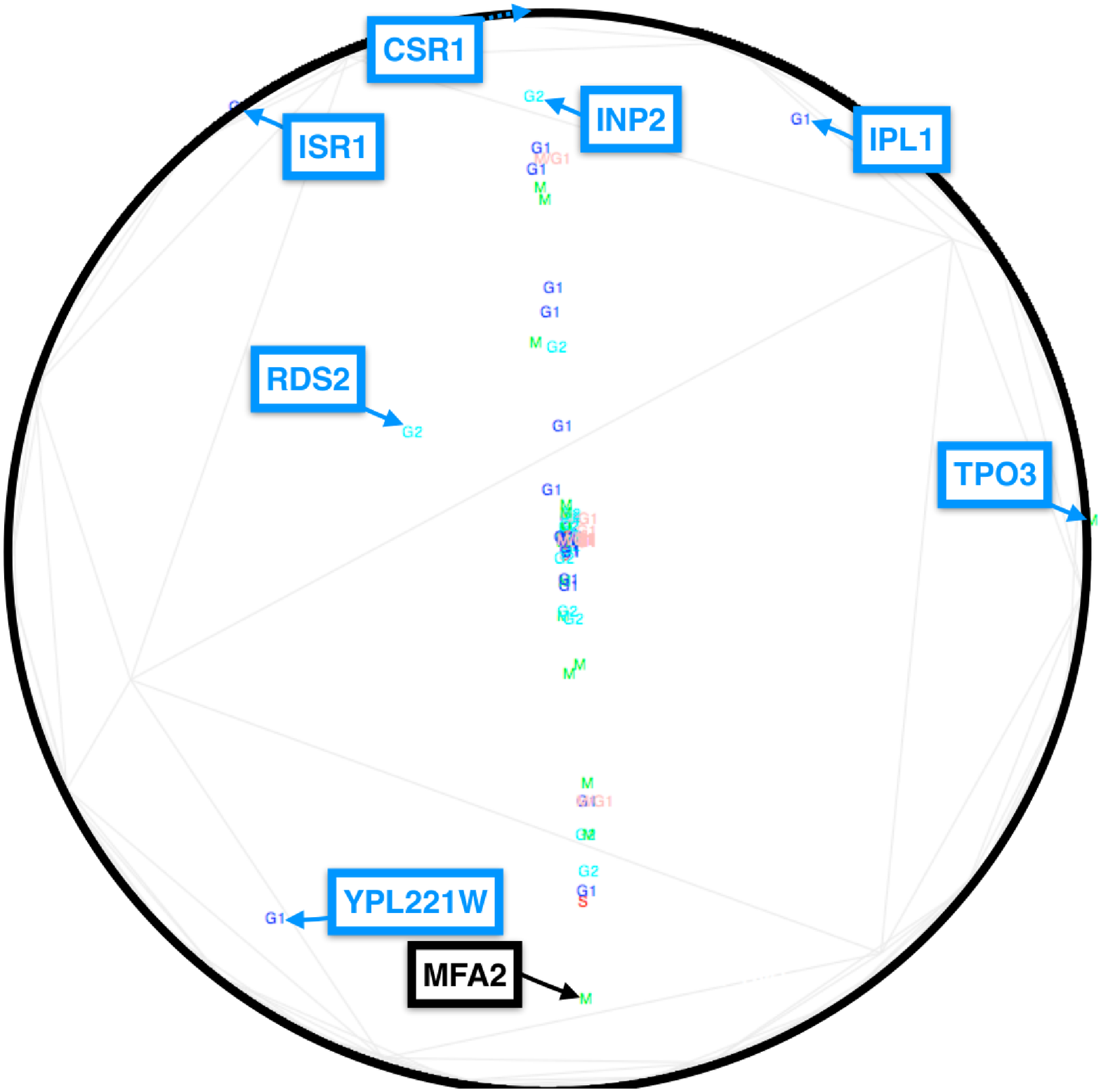} 
& \includegraphics[trim=120bp 0bp 130bp 25bp,clip,width=.45\linewidth]{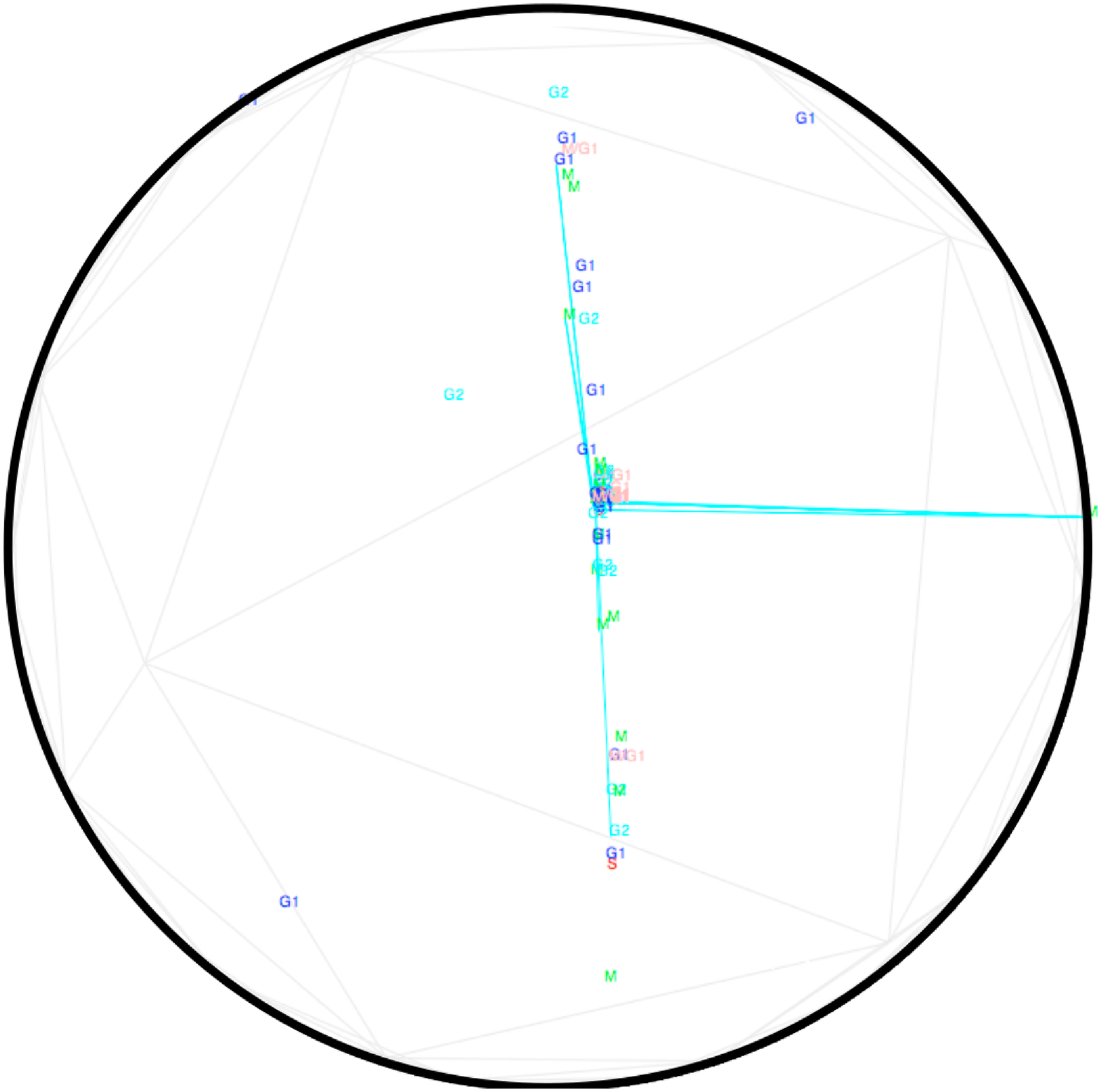} 
\end{tabular}
}
\caption{Left: manifold learned from \cpc, using the same convention as
  for \latnet. Remark that none of the \textit{edges} learned by \cpc~appears,
  because they are all concentrated inside several blobs that belong
  to the visible vertical line in the center. Genes displayed in blue
  are those extremely localed genes that do \textit{not} belong to the
  genes in \citep[Table 4 SI]{rkt+HR}. Right: we have substituted the
  edges learned by \cpc~by ours, showing that the most important genes
  in fact belong to the central blob of the picture, therefore not
  discriminative of the YCC genes.\label{fig:MANICPC}}
\end{figure}

\subsection{Analysis of the complete yeast genome}\label{sec-yeast}

\begin{figure}[t]
\centering
\scalebox{0.925}{
\begin{tabular}{ccc}
\includegraphics[trim=20bp 0bp 0bp 0bp,clip,width=.32\linewidth]{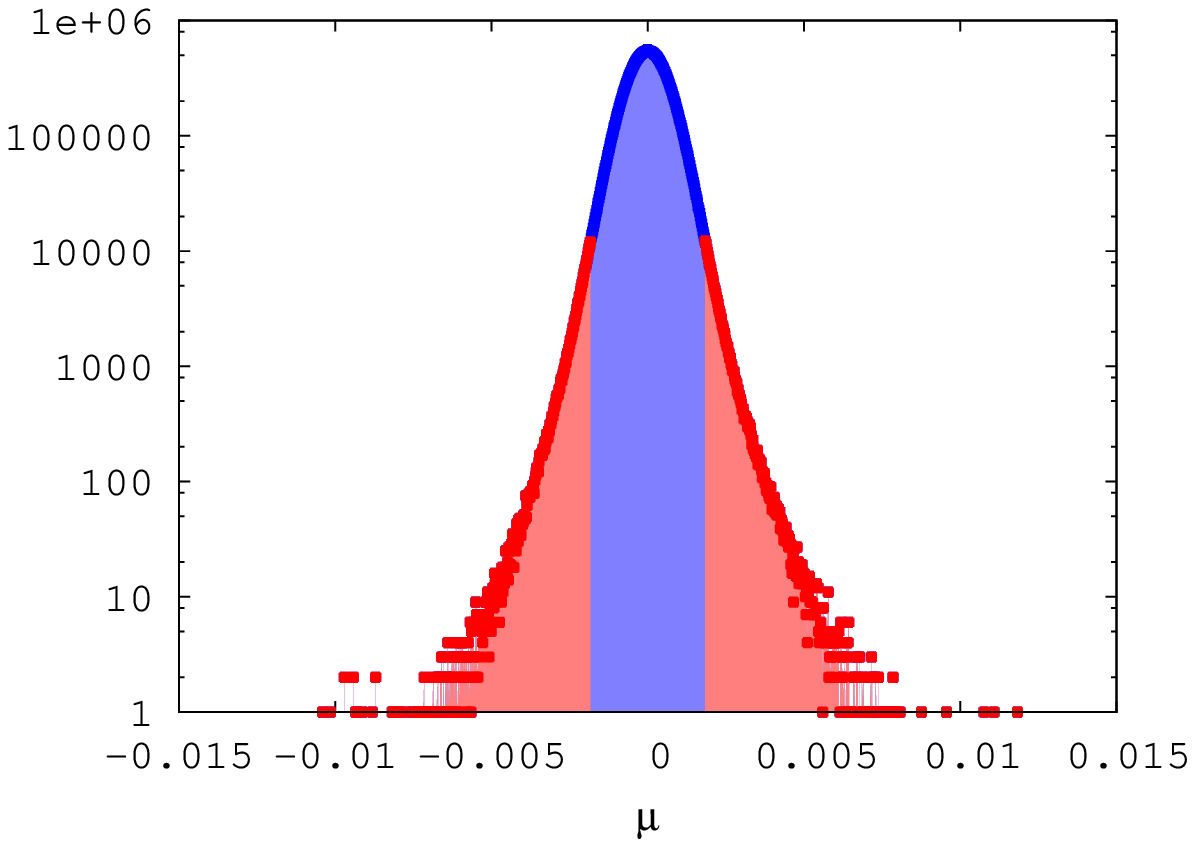}
& \includegraphics[trim=20bp 0bp 0bp 0bp,clip,width=.32\linewidth]{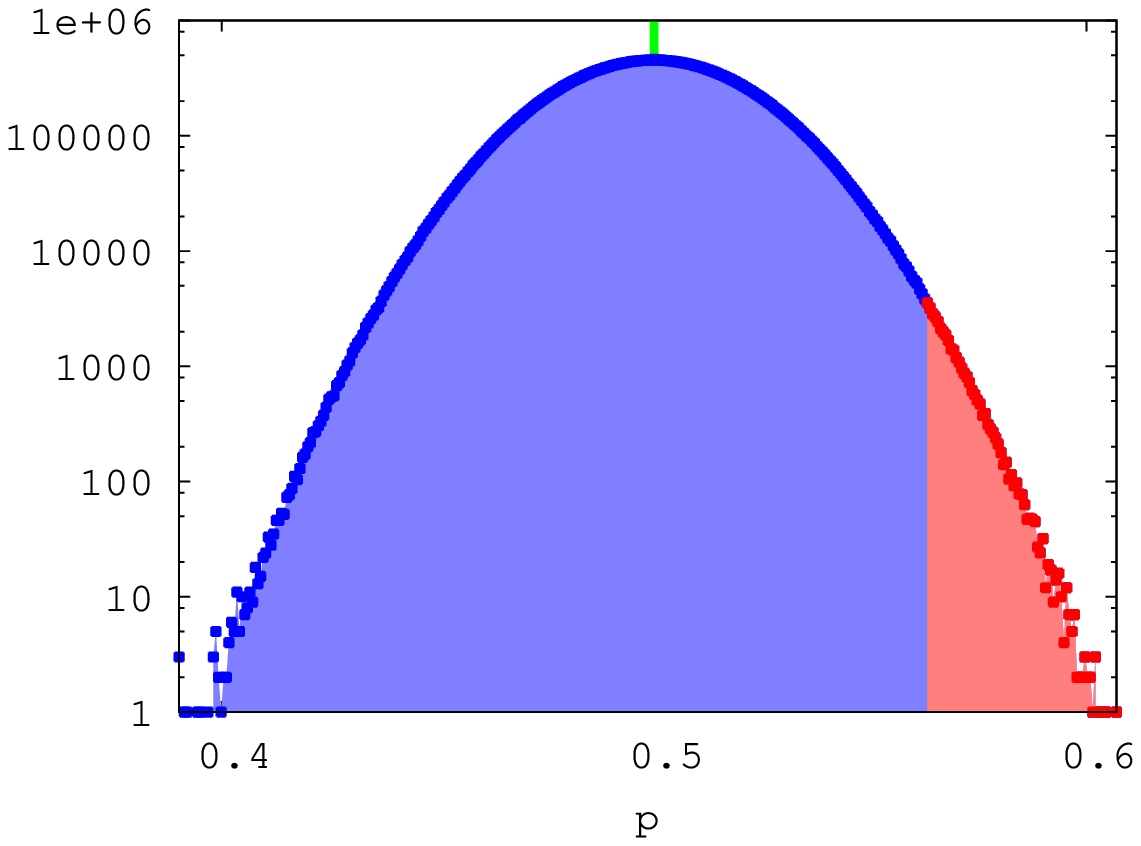} 
& \includegraphics[trim=20bp 0bp 0bp 0bp,clip,width=.32\linewidth]{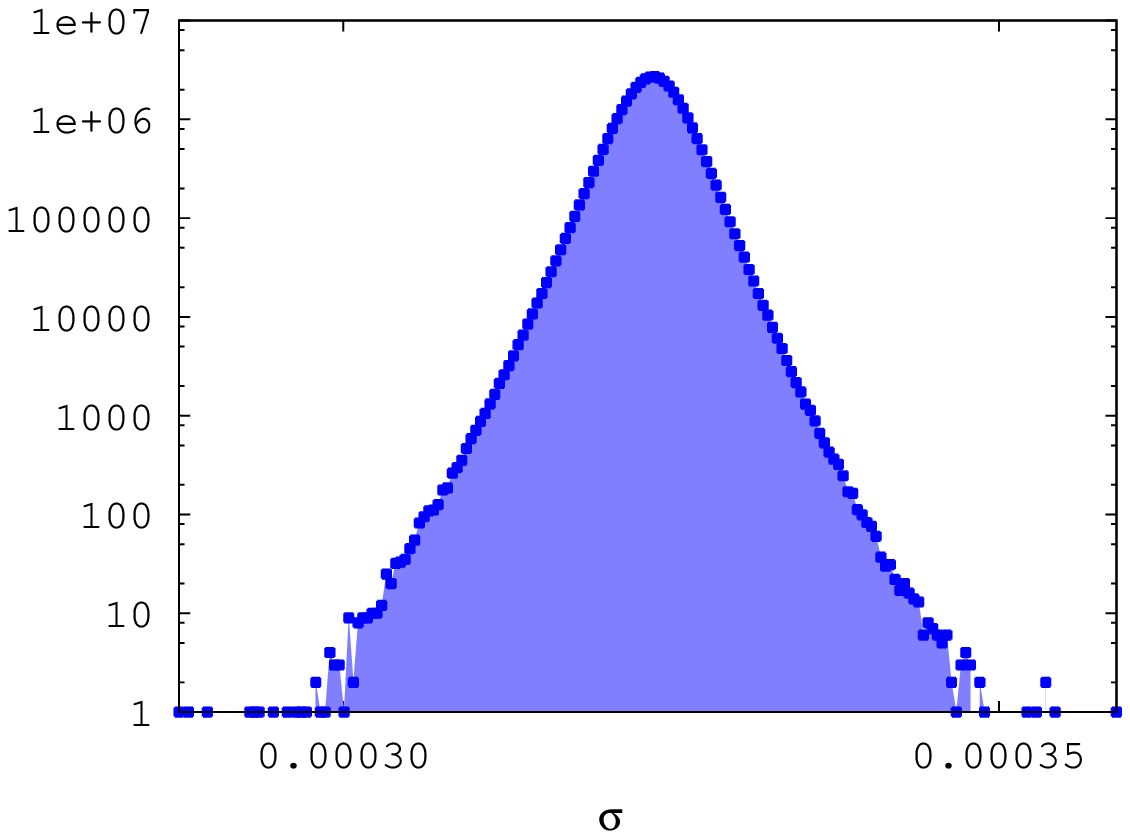} 
\end{tabular}
}
\caption{Counting histograms ($y$, blue + red) for the values of $\mu$ (left), $p$
  (center) and $\sigma$ (right, $y$-scales are
  log-scales). Conventions follow Figure \ref{fig:MUP}. A tiny
  fraction ($<1\text{\textperthousand}$) of arcs found have $p$ or
  $\sigma$ close to zero; they are not shown to save readability.\label{fig:MUPALL}}
\end{figure}

We have analyzed the complete set of 6178 genes (representing now
$100,000+$ data points) in the yeast genome
data \cite{ssz+CI}. Not that this time, this represents a maximum of more than 38
millions arcs in total in the network.

Table \ref{tab:YCOMPLETE} presents the breakdown in percentages
between YCC genes and non-YCC genes. Clearly, the graph is heavily non
symmetric: while roughly 11$\%$ of strong arcs come from outside of
the YCC to inside the YCC, more than 25$\%$ of these strong arcs come
from inside the YCC to outside the YCC. The largest percentage of arcs
between YCC - nonYCC is obtained from G1 onto the non YCC genes
($>10\%$), which seems to be plausible, since G1 is a gap phase
involving a lot of interactions with the environment, testing for
nutrient supply and growth availability. Interestingly, the strong
arcs are recalibrated to take into account the complete set of genes
(strong arcs are defined with respect to quantiles in data), yet the
relative proportions in the YCC still denote the predominance of M and
G1, and the very small percentage of strong arcs for phases S and
G2. We notice that the predominance of phase G1 compared to G2 is in
perfect accordance with the fact that \cite{ssz+CI} picked the yeast
\textit{Saccharomyces cerevisiae} which is indeed known to possess
long G1 phases (compared to \textit{e.g.} \textit{Saccharomyces
  pombe}).

\begin{figure}[t]
\centering
\scalebox{0.925}{
\begin{tabular}{c}
\includegraphics[trim=20bp 0bp 30bp 5bp,clip,width=.80\linewidth]{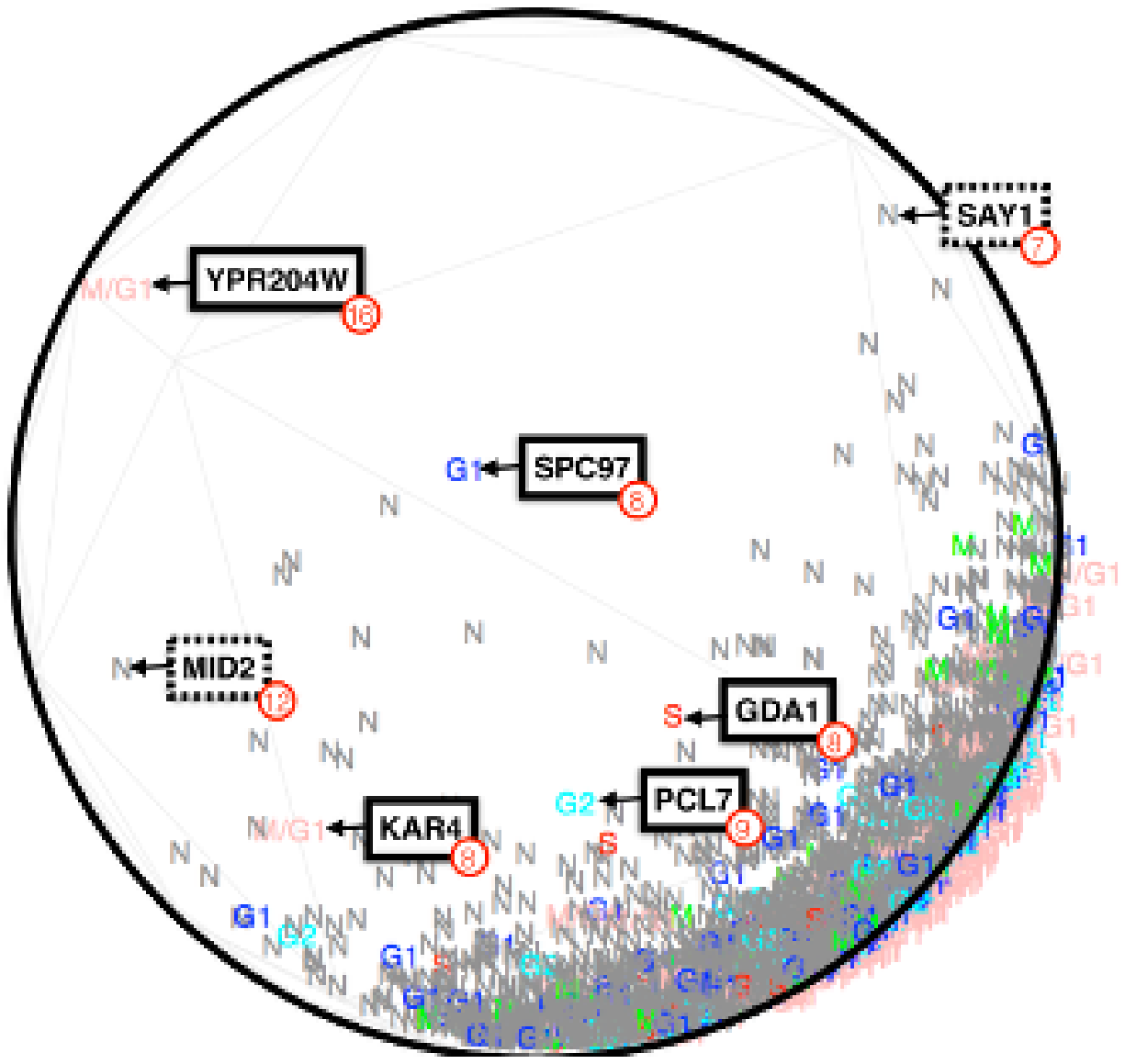} 
\end{tabular}
}
\caption{manifold obtained for \latnet~in the whole yeast genome, conventions follow
  Figure \ref{fig:MANI} (strong arcs not displayed for readability).\label{fig:MANIALL}}
\end{figure}

Finally, Figure \ref{fig:MANIALL} displays the manifold obtained for
the complete genome. We represent only a corner of the manifold of 6K+
genes, which displays this time the importance of other YCC genes,
including in particular YPR204W. This comes at no surprise: this gene
codes for a DNA helicase, a motor protein tht separates DNA
strands. DNA helicases are involved in a number of processes and not just the
YCC. We do not show strong arcs in the picture, but it is worthwhile
remarking that the relative predominance of the most prevalent YCC
genes is still here, in the whole genome analysis: WSC4, SPO16 and SLD2 are in the top-5 of
out-degree measures with strong arcs.

\begin{table}[t]
\centering
\scalebox{0.925}{
\begin{tabular}{ccccccc}\hline\hline
 & M & M/G1 & G1 & S & G2 & N\\ \hline
M & 0.6 & $\epsilon$ & 1.2 & $\epsilon$ & $\epsilon$ & 2.1\\
M/G1 & $\epsilon$ & $\epsilon$ & 0.2 & $\epsilon$ & $\epsilon$ & 1.4\\ 
G1 & 1.0 & $\epsilon$ & 1.0 & 0.2 & $\epsilon$ & 4.5\\
S & 0.2 & $\epsilon$ & 0.2 & $\epsilon$ & 0.2 & 2.0\\
G2 & 0.2 & $\epsilon$ & 0.2 & $\epsilon$ & $\epsilon$ & 1.0\\ 
N & 6.6 & 3.5 & 10.3 & 2.3 & 2.9 & 58.5\\\hline
\end{tabular}
}
\caption{Distribution of strong arcs ($p$ in top 99.9$\%$ quantile, $|\mu|$ in
  top 99$\%$ quantile) for the \textit{complete genome} of the
  yeast, including the breakdown for the YCC phases (see \textit{e.g.}
  Table \ref{tab:YCC}). "$\epsilon$" means $<0.1\%$ and "N" stands for "None" (Gene
  not in the sentinels of the YCC").\label{tab:YCOMPLETE}}
\end{table}

\subsection{Sydney property prices data}\label{sec-syd}
\begin{figure}[t]
  \centering
  \includegraphics[scale=1]{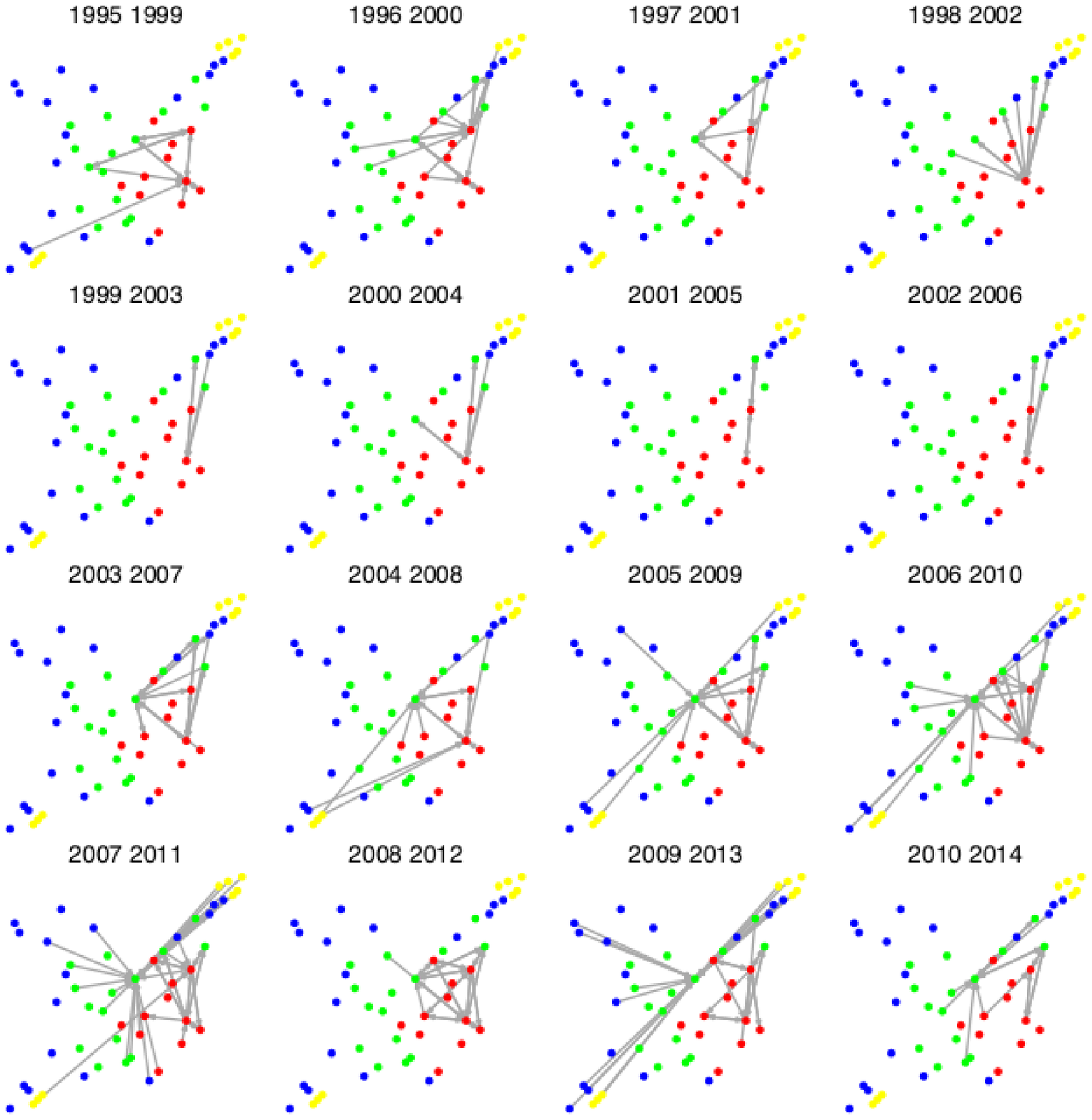}
	\caption{Associations between median house prices in different suburbs discovered by \latnet}.
	\label{fig:housing_latnet}
\end{figure}

\begin{figure}[t]
	\centering
	\includegraphics[scale=1]{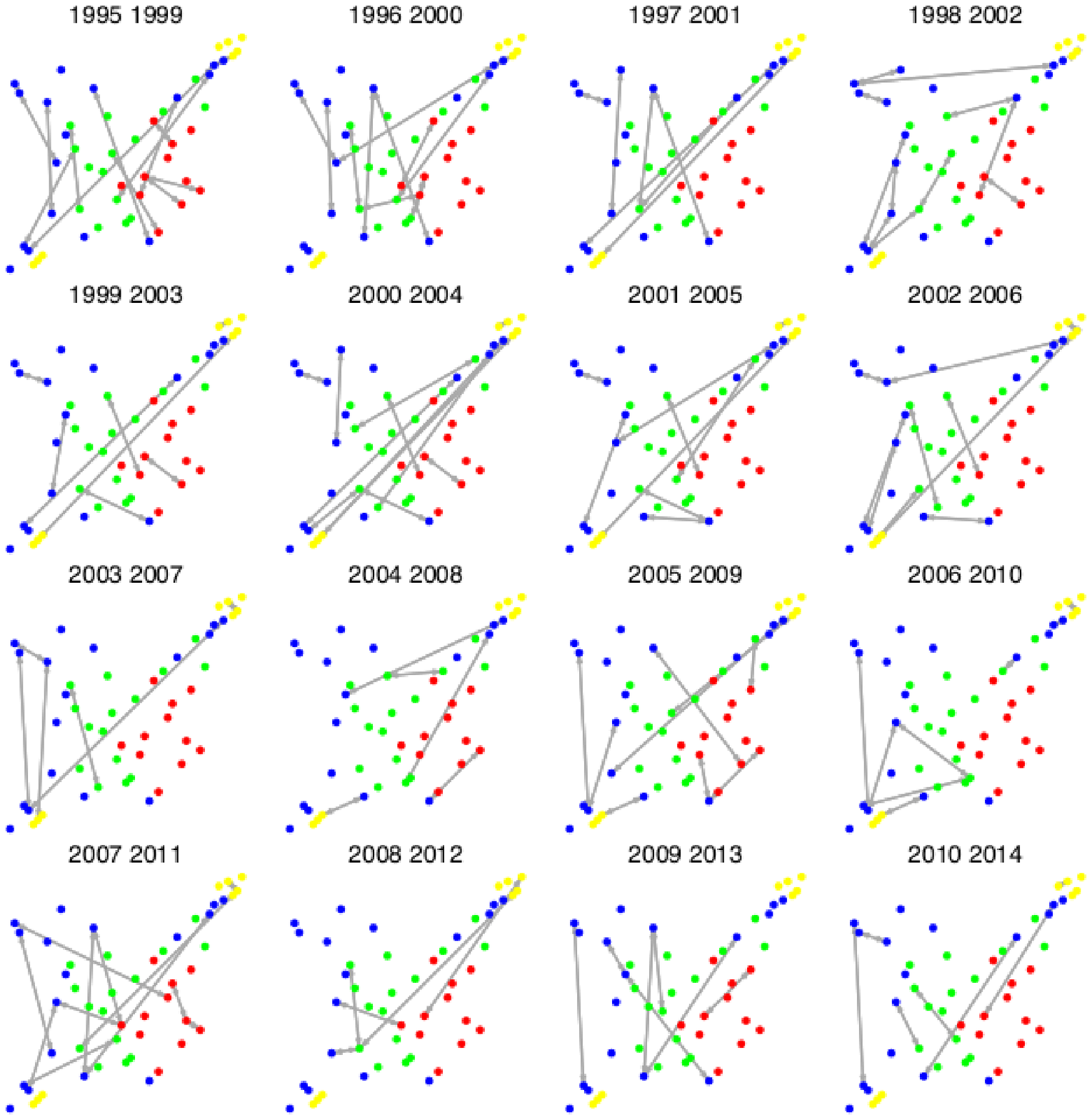}
	\caption{Associations between median house prices in different suburbs discovered by \cpc}.
	\label{fig:housing_cpc}
\end{figure}

\begin{figure}[t]
	\centering
	\includegraphics[scale=1]{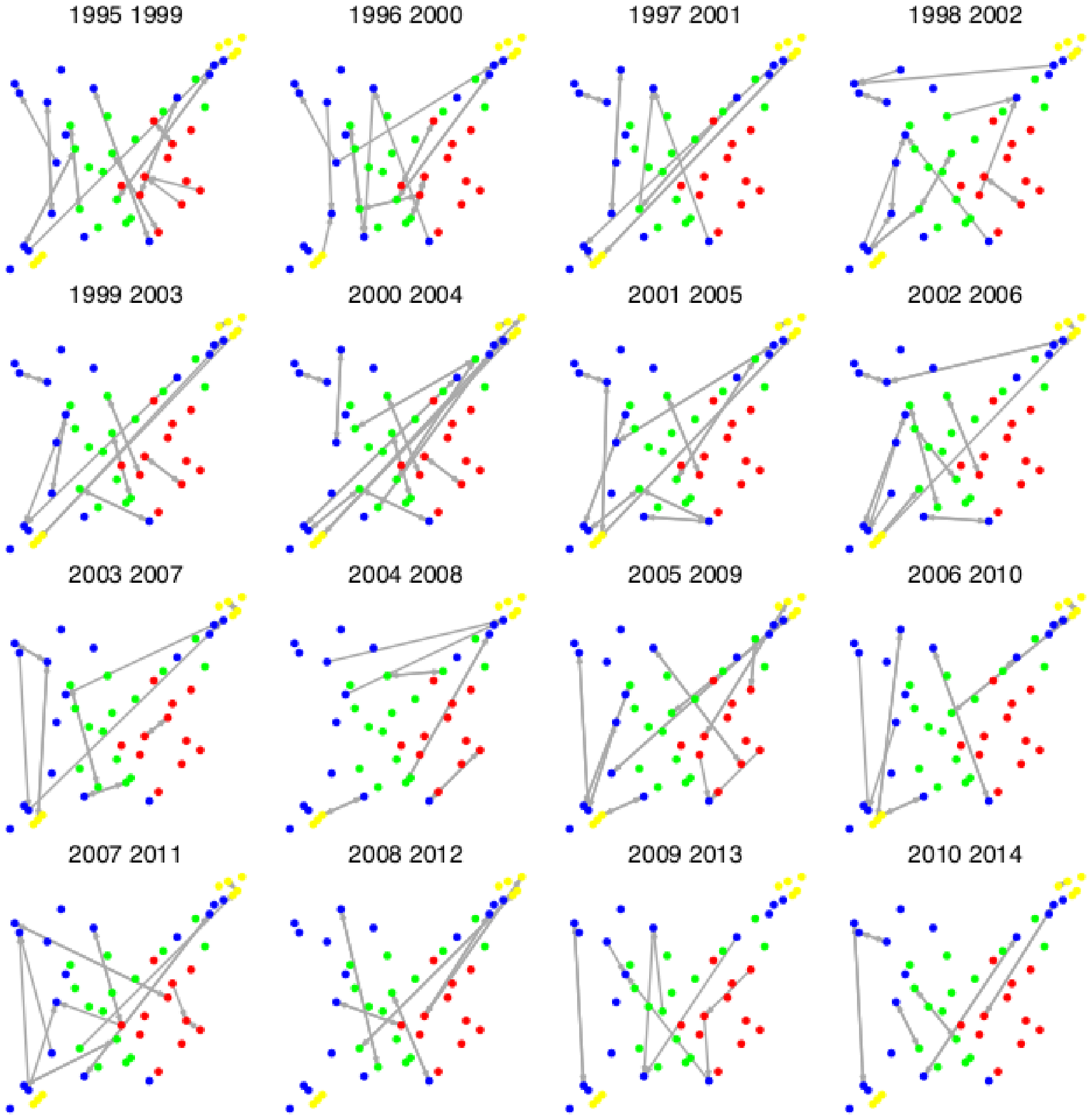}
	\caption{Associations between median house prices in different suburbs discovered by \pc}.
	\label{fig:housing_pc}
\end{figure}

\begin{figure}[t]
	\centering
	\includegraphics[scale=1]{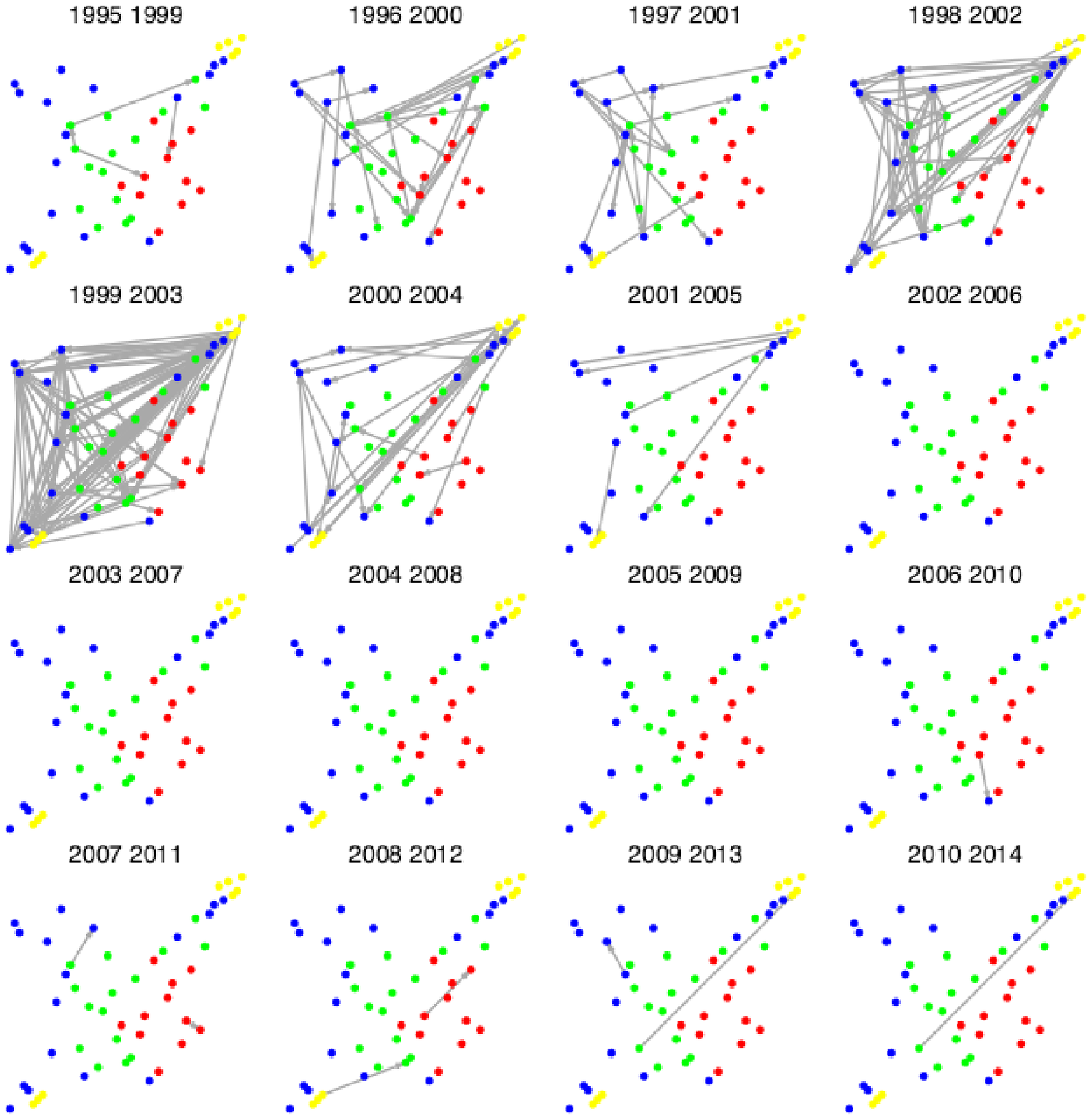}
	\caption{Associations between median house prices in different suburbs discovered by \pwlingam}.
	\label{fig:housing_pwlingam}
\end{figure}

\begin{figure}[t]
	\centering
	\includegraphics[scale=1]{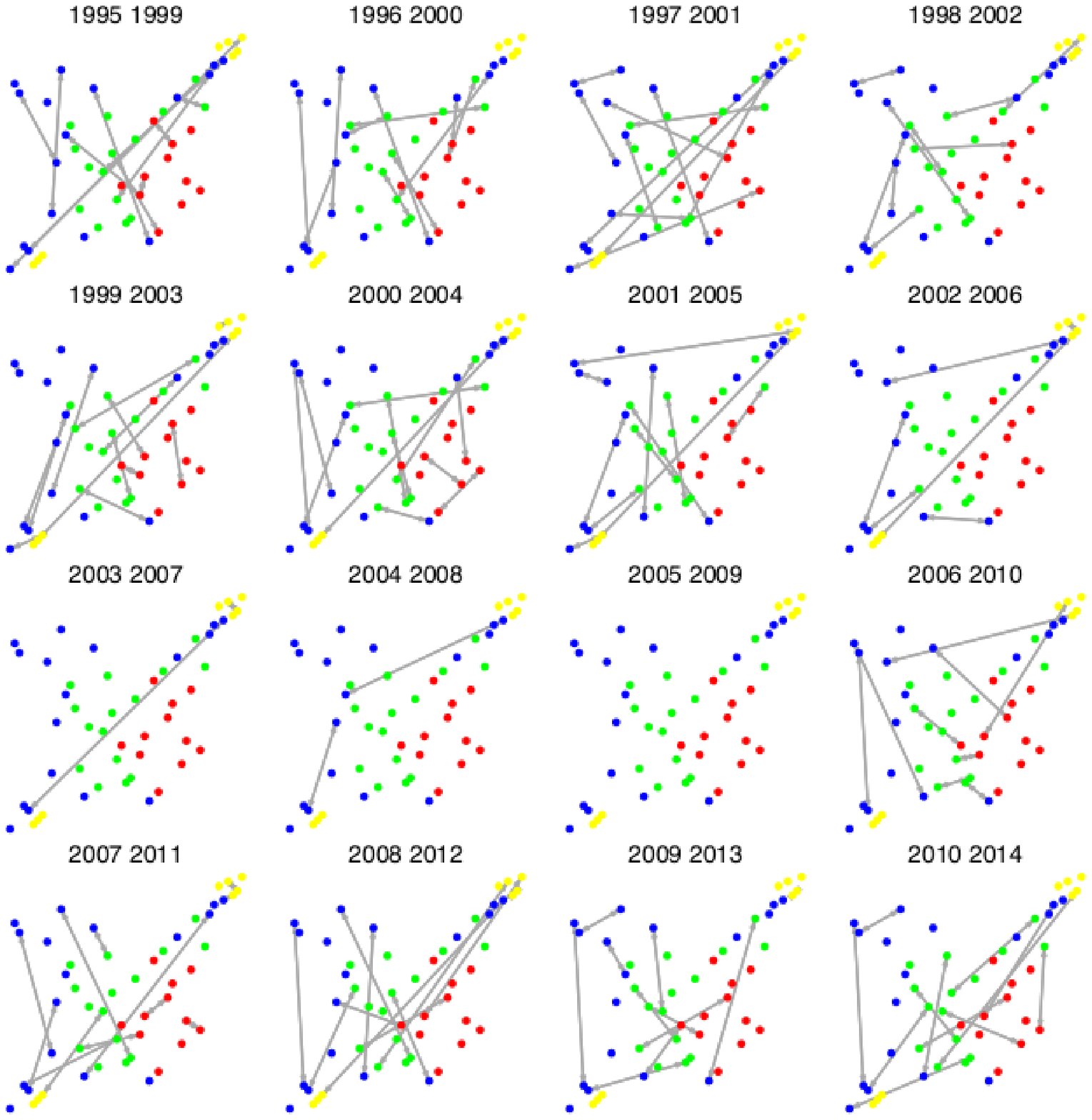}
	\caption{Associations between median house prices in different suburbs discovered by \iamb}.
	\label{fig:housing_iamb}
\end{figure}

 We set the discrimination threshold for each method so that on average each method finds 17-19 edges in the network ($p_{ij}=0.597$ for \latnet; $p$-value=$0.015$ for \pc; $p$-value=$0.012$ for \cpc; $p$-value=$10^{-7}$ for \iamb; partial correlation$=0.5$ for \pwlingam). 
Figures~\ref{fig:housing_latnet},\ref{fig:housing_cpc},\ref{fig:housing_pc},\ref{fig:housing_iamb},\ref{fig:housing_pwlingam} show the results of \latnet, \cpc, \pc\, \pwlingam\ and \iamb\ algorithms on Sydney property price data. Suburbs were ranked geographically according to their latitude and longitude coordinates, and their locations in the graphs are assigned according to their ranks. We used the ranks instead of actual coordinates of the suburbs in order to be able to better visualize connections in the inner ring. Each panel in the graphs shows the results for a certain period of time, indicated by the label above the panel. Suburbs in Sydney are divided into four groups according to their locations: inner ring (red points), middle ring (green points), outer ring (blue points), and Greater Metropolitan Region (GMR; yellow points).

Data is downloaded from: \\ \textit{http://www.housing.nsw.gov.au/about-us/reports-plans-and-papers/rent-and-sales-reports/back-issues/issue-111}

\end{document}